\newif\ifshortversion

\shortversionfalse

\ifshortversion
    \documentclass{article}
    \usepackage{iclr_files/iclr2025_conference,times,relsize,microtype}

    \title{STEER-ME: Assessing the Microeconomic Reasoning of Large Language Models}

    \author{Narun K.~Raman, Taylor Lundy, Kevin Leyton-Brown  \\
    Department of Computer Science\\
    University of British Columbia\\
    Vancouver, BC, Canada \\
    \texttt{\{narunram,tlundy,klb\}@cs.ubc.ca} \\
    \And
    Jesse Perla \\
    Department of Economics \\
    University of British Columbia \\
    Vancouver, BC, Canada \\
    \texttt{jesse.perla@ubc.ca} \\
    }
    \iclrfinalcopy 
\else
    \documentclass[format=acmsmall]{acmart}
    \usepackage{acm-ec-25}
    \usepackage{booktabs} 
    \usepackage[ruled]{algorithm2e} 
    
    \SetAlFnt{\small}
    \SetAlCapFnt{\small}
    \SetAlCapNameFnt{\small}
    \SetAlCapHSkip{0pt}
    \IncMargin{-\parindent}

    \setcitestyle{authoryear}
    
    \title[STEER-ME]{STEER-ME: Assessing the Microeconomic Reasoning of Large Language Models}

    \author{Narun K.~Raman}
    \affiliation{%
      \institution{University of British Columbia}
      \city{Vancouver}
      \country{Canada}
    }
    \email{narunram@cs.ubc.ca}
    
    \author{Taylor Lundy}
    \affiliation{%
      \institution{University of British Columbia}
      \city{Vancouver}
      \country{Canada}
    }
    \email{tlundy@cs.ubc.ca}
    
    \author{Thiago Amin}
    \affiliation{%
      \institution{University of British Columbia}
      \city{Vancouver}
      \country{Canada}
    }
    \email{thiagoamin22@gmail.com}
    
    \author{Jesse Perla}
    \affiliation{%
      \institution{University of British Columbia}
      \city{Vancouver}
      \country{Canada}
    }
    \email{jesse.perla@ubc.ca}
    
    \author{Kevin Leyton-Brown}
    \affiliation{%
      \institution{University of British Columbia}
      \city{Vancouver}
      \country{Canada}
    }
    \email{kevinlb@cs.ubc.ca}

    \begin{abstract}
        How should one judge whether a given large language model (\model) can reliably perform economic reasoning? Most existing \model benchmarks focus on specific applications and fail to present the model with a rich variety of economic tasks. A notable exception is \citet{ramansteer}, who offer an approach for comprehensively benchmarking strategic decision-making; however, this approach fails to address the non-strategic settings prevalent in microeconomics, such as supply-and-demand analysis. We address this gap by taxonomizing microeconomic reasoning into \num{58} distinct elements, focusing on the logic of supply and demand, each grounded in up to \num{10} distinct \domains, \num{5} \Tags, and \num{3} \qtypes. The generation of benchmark data across this combinatorial space is powered by a novel \model-assisted data generation protocol that we dub \autosteer, which generates a set of questions by adapting handwritten templates to target new domains and perspectives. Because it offers an automated way of generating fresh questions, \autosteer mitigates the risk that \models will be trained to over-fit evaluation benchmarks; we thus hope that it will serve as a useful tool both for evaluating and fine-tuning models for years to come. We demonstrate the usefulness of our benchmark via a case study on \num{27} \models, ranging from small open-source models to the current state of the art. We examined each model's ability to solve microeconomic problems across our whole taxonomy and present the results across a range of prompting strategies and scoring metrics.
    \end{abstract}
    
\fi

\usepackage[most]{tcolorbox}
\usepackage{lipsum} 
\usepackage{pgfkeys}
\usepackage{multirow}
\usepackage{colortbl}            
\usepackage{hyperref}            
\usepackage{booktabs}            
\usepackage{enumitem}            
\usepackage{fontawesome}         
\usepackage{float}               

\usepackage{multiaudience}

\SetNewAudience{arxiv}
\SetNewAudience{iclr}

\definecolor{setting1}{HTML}{F4CCCC}
\definecolor{setting2}{HTML}{D9EAD3}
\definecolor{setting3}{HTML}{CFE2F3}
\definecolor{setting4}{HTML}{FFF2CC}
\definecolor{setting5}{HTML}{D9D2E9}

\usepackage{tabularx}
\usepackage{enumitem}

\setlist{nolistsep}
\definecolor{green}{HTML}{66FF66}
\definecolor{myGreen}{HTML}{009900}

\usepackage{amsthm}

\usepackage[nameinlink]{cleveref}

\usepackage{nicefrac}

\usepackage{bbm}

\usepackage[T1]{fontenc}
\usepackage[utf8]{inputenc}

\let\1\undefined 
\usepackage{outlines}

\let\models\undefined 

\usepackage{wrapfig}
\usepackage{enumitem}

\usepackage{array}
\usepackage{longtable}
\usepackage[tableposition=below]{caption}
\usepackage{caption}

\newtheoremstyle{elementstyle}
  {3pt} 
  {3pt} 
  {\itshape} 
  {} 
  {\bfseries} 
  {.} 
  { } 
  {\thmname{#1}\thmnumber{ #2}\thmnote{ (#3)}} 

\newtheoremstyle{illustrationstyle}
  {3pt} 
  {3pt} 
  {\itshape} 
  {} 
  {\bfseries} 
  {.} 
  { } 
  {\thmname{#1}\thmnumber{ #2}} 

\theoremstyle{elementstyle}
\newtheorem{el}{Element}[section]

\theoremstyle{illustrationstyle}

\makeatletter
\@addtoreset{el}{section}
\@addtoreset{illustration}{section}
\makeatother

\AtBeginDocument{%
  \setcounter{el}{0}
  \setcounter{illustration}{0}
}

\usepackage{siunitx}  
\usepackage{hyperref}
\usepackage{url}
\usepackage{subcaption}
\usepackage{placeins}

\usepackage{environ}
\usepackage{etoolbox}

\usepackage{fontawesome}

\begin{document}

\newif\ifcomments
\commentstrue
\ifcomments
    \providecommand{\narun}[1]{{\protect\color{blue}{[Narun: #1]}}}
    \providecommand{\jesse}[1]{{\protect\color{teal}{[Jesse: #1]}}} 
    \providecommand{\tl}[1]{{\protect\color{magenta}{[Taylor: #1]}}}  
    \providecommand{\klb}[1]{{\protect\color{red}{[Kevin: #1]}}} 
\else
    \providecommand{\narun}[1]{}
    \providecommand{\jesse}[1]{} 
    \providecommand{\tl}[1]{} 
    \providecommand{\klb}[1]{} 
\fi

\newcommand{\firstParent}{\textsc{Foundations}\xspace}
\newcommand{\secondParent}{\textsc{Decisions in Single-Agent Environments}\xspace}
\newcommand{\thirdParent}{\textsc{Decisions in Multi-Agent Environments}\xspace}
\newcommand{\fourthParent}{\textsc{Decisions on Behalf of Other Agents}\xspace}
\newcommand{\fifthParent}{\textsc{Decisions on Consumption in Non-Strategic Environments}\xspace}
\newcommand{\sixthParent}{\textsc{Decisions on Production in Non-Strategic Environments}\xspace}
\newcommand{\seventhParent}{\textsc{Decisions in Multi-Agent Non-Strategic Environments}\xspace}
\newcommand{\eighthParent}{\textsc{Evaluating Equilibria and Externalities}\xspace}

\newcommand{\Parent}{Setting\xspace}
\newcommand{\parent}{setting\xspace}
\newcommand{\Child}{Module\xspace}
\newcommand{\child}{module\xspace}
\newcommand{\parentSec}[1]{\subsection{\textsc{\Parent \arabic{subsection}: #1}}}
\newcommand{\childSec}[1]{\subsubsection{\textsc{\Child \arabic{subsection}.\arabic{subsubsection}: #1}}}
\newcommand{\children}{{\child}s\xspace}
\newcommand{\qtype}{type\xspace}
\newcommand{\qtypes}{{\qtype}s\xspace}
\newcommand{\domain}{domain\xspace}
\newcommand{\domains}{{\domain}s\xspace}
\newcommand{\grade}{grade level\xspace}
\newcommand{\grades}{{\grade}s\xspace}
\newcommand{\tags}{perspective\xspace}
\newcommand{\Tags}{{\tags}s\xspace}

\newcommand{\model}{LLM\xspace}
\newcommand{\models}{LLMs\xspace}
\newcommand{\generator}{LLM\xspace}

\newcommand{\ttname}[1]{{\fontsize{9pt}{10pt}\selectfont{\texttt{#1}}}\xspace}
\newcommand{\steer}{\ttname{STEER}}
\newcommand{\benchmark}{\ttname{STEER-ME}}
\newcommand{\autosteer}{\ttname{auto-STEER}}
\newcommand{\steerbig}{\texttt{STEER}\xspace}
\newcommand{\benchmarkbig}{\texttt{STEER-ME}\xspace}
\newcommand{\autosteerbig}{\texttt{auto-STEER}\xspace}

\newcommand{\src}{\ttname{SRC}}
\newcommand{\srcs}{\ttname{SRCs}}

\newcommand{\reportcard}{\ttname{STEER Report Card}}
\newcommand{\reportcards}{\ttname{STEER Report Cards}}

\newcommand{\hidden}{hidden\xspace}
\newcommand{\shown}{shown\xspace}

\newcommand{\outputtoken}{top token\xspace}

\newcommand{\economicDomain}{non-strategic microeconomics\xspace}


\newcommand{\element}[3]{
        \begin{el}[#1]
          #2
        \end{el}

        \begin{tcolorbox}[colback=black!5!white,colframe=black!5!white,title=Illustration:,coltitle=black,left=5pt,toptitle=1mm,fonttitle=\bfseries]
            #3
        \end{tcolorbox}
}



        \maketitle


\begin{titlepage}
    
\end{titlepage}

\begin{shownto}{iclr}

    \maketitle
    \begin{abstract}
    Large language models (\models) are increasingly being applied to economic tasks like stock picking and financial analysis. Existing \model benchmarks tend to focus on specific applications and often fail to describe a rich variety of economic tasks. \cite{ramansteer} offer a blueprint for comprehensively benchmarking strategic decision-making. However, their work failed to address the non-strategic settings prevalent in micro-economics. We address this gap by taxonomizing micro-economic reasoning into $58$ distinct elements, each grounded in up to $10$ distinct \domains, $5$ \Tags, and $3$ \qtypes. The generation of benchmark data across this combinatorial space is powered by a novel LLM-assisted data generation protocol that we dub \autosteer, which generates a set of questions by adapting handwritten templates to target new \domains and \Tags. By generating fresh questions for each element, \autosteer  helps reduce the risk of data contamination, ensuring that \model evaluations remain valuable over time. We leveraged our benchmark to evaluate $27$ \models over each of the instantiated elements, examined their ability to reason through and solve microeconomic problems and compared \model performance across a suite of adaptations and metrics. Our work provides insights into the current capabilities and limitations of \models in non-strategic economic decision-making and a tool for fine-tuning these models to improve performance. 
    \end{abstract}
\end{shownto}

\section{Introduction}

There is much recent interest in using language models (\models) to reason about economic topics. Some prominent examples include financial sentiment analysis, where \models are tasked with analyzing the sentiment information of financial texts \citep{malo2013gooddebtbaddebt, maia201818, finbert, yang2020finbert}; question answering, where \models are tasked with answering an economic question based on the provided information \citep{maia201818, finqa_benchmark, chen2022convfinqaexploringchainnumerical, flue, xie2023pixiu, ramansteer}; financial text summarization, which entails condensing long unstructured financial texts into short summaries that capture crucial information and maintain factual consistency with the original long texts \citep{mukherjee2022ectsum, zhou2021trade}; and Named Entity Recognition, which asks the model to detect critical financial entities such as persons, organizations, and locations \citep{salinas-alvarado-etal-2015-domain, flue}. More open-ended applications are also starting to emerge. \models such as WallStreetBERT, TradingGPT, FinGPT, FinTral, and BloombergGPT are already giving advice to investors and financial advisors \citep{xie2023wall, li2023tradinggpt, yang2023fingptopensourcefinanciallarge, bhatia2024fintralfamilygpt4level, bloomberggpt}. \models can help to automate budgetary planning and allocation \citep{chen2023emergenceeconomicrationalitygpt}. \models are also being deployed as agents in simulations to analyze the impact of policy changes on key indicators like inflation and GDP growth \citep{carriero2024macroeconomicforecastinglargelanguage, li2024econagentlargelanguagemodelempowered}.

Before \models should be trusted in such open-ended applications, they should demonstrate robustly strong performance on the fundamentals of economic reasoning (just as, e.g., financial advisors, budget planners, and economists are required to do). Many existing benchmarks have been proposed, many of which were introduced in papers cited above. However, most of these are quite narrowly focused on a single task and/or application, rather than assessing economic reasoning more broadly. A second---useful but insufficient---category of benchmarks tests foundational concepts in mathematics, ranging from basic arithmetic to complex problem-solving tasks \citep{huang2016well, ling2017program, amini2019mathqa, lample2019deep, zhao2020ape210k}. Notable benchmarks include GSM8K \citep{cobbe2021training}, a small but varied dataset that contains moderately difficult math problems and MATH \citep{hendrycks2021measuring}, a challenging benchmark for which no evaluated model has yet attained expert-level performance across any of the $57$ tested scenarios.

What might it look like to assess an \model's economic reasoning more comprehensively? Economics encompasses a wide array of problems, such as determining optimal consumption bundles, forecasting profit in the face of uncertainty, or analyzing how a shift in supply impacts equilibrium prices and quantities. Each of these problems can occur in a wide range of contexts such as labor markets, consumer product markets, financial markets, or public policy. Beyond the breadth of inputs that must be considered, evaluating \models presents further challenges to benchmark designers. There is no guarantee that an \model will perform equally well on problems that appear similar or are conceptually related \citep[e.g.,][]{hendrycks2021measuringmassivemultitasklanguage}. For instance, an \model that excels at maximizing profit may struggle with minimizing cost. Similarly, \models can be susceptible to perturbations in the text of a question, which can impact their performance on otherwise similar problems \citep{ribeiro-etal-2020-beyond}. For example, \models may excel in allocating budgets as a doctor, but struggle to allocate budgets as an educator. Finally, \models may reason correctly about their own incentives, but fail to apply this logic to other participants and hence have difficulty understanding market or aggregate level responses (e.g., total supply, demand, and prices). Therefore, in order to be comprehensive, a microeconomic benchmark must exhibit broad variation across problems, contexts, and textual perturbations.
It is similarly nontrivial actually to conduct experiments that comprehensively assesses how well different \models perform at economic reasoning tasks. Different models may leverage distinct architectures, driving performance differences \citep{sanh2020distilbertdistilledversionbert, islam2023comprehensivesurveyapplicationstransformers, ramansteer}. Additionally, adaptation strategies---such as fine-tuning, prompt engineering, and output distribution modification---can dramatically influence a model's effectiveness \citep{gpt3, lester2021powerscaleparameterefficientprompt, kojima2023largelanguagemodelszeroshot}. Under the right adaptations, models with as few as 7B parameters can achieve state-of-the-art performance \cite[e.g.,][]{bhatia2024fintralfamilygpt4level}. Furthermore, robustness across multiple task formats (e.g., multiple-choice QA, free-text QA, etc.) is crucial for understanding the gaps in an \model's reasoning capabilities. A model that performs well on one task format may underperform on others, which suggests gaps in its reasoning processes. Finally, scoring performance using only a single metric can give a skewed understanding of an \model's abilities and limitations \citep{schaeffer2023emergentabilitieslargelanguage}, or obscure tradeoffs that are relevant to practitioners \citep{ethayarajh-jurafsky-2020-utility}. Without a comprehensive evaluation, we risk misattributing performance to a \model when it is instead driven by an adaptation strategy or is an artifact of the metric used. 

A recent paper by \citet{ramansteer} developed a benchmark distribution for assessing economic reasoning in strategic settings that aims for comprehensiveness in the senses just described. This work serves as a starting point for our own paper, and so we describe it in detail. First, they developed a taxonomy that divided the space of game theory and foundational decision theory into $64$ distinct ``elements of economic rationality,'' ensuring that the elements in the benchmark covered a wide range of strategic contexts and decision-making problems. Second, they formalized a hierarchy across elements so that an \model's performance could be better understood in the context of its dependent subtasks. They generated a huge set of questions from this taxonomy, dubbed \steer, which vary in their difficulty and domain (e.g., finance, medicine, public policy). Finally, they evaluated a spectrum of \models over two adaptation strategies and scored with a suite of metrics. They defined this evaluation framework as a \reportcard (\src), a flexible scoring rubric that can be tuned by the user for their particular needs. 

A key drawback of \steer is that, in its focus on game-theoretic reasoning, it neglects much of the subject matter of microeconomics: multiagent settings in which agents nevertheless act nonstrategically. Such reasoning is widespread in competitive markets, where each agent's impact on the market is too small to affect prices unilaterally. For example, while a mobile phone manufacturer might make a strategic decision about the number of handsets to produce and the price to sell them at, a small farm's decision to produce wheat instead of corn given market prices is non-strategic. We employ---and expand upon---the \steer blueprint to construct a benchmark for testing \models on economics in non-strategic environments. Following \citet{ramansteer}, we built a taxonomy of non-strategic economics consisting of $58$ elements. We then instantiated each element in the taxonomy across $8$--$10$ domains. From here, we expanded on the blueprint in two ways. First, we increased the diversity of the questions in the dataset and instantiated each element in $5$ different \emph{\Tags} and up to $3$ \emph{\qtypes} (as defined in \Cref{subsec:dataset}). Second, we expanded their evaluation framework to include newer \models ($27$ in total), some new adaptations ($3$ that we developed and $2$ more from the literature), and many new scoring metrics (a family of $4$ calibration metrics). We dub our benchmark \benchmark, reflecting both its conceptual links to the original \steer and its novel focus on microeconomics.

Even given the best possible \model benchmark, data contamination poses an increasingly important challenge \citep{sainz-etal-2023-nlp, deng2023investigating, ravaut2024much}. Data contamination occurs when the test data used to evaluate an \model is similar or identical to data the \model encountered during training, leading to inflated performance metrics that do not accurately reflect the \model's true capabilities. To tackle this issue, we introduce a new dynamic data generation process called \autosteer which we used to generate all of the questions in \benchmark. \autosteer combines many of the features present in existing dynamic and modular frameworks \citep{gioacchini2024agentquest, wang2024benchmark, white2024livebench} that we detail in \Cref{app:autosteer}. 

In what follows, \Cref{sec:taxonomy} gives an overview of our taxonomy; for space reasons we defer definitions and examples of each element---which are extensive---to \Cref{app:taxonomy}. \Cref{sec:steer_benchmark} describes how we used this taxonomy to build the benchmark distribution. For $37$ elements, we have written {\model} prompts to synthetically generate \num{1000}--\num{5000} multiple-choice questions and manually validated \num{500} generations per element. \Cref{sec:exp_setup} describes the setup of an experiment in which we generated full \srcs for $27$ \models, ranging from Llama-\num{2} \num{7}B to o1-preview, evaluated on a total of \num{21000} test questions. We spent \$\num{12439.54} making requests to OpenAI and Anthropic's API and \num{9.81} GPU years of compute to evaluate open-source models. 
Finally, we survey our experimental results in \Cref{sec:results}. Here, we offer a few highlights. We observed a significant variation in performance across both \models and elements. Even among large models, most underperform on at least a few tasks, indicating that a model's size alone is not sufficient to predict its degree of success across our benchmark. The one exception is o1-preview, which consistently achieved top performance on every element we tested, standing out as the most robust and accurate model in our evaluations. Across \domains and \Tags, \models generally exhibited stable performance, but key error patterns emerged throughout the benchmark. In some instances, \models would ignore the question asked and instead provide ``near-miss'' solutions---those that were economically relevant but addressed a simpler or different problem. In addition, they also frequently bypassed the intended reasoning process by relying on answer choices to reach the correct response rather than deriving solutions independently. Finally, our richest \parent produced a particularly large number of errors: even models as large as GPT-4o and Claude 3.5 Sonnet consistently miscalculated a straightforward concept like deadweight loss by applying incorrect formulas and misinterpreting marginal cost. 

We release all model outputs to support evaluation research and contributions via our website \url{steer-benchmark.cs.ubc.ca}, allowing users to deeply probe all of our experimental results and the underlying model prediction details. Finally, we will release an extensible codebase to support the community in taking \benchmark further.

\begin{table}[ht]
\centering
\setlength{\tabcolsep}{8pt} 
\renewcommand{\arraystretch}{1.2} 
\begin{tabular}{@{} p{9cm} p{4cm} @{}}

\rowcolor{setting1}
\multicolumn{2}{@{} l}{\textbf{Setting 1: Foundations}} \\
\scriptsize
\begin{tabular}[t]{@{}l@{}}
    \textbf{Module 1.1:} Optimization \hyperlink{app:taxonomy}{\faInfoCircle} \\
    \textbf{Module 1.2:} Systems of Equations \hyperlink{app:taxonomy}{\faInfoCircle} \\
    \textbf{Module 1.3:} Derivatives and Homotheticity \hyperlink{app:taxonomy}{\faInfoCircle} \\
\end{tabular}
&
\scriptsize
\begin{tabular}[t]{@{}l@{}}
Number of elements: $6$ \\
Number of questions: $127,342$ \\
Average \# of characters: $134.2$ \\
Number of types: $1$
\end{tabular}
\\
\rowcolor{setting2}
\multicolumn{2}{@{} l}{\textbf{Setting 2: Consumption Decisions in Non-Strategic Environments}} \\
\scriptsize
\begin{tabular}[t]{@{}l@{}}
    \textbf{Module 2.1:} Properties of Utility Functions \hyperref[mod:utility_properties]{\faInfoCircle} \\
    \textbf{Module 2.2:} Deriving Demand \hyperref[mod:deriving_demand]{\faInfoCircle} \\
    \textbf{Module 2.3:} Comparative Statics of Demand \hyperref[mod:properties_demand]{\faInfoCircle} \\
    \textbf{Module 2.4:} Labor Supply \hyperref[mod:labor_supply]{\faInfoCircle} \\
    \textbf{Module 2.5:} Dynamic Consumption Decisions \hyperref[mod:portfolio_choice]{\faInfoCircle} \\
\end{tabular}
&
\scriptsize
\begin{tabular}[t]{@{}l@{}}
Number of elements: $22$ \\
\# of questions: $3,295,770$ \\
Avg. \# chars: $458.35$ \\
Number of \qtypes: $14$
\end{tabular}
\\
\rowcolor{setting3}
\multicolumn{2}{@{} l}{\textbf{Setting 3: Production Decisions in Non-Strategic Environments}} \\
\scriptsize
\begin{tabular}[t]{@{}l@{}}
    \textbf{Module 3.1:} Properties of Production Functions \hyperref[mod:properties_production]{\faInfoCircle} \\
    \textbf{Module 3.2:} Deriving Factor Demand \hyperref[mod:deriving_factor_demand]{\faInfoCircle} \\
    \textbf{Module 3.3:} Comparative Statics with Production \hyperref[mod:production_statics]{\faInfoCircle} \\
    \textbf{Module 3.4:} Dynamic Production Decisions \hyperref[mod:dynamic_production]{\faInfoCircle} \\
\end{tabular}
&
\scriptsize
\begin{tabular}[t]{@{}l@{}}
Number of elements: $16$ \\
\# of questions: $1,333,330$ \\
Avg. \# chars: $434.48$ \\
Number of types: $20$
\end{tabular}
\\
\rowcolor{setting4}
\multicolumn{2}{@{} l}{\textbf{Setting 4: Non-Strategic Decisions in Multi-Agent Environments}} \\
\scriptsize
\begin{tabular}[t]{@{}l@{}}
    \textbf{Module 4.1:} Consumer Goods Market Aggregation \hyperref[mod:consumer_aggregation]{\faInfoCircle}\\
    \textbf{Module 4.2:} Factor Market Aggregation \hyperref[mod:factor_aggregation]{\faInfoCircle}\\
    \textbf{Module 4.3:} Prices in Static Market Equilibrium \hyperref[mod:static_equilibrium]{\faInfoCircle}\\
    \textbf{Module 4.4:} Comparative Statics of Equilibrium Prices \hyperref[mod:comparative_equilibrium]{\faInfoCircle} \\
\end{tabular}
&
\scriptsize 
\begin{tabular}[t]{@{}l@{}}
Number of elements: $10$ \\
\# of questions: $750,060$ \\
Avg. \# chars: $362.69$ \\
Number of types: $6$
\end{tabular}
\\
\rowcolor{setting5}
\multicolumn{2}{@{} l}{\textbf{Setting 5: Evaluating Equilibria and Externalities}} \\
\scriptsize
\begin{tabular}[t]{@{}l@{}}
    \textbf{Module 5.1:} Welfare and Decentralization \hyperref[mod:welfare]{\faInfoCircle} \\
    \textbf{Module 5.2:} Welfare Analysis of Market Equilibrium \hyperref[mod:analysis_equilibrium]{\faInfoCircle} \\
\end{tabular}
&
\scriptsize
\begin{tabular}[t]{@{}l@{}}
Number of elements: $10$ \\ 
\# of questions: $698,367$ \\
Avg. \# chars: $311.50$ \\
Number of types: $5$
\end{tabular}
\\

\end{tabular}
\caption{High-level diagram of the taxonomy of elements of rationality. At the top level, we divide the space of decision making into 5 {\parent}s; we further subdivide {\parent}s into \children (e.g., Comparative Statics of Demand) that capture conceptually similar behaviors. We also include a few summary statistics about the dataset. Each \faInfoCircle~ icon is a hyperlink to the corresponding module in our appendix.}
\label{tab:taxonomy}
\end{table}

\section{Elements of Economic Rationality}\label{sec:taxonomy}

Our first step in generating a benchmark for \economicDomain is to taxonomize this space. Previous work by \citet{ramansteer} developed a taxonomy for economic rationality within strategic domains. Their approach involved identifying foundational principles that define how agents should make decisions in specific environments and then organizing these principles, or ``elements,'' into progressively more complex decision-making scenarios. We adopt a similar hierarchical approach for \benchmark, focusing on organizing economic decision-making principles into structured categories. However, unlike \steer, which assesses decision-making in strategic environments, our focus is assessing how agents make decisions given prices and quantities that are determined by the forces of supply and demand. We call this sub-field non-strategic microeconomics.

Two of the settings from \steer remain directly relevant to \economicDomain: \firstParent and \secondParent. As we describe our taxonomy, we begin with these foundational settings. The elements we incorporate from \firstParent---arithmetic, optimization, probability, and logic---are core mathematical skills essential for microeconomic reasoning and are already present in \steer. In \benchmark, we expand this setting by adding elements that test basic calculus, such as single-variable derivatives and linear systems of equations. In \steer, \secondParent focused on testing whether an agent can adhere to the von Neumann-Morgenstern utility axioms when making decisions over a set of alternative choices. We include those axiomatic elements and extend this \parent to include testing the properties of commonly used parameterizations of utility functions in non-strategic microeconomic contexts, such as utility functions with satiation points, monotone preferences, and budget constraints.

Building directly on these foundational settings, we introduce the next \parent, \fifthParent, which tests an agent's ability to optimally exchange time and money for desired goods and services. Elements in this \parent assume that the agent is a price taker, meaning that the agent accepts market prices as given rather than forecasting how a purchase
might move the market. First, we test the agent's ability to derive demand functions consistent with the axioms and functional forms from \secondParent. These foundational elements are useful in assessing whether an agent can make consistent, rational choices in response to market prices. We then include elements testing the agent's ability to determine optimal consumption bundles, decide when to leave the workforce, and conduct comparative statics with demand functions.

\sixthParent tests an agent's ability to decide on the combination of inputs to efficiently produce goods and services to maximize profits. The \parent starts by assessing the agent's ability to identify and analyze basic properties of production functions, such as the relationship between input quantities and output levels. This includes concepts like returns to scale, diminishing marginal returns, and the technological constraints that shape production capabilities. We then test the agent's ability to conduct expenditure minimization and its dual, profit maximization. This involves solving optimization problems where the agent must use marginal analysis to determine the quantity of output that maximizes profit (i.e., minimizes cost).

\seventhParent considers consumers and producers who each reason according to the principles just described to trade with each other. This more complex setting requires an agent to reason about how the aggregated behaviors of consumers and producers lead to market-clearing prices that balance supply and demand. This \parent covers elements such as finding market-clearing prices, computing competitive equilibria, and analyzing the comparative statics of equilibrium in markets where individual actions do not directly impact others.

Our last setting, \eighthParent, tests agents on their ability to evaluate whether equilibria are efficient and to analyze the effects of interventions, such as taxes or price ceilings, on welfare. In this \parent, agents must not only be able to analyze how supply and demand dynamics establish equilibrium prices but also consider how external interventions shift these dynamics and alter the behavior of both consumers and producers. The elements in this \parent can be relatively simple (e.g., compute consumer/producer surplus) or involve detailed counterfactual analysis (e.g., predict how interventions impact prices, the allocation of resources, and welfare outcomes). 


\begin{shownto}{iclr}
    For a more detailed discussion on the structure of these elements and the methodology we used to group the elements, including formal definitions, we refer the reader to \Cref{app:taxonomy}.
\end{shownto}

\section{The \benchmarkbig Benchmark} \label{sec:steer_benchmark}
We first give an overview of \benchmark dataset and then explain the process we used to generate and validate these questions, which we call \autosteer. Finally, we describe our evaluation framework. 

\subsection{Dataset} \label{subsec:dataset}
We adopted the widely used Multiple-Choice Question Answering (MCQA) format for our benchmark \citep[see, e.g.,][] {rajpurkar2016squad, wang2018glue, wang2019superglue, hellaswag, mmlu, flue, liang2022holistic,big_bench_hard}. In this format, each test question presents a decision-making scenario along with several candidate options, where only one is correct. As an evaluation paradigm, a key benefit of MCQA is that it provides a standardized way to evaluate an \model's ability to correctly respond to given prompts. MCQA tasks admit well-established metrics like exact-match accuracy or expected calibrated error that provide interpretable measures of how well an \model answers questions \citep{liang2022holistic, li-etal-2024-multiple}. Furthermore, many real-world applications of \models in economics involve answering questions: e.g., chatbots \citep{inserte2024large} and virtual assistants \citep[BloombergGPT][]{wu2023bloomberggptlargelanguagemodel}.

Our own benchmark consists of a total of $30$ instantiated elements, each containing 5,000--20,000 MCQA questions. Each question is characterized by a (\qtype, \domain, \tags) tuple. Different \textit{\qtypes} represent distinct ways of testing an agent's abilities within an element. For example, we could assess an agent's ability to perform profit maximization by asking ``What is the maximum profit?'' or ``How much labor is needed to maximize profit?''  The \textit{\domain} of a question indicates which of $10$ predefined topic areas it pertains to: consumer goods, medical, finance, education, technology, entertainment, environmental policy, politics, sports, or gambling. Finally, the \textit{\tags} of a question represents which of the $5$ predefined \tags the question was written in: first-person, second-person, third-person anonymous, third-person female and third-person male. We disallow over (\qtype, \domain, \tags) combinations that do not lead to coherent questions; for example, questions about welfare theorems do not make sense in gambling settings.

\subsection{\autosteerbig}
Like \citet{ramansteer}, we leveraged a state-of-the-art \generator to help generate our dataset. We substantially extended their methodology, however, by adding an additional style-transfer step where we asked the \generator to rewrite questions in new domains or perspectives. This greatly increased the variety of questions we were able to add. This section describes how we used our new approach to design \benchmark. 

\begin{figure}
    \centering
    \includegraphics[trim={0.25cm 0.2cm 0 0},clip,width=0.65\textwidth]{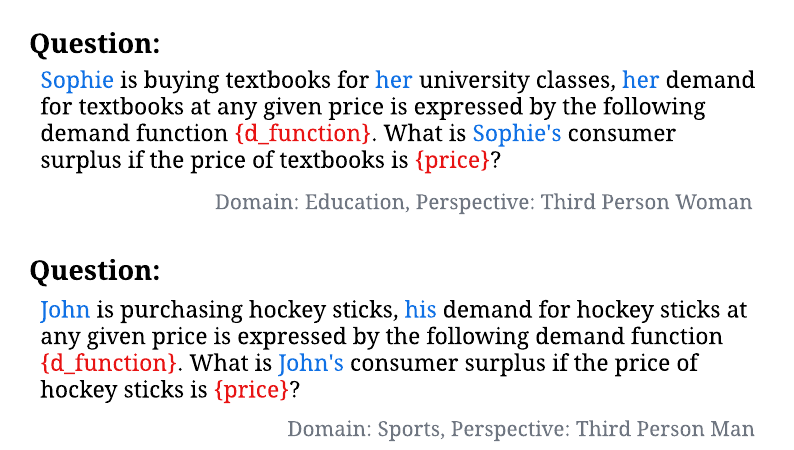}
    \caption{This figure depicts two questions in the consumer surplus element with different domains and perspectives. The text colored in red are the labeled fields that will be filled for test time and the text in blue is the perspective. On top, a question is framed in the education domain from a third-person woman perspective, while on the bottom, the same question is written for the sports domain from a third person man perspective. These were both generated during the style-transfer step in the data generation process.}
    \begin{shownto}{iclr}
        \vspace{-30pt} 
    \end{shownto}
    \begin{shownto}{arxiv}
        \Description[]{This figure depicts two questions in the consumer surplus element with different domains and perspectives. The text colored in red are the labeled fields that will be filled for test time and the text in blue is the perspective. On top, a question is framed in the education domain from a third-person woman perspective, while on the bottom, the same question is written for the sports domain from a third person man perspective. These were both generated during the style-transfer step in the data generation process.}
    \end{shownto}
    \label{fig:domain_perspective}
\end{figure}

First, for each \qtype we hand-wrote a set of gold-standard example templates that served as the seeds for the data generating process. As can be seen in \Cref{fig:template_writing}, these templates were tagged with a \domain, a \tags, and a \qtype, if appropriate. The majority of these questions had \emph{labeled fields} for numbers (e.g., ``\dots the cost of labor is \{cost\}\dots'') which were programmatically filled for test time. See \Cref{fig:domain_perspective} for an example.

Next, we asked the \generator to style-transfer these templates into each of the domains. We primarily leveraged gpt-4o to generate our benchmark, but as we show in \Cref{app:claude_elements}, using claude-3-5-sonnet did not change \model performance. Our prompt included explicit instructions to maintain the same set of labeled fields as the hand-written templates. \Cref{fig:template_generation} depicts the style-transfer page in our web application along with the prompting instructions. \models can be inconsistent in maintaining the economic meaning of questions after \domain style transfer, so we hand-checked each of the outputted templates and edited them when necessary. All of these operations are supported by a web application we built: see \Cref{fig:template_validation} in \Cref{appendix:web}. We then further style-transferred each of these newly generated templates into each perspective, resulting in up to $40$ unique domain-perspective pairs for each \qtype. 
We ran an additional check on the style-transfer process by filling the labeled fields in the templates with values and asking the \generator to solve the questions as written, which we found could highlight mistakes in question wording or in programmatically filled values; see \Cref{fig:template_ai_validation}.
\begin{shownto}{iclr}
    We ran an additional check on the style-transfer process by filling the labeled fields in the templates with values and asking the \generator to solve the questions as written, which we found could highlight mistakes in question wording or in programmatically filled values; see \Cref{fig:template_ai_validation} in \Cref{appendix:web}.
\end{shownto}
(We were careful only to use his procedure to correct mistakes in the templates, not to tune the difficulty of the questions in a way that would bias our benchmark.)

\begin{figure}[h]
    \centering
    \includegraphics[trim={0 0cm 0 1.6cm},clip,width=\linewidth]{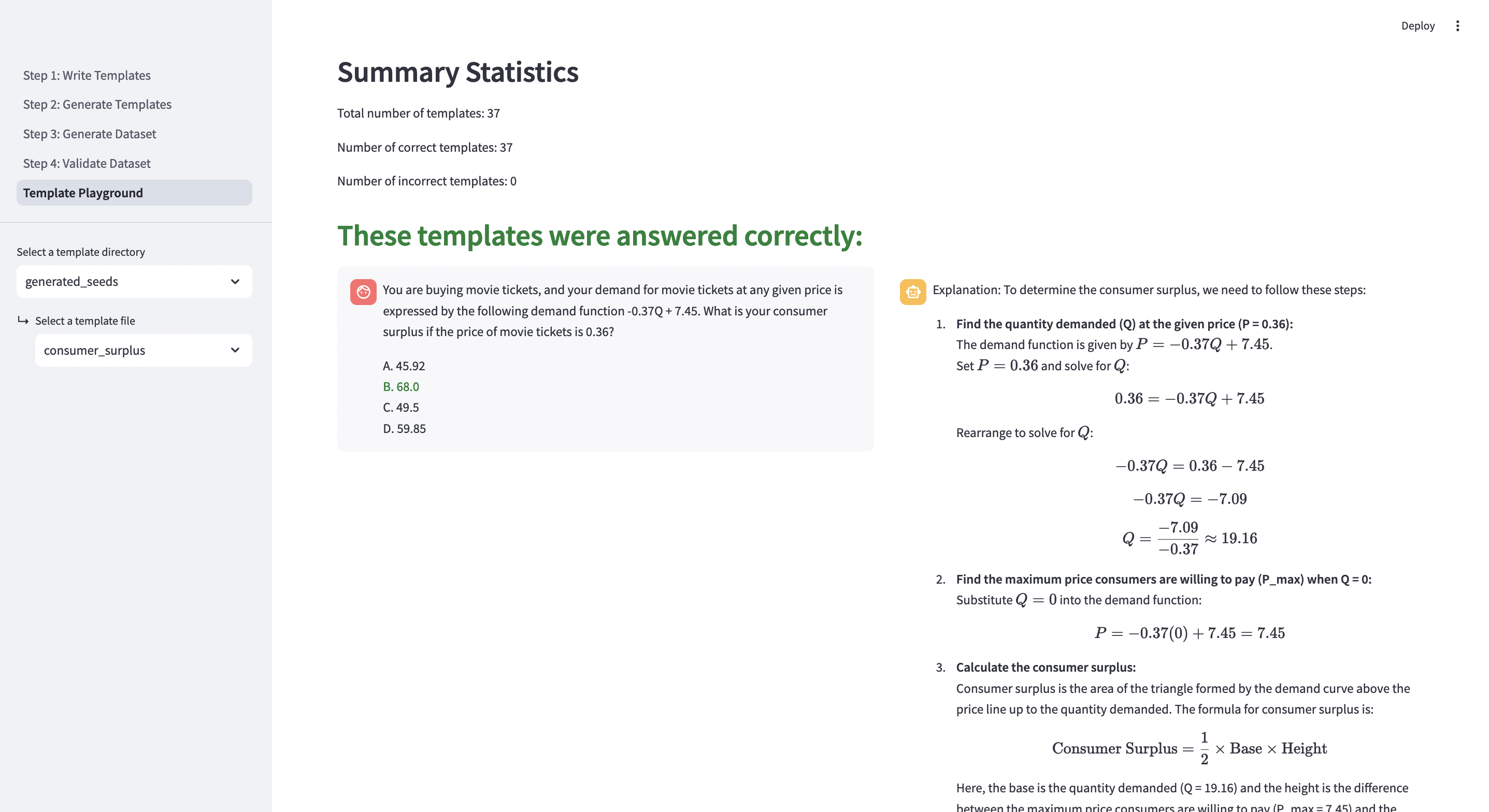}
    \caption{The web app user interface for template AI double-checking. This page instantiates and fills a set of question using a generated or example seed and then generates a response using an OpenAI model. The page also reports the number of questions answered correctly as well as the responses from the model.}
    \label{fig:template_ai_validation}
    \begin{shownto}{arxiv}
        \Description[]{The web app user interface for template AI double-checking. This page instantiates and fills a set of question using a generated or example seed and then generates a response using an OpenAI model. The page also reports the number of questions answered correctly as well as the responses from the model.}
    \end{shownto}
\end{figure}

We then took each of these templates and asked the \generator to replicate the template, keeping the domain, perspective and labeled fields fixed but modifying exact words or objects used in the question. We generated $100$ new templates for each element, crossing every \domain and \tags pair, resulting in \num{30000} templates across the dataset. We then spot-checked $500$ of the resulting templates for each element, and flagged $99.88$\% of the templates as valid.

Finally, we created 20 instantiated questions from each template by filling its labeled fields with randomly generated values. We restricted the random generator to output numbers that were appropriate given the context: e.g., demand functions had negative slopes, positive values for equilibrium prices, etc. We programmatically solved each question and filled in the appropriate options and answer. In the end, we produced \num{1000} questions per (\domain, \tags) pair and up to \num{40000} per \qtype.

\subsection{Evaluation Framework}

We now turn to describing our evaluation framework. Following other work in this space, we consider an \model as a black box to which we provide inputs in the form of prompts (i.e., strings) and adjust the decoding parameters (e.g., temperature) to analyze the resulting output completions (i.e., strings) and log probabilities, when available. Within this black-box framework, we consider two classes of adaptations: performance adaptations, which modify inputs to affect performance on a task, and diagnostic adaptations, which aim to analyze specific behaviors or model characteristics. We then score \models across a suite of metrics.

We follow \citet{ramansteer} by allowing a user to tune the evaluation framework for their specific needs by choosing for their set of \models: the set of elements in the evaluation, the adaptation chosen for each \model and a scoring metric. For instance, one may only want to evaluate specific economic \children in our taxonomy (e.g., utility maximization for individual decision-making in \fifthParent or production optimization scenarios in \sixthParent), or conduct comparative assessments across adaptation strategies, or evaluate targeted use cases like medical or financial decision-making. We provide a number of predefined evaluation frameworks in our web application as well as allowing users to create new evaluation frameworks.

We classify any adaptation as a performance adaptation when the inputs are modified in a way that is intended to increase an \model's performance on a task. Common performance adaptations are chain-of-thought reasoning \citep{CoT, meta_cot, self_improve, kojima2023largelanguagemodelszeroshot} and few-shot prompting \citep{gpt3, perez2021true}. We focus on zero-shot chain-of-thought reasoning.

\textbf{Zero-Shot Chain-of-Thought (0-CoT).}
There has been work showing that performance can be improved by asking an \model to explain its reasoning before outputting an answer \citep{CoT, meta_cot, self_improve, kojima2023largelanguagemodelszeroshot}. We follow \citet{kojima2023largelanguagemodelszeroshot} in implementing 0-CoT by first asking the \model to explain its reasoning and then subsequently asking it to select the correct answer. We take two approaches to adapting 0-CoT to MCQA, which we denote \emph{\hidden} and \emph{\shown}. In the \hidden approach, we give the \model the question text and ask it to explain its reasoning---we only provide the candidate options in the second step. In the \shown approach, the \model is given both the question text and candidate options when it is asked to explain its reasoning. See \Cref{fig:hidden_vs_shown} in \Cref{appendix:sec3} for an example.

\subsubsection{Diagnostic Adaptations}\label{subsubsec:diagnostic}
\emph{Diagnostic adaptations} alter the prompt or decoding parameters not to improve performance, but rather to gain a better understanding of an \model's behavior. 

\textbf{Calibrated Answer Replacement (CAR).} In CAR, we modify the candidate options by replacing one of the options with the  string ``No other option is correct.'' For a test containing questions with $n$ options, we replace the correct answer with this placeholder in a $1/n$ fraction of questions. For the remaining questions, we replace one of the incorrect answers instead. This ensures that an \model that always chooses ``No other option is correct'' achieves the same accuracy as random guessing.

\textbf{Reshaped Probability Mapping (RPM).} Sometimes, \models can assign nonzero probability to tokens that do not correspond to any of the options available. Such errors are trivial to fix in any downstream application. However, if not corrected for, such errors can distort performance metrics, e.g., leading models to appear to perform worse than random guessing. We call the adaptation that addresses this issue RPM and take two approaches to reshaping the outputs. The first approach is renormalizing the output distribution to distribute all probability mass to valid options. However, in cases where the model puts very little weight on \emph{any} correct option this renormalization can make the model appear overconfident. Our second approach attempts to deal with this by mixing the output distribution with a uniform distribution over valid options. This means if very little probability mass is given to any correct option its output will look more uniform and hence less confident in its answer. We define these adaptations and offer further discussion in \Cref{app:adaptations}. Importantly, neither implementation changes which of the valid option tokens receives the largest weight in the output distribution, and therefore assessments of the \model's accuracy. 

\textbf{Free-Text QA.} In addition to the diagnostic adaptations discussed earlier, we conducted experiments involving free-text generation question answering to more closely align with real-world use cases. 
We ask an evaluator \model to report the answer the chain of thought reasoning arrived at and None if there is no easily findable answer. We then scored a model's answer as correct if it was within 98\% of the correct answer value and is closer to the correct answer than any other option. We include the prompt we used in \Cref{app:free_text_prompt}.

\subsubsection{Scoring}\label{subsubsec:scoring}

Given a complete set of model responses, it is far from straightforward to choose a way of computing a single, overall performance score. Consequently, benchmarks often employ a suite of metrics to provide a more comprehensive assessment of performance \citep{wang2019superglue, gehrmann-etal-2021-gem, liang2022holistic, srivastava2023imitationgamequantifyingextrapolating}. We evaluate \models using three categories of metrics: accuracy, calibration, and robustness. We leave the discussion and definitions of our scoring metrics in \Cref{app:metrics} and simply list the metrics below:

\begin{itemize}
    \item Accuracy: \nameref{metric:exact_match} and \nameref{metric:normalized_accuracy}
    \item Calibration: \nameref{metric:ece}, \nameref{metric:brier_score}, and \nameref{metric:epa}
    \item Robustness: \hyperref[app:robustness]{Domain Robustness} and \hyperref[app:robustness]{Type Robustness}.
\end{itemize}

In this paper we score \models on their restricted output distributions over valid option tokens, modified using the diagnostic adaptation RPM as described in \Cref{subsubsec:diagnostic}. For each model, we also report the proportion of responses where the \outputtoken is not a valid option token.

A \model's score on an element is the average taken over all questions in an element. We consider an element a base concept in our benchmark and therefore define the accuracy and confidence metrics with respect to an element.

\section{Experimental Setup}\label{sec:exp_setup}
\Cref{tbl:models} in \Cref{appendix:models} lists the \num{27} \models we evaluated. We ran gpt-$4$o, gpt-$4$o-mini, and o$1$-preview using OpenAI's API \citep{openai_api}; claude-$3$-$5$-sonnet and claude-$3$-haiku using Anthropic's API \citep{anthropic-api}. We obtained $22$ open-source \models from the HuggingFace Hub \citep{huggingface} and ran them on between $1$ and $4$ A$100$, Tesla M$60$, and V$100$ GPUs (depending on model size) on one of several dedicated compute clusters to which we have access.

In multiple-choice classification, there are a few ways one might represent the input to an \model. We follow prior work by \citet{hendrycks2020measuring} who introduced the \emph{joint} approach where all answer choices are combined with the question into a single prompt, and the \model predicts the most likely option letter.\footnote{There is another approach, called \emph{separate} and employed by \citet{gpt3} However, this approach is better suited to tasks where the answer choices are long-form generations.} We then decoded valid multiple choice responses from all \models as described in \Cref{subsubsec:scoring}. For those \models where we had no access to the output distribution (claude-3-5-sonnet, claude-3-haiku, o1-preview) we took the \outputtoken.\footnote{OpenAI models only return the top $20$ tokens, however, we never saw a valid option token not present in those top $20$ tokens.} In the free-text QA adaptation, we used gpt-4o-mini as the evaluator \model due to its low cost and high performance in text retrieval. 

Due to time and budget constraints we evaluated the closed-source \models, claude-3-5-sonnet, claude-3-haiku, gpt-4o, and gpt-4o-mini, on all \num{35} of the instantiated elements, all open-source models on $20$ of the instantiated elements, and o1-preview on $13$ elements. We applied our benchmark across all combinations of adaptations and \models, except for in the case of o$1$-preview. We did not explicitly ask o1-preview to conduct $0$-CoT reasoning since the model was already explicitly trained to perform reasoning; thus, we simply asked o1-preview for the \outputtoken. Additionally, since o1-preview always performs reasoning we do not run the \hidden implementation as its impossible to keep it from doing an additional reasoning step once it sees the candidate answers.
This led to a total of $4$ experiments per element for o1-preview and $8$ for all other \models.

\section{Results}\label{sec:results}

\begin{shownto}{iclr}
    \begin{figure}
        \centering
        \includegraphics[width=\linewidth]{images/heatmap_new.png}
        \caption{This figure plots a heatmap of the closed-source \model performance measured with normalized accuracy on the $30$ elements we instantiated. The \models, on the y-axis, are sorted in terms of parameter size. The elements, on the x-axis, are grouped by \parent.}
        \label{fig:element_heatmap}
    \end{figure}
\end{shownto}

\begin{shownto}{iclr}
    \Cref{fig:element_heatmap} depicts aggregate performance across our whole benchmark, using normalized accuracy with the \shown implementation of $0$-CoT and without CAR. We chose these adaptations as we observed that \models performed the best on that adaptation configuration on average. We plot the models in descending order of parameter size and the elements in taxonomical order (i.e., \firstParent elements first) breaking ties alphabetically. Due to space constraints we only include \models that performed sufficiently better than random guessing: with normalized accuracy greater than $0.2$ on average (see \Cref{fig:small_model_heatmap} in the appendix for the remaining models). Furthermore, we observed that for the \models that we plot, our calibration metrics were correlated with normalized accuracy and hereafter focus mainly on normalized accuracy.
    
    All \models strugged with a least some elements in each of the settings in our benchmark (i.e., from \firstParent to \eighthParent); however, on the $13$ elements that we tested, o1-preview was the most accurate model (see the top row in \Cref{fig:element_heatmap}). Even in elements where every other model was close to random guessing (e.g., \nameref{el:profit_max} and \nameref{el:dynamic_profit_max}) o1-preview obtained high accuracy. With the exception of o1-preview, no single \model consistently surpassed all others across our entire benchmark. 

    A common struggle for \models was the precision required to solve optimization problems, particularly those that involve multiple sequential steps of computation and economic interpretation. For instance, in a challenging task like \nameref{el:dynamic_profit_max}, \models are tasked with solving a $2$-stage optimization problem that requires accurately performing a series of interdependent calculations. Each step, from identifying the correct approach to interpreting the economic implications and executing precise computations, presents opportunities for errors to accumulate.
    
    However, even elements with simple mathematical problems presented opportunities for errors. None of the closed-source \models, except for o1-preview were able to consistently compute the \nameref{el:deadweight_loss}; an element whose primary mathematical requirement is computing the area of a triangle. We discovered that models like claude-3-5-sonnet and gpt-4o often used an incorrect formula for computing deadweight loss and made errors in interpreting the marginal cost, a crucial step in the problem-solving process. To better understand these errors we investigated model performance in the free-text QA adaptation. \Cref{fig:sonnet_dwl} and \Cref{fig:4o_dwl} show the distribution of correct responses and specific errors for claude-3-5-sonnet and gpt-4o, respectively. While gpt-4o displayed performance better than random guessing, errors stemming from the use of an incorrect formula consisted of the majority of responses. claude-3-5-sonnet, on the other hand, exhibited a higher prevalence of incorrect formula errors, with nearly 44\% of its responses relying on a particular incorrect formula for deadweight loss. Furthermore, gpt-4o was more susceptible to compounding issues, incorrectly computing marginal cost and using an incorrect deadweight loss formula, than claude-3-5-sonnet. We describe these errors in more detail in \Cref{app:deadweight_loss}.

    \subsection{Robustness}
    
    \emph{Domain Robustness.} While overall the variation across domains was limited, we observed noticeable differences in specific elements. In particular,  elements testing conceptual understanding of foundational principles (e.g., first welfare theorem) showed that certain domains provided more effective contextual cues for the \models. For example, in the consumer goods domain—where items like apples, chairs, or mugs are familiar in economic word problems—\models were more likely to recognize the task as an economic problem and anchor their reasoning in classical economic principles. 
    
    In contrast, the technology \domain, where the economic context could be interpreted as a real-world scenario presented more challenges. The \models often failed to recognize what was being asked and equivocated when reasoning about the problem. The largest performance gaps appeared in the \nameref{el:welfare_theorem_1} and \nameref{el:welfare_theorem_2} elements. For instance, claude-3-5-sonnet exhibited a gap of $0.657$ in accuracy between the consumer goods and technology domains, claude-3-haiku had a gap of $0.48$, and gpt-4o-mini showed a gap of $0.278$. 
    
    \emph{Type Robustness.} Here, we examine \model performance across different families of functions used in economic reasoning. These include Cobb-Douglas, Leontief, linear, and non-linear functions. Each family of functions poses distinct challenges depending on the mathematical operations and economic concepts being tested. While Cobb-Douglas functions are ubiquitous in economics, they can often be more challenging for language models as they feature non-integral exponents, which add a layer of difficulty in operations like differentiation. For instance, in \Cref{fig:type_robustness}, we observe that, with the exception of claude-3-haiku, performance on non-linear functions (polynomials with integer exponents of degree $\leq$ 3) surpasses performance on Cobb-Douglas functions.
    
    For any given element, the family of functions that is the most difficult can vary. For example, computing the \nameref{el:returns_to_scale} of a Cobb-Douglas production function is the sum of the exponents and computing the \nameref{el:output_elasticity} corresponds to the exponent on the input.

    \subsection{Adaptations}
    
    We observed that in the \hidden implementation, \model performance was worse overall compared to the \shown implementation. This suggests that \models benefit from being able to reason directly over the options. 
    
    One pattern we observed was models exploiting the provided options to ``cheat'' the question. Instead of deriving the answer from first principles, \models would insert the candidate options directly into functions in the question text and select the correct answer based on which option produced the best result. This strategy was particularly prevalent in the \nameref{el:profit_max} element, where models were asked to find the amount of labor to employ that maximizes a profit function. While the intended approach was for the model to take the derivative of the profit function and identify the profit-maximizing labor, \models often bypassed this by simply plugging in each of the given options and selecting the one that resulted in the highest profit. We observed this behavior in every question that we spot-checked where gpt-$4$o answered correctly (see \Cref{app:profit_max_example}). 
    
    We also found that the inclusion of options could signal how to reason about the question. This was particularly prevalent in the aggregation elements in \eighthParent and especially in the \nameref{el:agg_consumer_demand} element, which ask models to aggregate the quantity demanded for some number of consumers. In the \hidden implementation, models often failed to multiply the quantity demanded by the number of consumers in the market. When presented with the options, the additional signal in the magnitude of each of the candidate options increased performance. Providing evidence of this, we found that as the number of digits in the answer increased so too did the exact-match accuracy. \Cref{fig:agg_consumer_demand} (in the appendix) shows that as the number of digits in the answer increased, so too did the exact-match accuracy, providing evidence that models use the magnitude as a hint for reasoning. We show an example of this behavior in \Cref{app:agg_consumer_demand_example}.

    To further investigate this effect, we examined four elements (\nameref{el:intertemporal_consumption_smoothing}, \nameref{el:profit_max}, \nameref{el:agg_consumer_demand}, and \nameref{el:producer_surplus}) that exhibited the largest gap in accuracy between hidden and shown adaptations. Our analysis revealed that performance was almost always worse under the free-text QA adaptation compared to the hidden adaptation, see \Cref{fig:hidden_vs_no_mc}. This performance gap appears to stem from the models' tendency to selecting the closest option to the free-text answer. \Cref{fig:percent_closest} shows the percentage of times that models were correct under the hidden adaptation but incorrect under the free-text adaptation due to guessing the closest answer. In almost all cases, the majority of the gap is due to this phenomenon. We offer more discussion in \Cref{app:free_vs_hidden}.
\end{shownto}

\begin{shownto}{arxiv}
    All \models strugged with a least some elements in each of the settings in our benchmark (i.e., from \firstParent to \eighthParent); however, on the $13$ elements that we tested, o1-preview was the most accurate model.
    Even in elements where every other model was close to random guessing (e.g., \nameref{el:profit_max} and \nameref{el:dynamic_profit_max}) o1-preview obtained high accuracy. Besides o$1$-preview, no \model consistently outperformed other \models across our benchmark. 

    For the remainder of this section, we organizing our discussion by economic setting, identifying key trends, successes, and failures in model performance. For each non-\firstParent \parent, we plot a heatmap of the performances of \models on the \shown adaptation and without CAR.  We selected these adaptations because we saw that LLMs performed best with that configuration.
    We plot the models in descending order of parameter size, breaking ties alphabetically, and aggregate element performance across all \domains, \qtypes, \tags. 
    We observed that for the \models that we plot, our calibration metrics were correlated with normalized accuracy; thus, our exposition in what follows focuses on normalized accuracy.

    \subsection{\firstParent}
    
    \begin{figure}[H]
        \centering
        \includegraphics[width=0.95\linewidth]{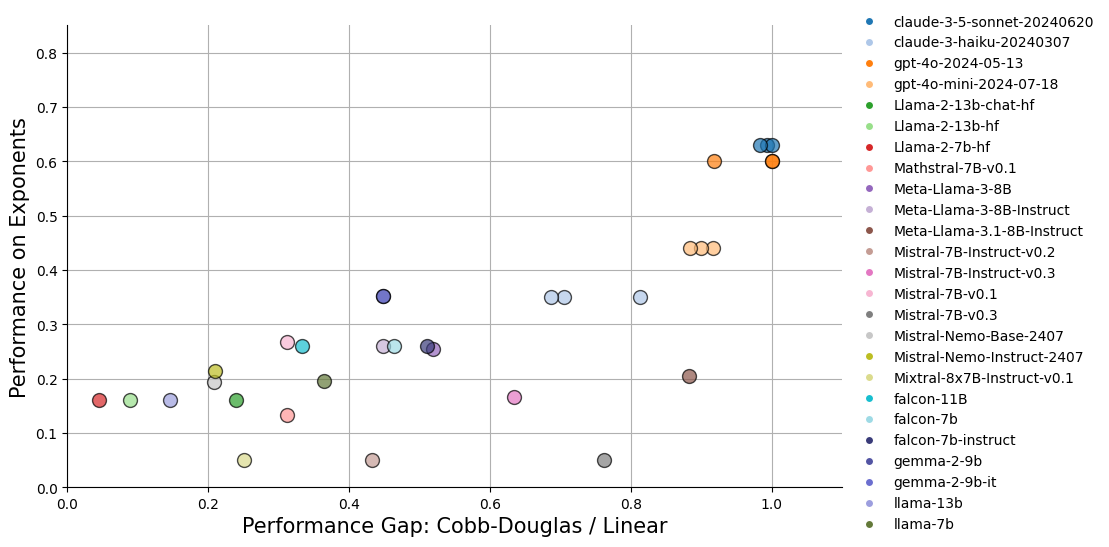}
        \caption{Scatter plot of calibrated performance on the Exponents element versus downstream performance gap across models. The x-axis shows the gap calculated as the quotient between a model's accuracy on real-valued exponent‐based (Cobb–Douglas) tasks and its accuracy on the linear version of those tasks for various downstream elements. The y-axis represents the model's performance on Exponents normalized by dividing by its average accuracy on the benchmark. Each point corresponds to a specific (model, downstream element) pair, with colors distinguishing different models.}
        \begin{shownto}{arxiv}
            \Description[]{Scatter plot of calibrated performance on the Exponents element versus downstream performance gap across models. The x-axis shows the gap calculated as the quotient between a model's accuracy on real-valued exponent‐based (Cobb–Douglas) tasks and its accuracy on the linear version of those tasks for various downstream elements. The y-axis represents the model's performance on Exponents normalized by dividing by its average accuracy on the benchmark. Each point corresponds to a specific (model, downstream element) pair, with colors distinguishing different models.}
        \end{shownto}
        \label{fig:foundations_ratio}
    \end{figure}

    A persistent challenge for many economic reasoning tasks is the precision required in multi-step optimization problems. These tasks not only demand accurate sequential calculations but also require the proper economic interpretation of intermediate results. In such settings, even minor errors in foundational computations---like those involving exponent manipulation---can compound and significantly undermine the final outcome. Recognizing this challenge, we investigated whether models that excel at basic exponent operations performed better on downstream tasks that rely on these operations.
    
    We first calibrated model performance on the Exponents element by normalizing it relative to the average accuracy across the benchmark. We then computed, for each downstream element, a performance gap defined as the difference between accuracy on tasks that involve exponentiation (those instantiated with Cobb–Douglas functions) and their linear counterparts that do not require exponent manipulation. Each (model, downstream element) pair is represented as a point in Figure \ref{fig:foundations_ratio}, where the x-axis shows the performance gap and the y-axis displays the performance on the Exponents element.

    The results reveal a clear trend: \models with higher performance on the Exponents element tended to exhibit a significantly reduced gap between their performance on exponent-based tasks and linear tasks. This relationship is statistically significant, with a Pearson correlation coefficient of \num{0.774} ($p = \num{1.23e-07}$), indicating that foundational mathematical skills accounted for a significant portion of the variance in downstream task performance.

    \subsection{\fifthParent}

    \begin{figure}[H]
        \centering
        \includegraphics[width=\linewidth]{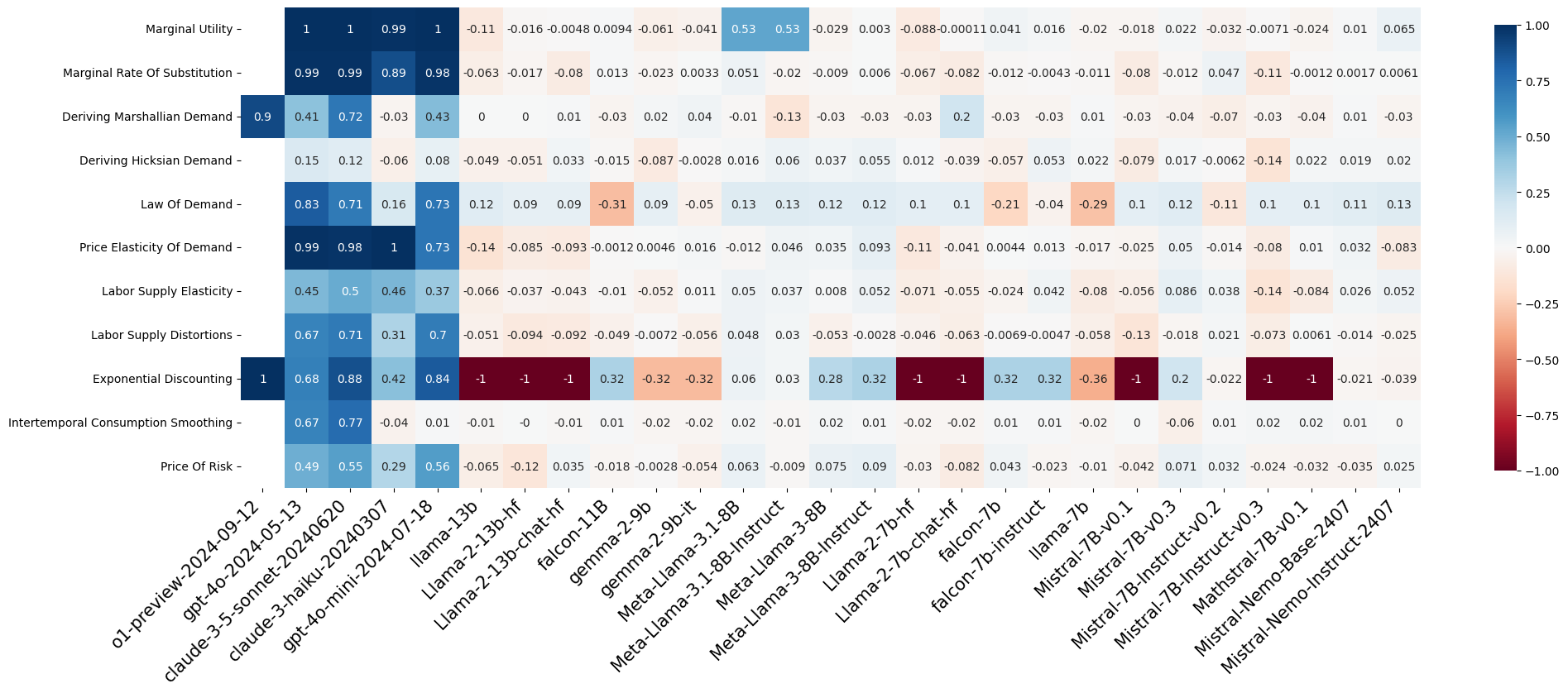}
        \caption{Heatmap plotting normalized accuracy performance of \models on elements within the \fifthParent \parent. Performance is on the \shown adaptation without CAR and we sort \models by parameter size (when available).}
        \begin{shownto}{arxiv}
            \Description[]{Heatmap plotting normalized accuracy performance of \models on elements within the \fifthParent \parent. Performance is on the \shown adaptation without CAR and we sort \models by parameter size (when available).}
        \end{shownto}
        \label{fig:consumers_heatmap}
    \end{figure}

    A key takeaway from \Cref{fig:consumers_heatmap} is that the closed‐source \models achieved relatively high average normalized accuracy on elements that assessed the properties of demand and labor supply functions, indicating that these models could generally interpret and compute straightforward statistics. However, their performance dropped considerably on more elements focusing on computational problems such as \nameref{el:intertemporal_consumption_smoothing} and \nameref{el:deriving_hicksian_demand}.

    Digging more deeply into these findings, this performance decline arose in part because the \models frequently ignored the specific problem posed and instead solved a simpler or otherwise different version of the problem. We examined model performance in the free-text QA adaptation and found that, in \num{36.2}\% of claude-3-5-sonnet's incorrect \nameref{el:intertemporal_consumption_smoothing} responses, a two‐year stream of future income was ``discounted'' as if it was all received one period into the future. In reality, given that cash flows occur in different periods (e.g., year \num{1} and year \num{2}) the optimal smoothing must discount each separately (e.g., using $0.99$ for year \num{1} and $0.99^2$ for year \num{2}) rather than lumping them together with a single division. Many other incorrect solutions were caused by smoothing consumption with a simpler Euler formula. While we stipulated that the subject was risk averse with a CRRA utility function, the Euler equation that the \models used was more appropriate for smoothing consumption when the utility is linear but ignores the smoothing factor induced by the curvature of CRRA utility. This occurred for claude-3-5-sonnet \num{11.0}\% of the time and gpt-4o \num{42.7}\% of the time. For more discussion on this see \Cref{app:intertemporal_consumption_smoothing}.

    A similar pattern emerged when the \models were tasked with deriving Hicksian demand functions. In many cases, the \models defaulted to the more familiar Marshallian approach---maximizing utility subject to a budget constraint---instead of minimizing expenditure subject to a utility constraint. This error did not occur uniformly across models. For example, gpt-4o erroneously applied a Marshallian derivation in roughly \num{51.2}\% of its incorrect responses, while gpt-4o-mini did so in about \num{23.4}\% of cases. Among the Anthropic models, claude-3-5-sonnet made this mistake in approximately \num{48.9}\% of incorrect instances, compared to only \num{15.2}\% for claude-3-haiku. These differences in error rates likely reflect the \models' varying capacities to engage with the question. More sophisticated models tended to generate responses that were more economically relevant, even if they did not always arrive at the correct derivation. It is also important to note that we can only identify an \model as using the Marshallian approach when it came to the right `Marshallian' answer.
    
    These findings underscore a broader concern: when faced with more complex tasks, sophisticated \models often give ``near-miss'' solutions that can appear correct at first glance but actually solve subtly different problems. This is especially worrisome in real-world use cases where a user may not easily be able to verify an answer.

    \subsection{\sixthParent}

    \begin{figure}[H]
        \centering
        \includegraphics[width=\linewidth]{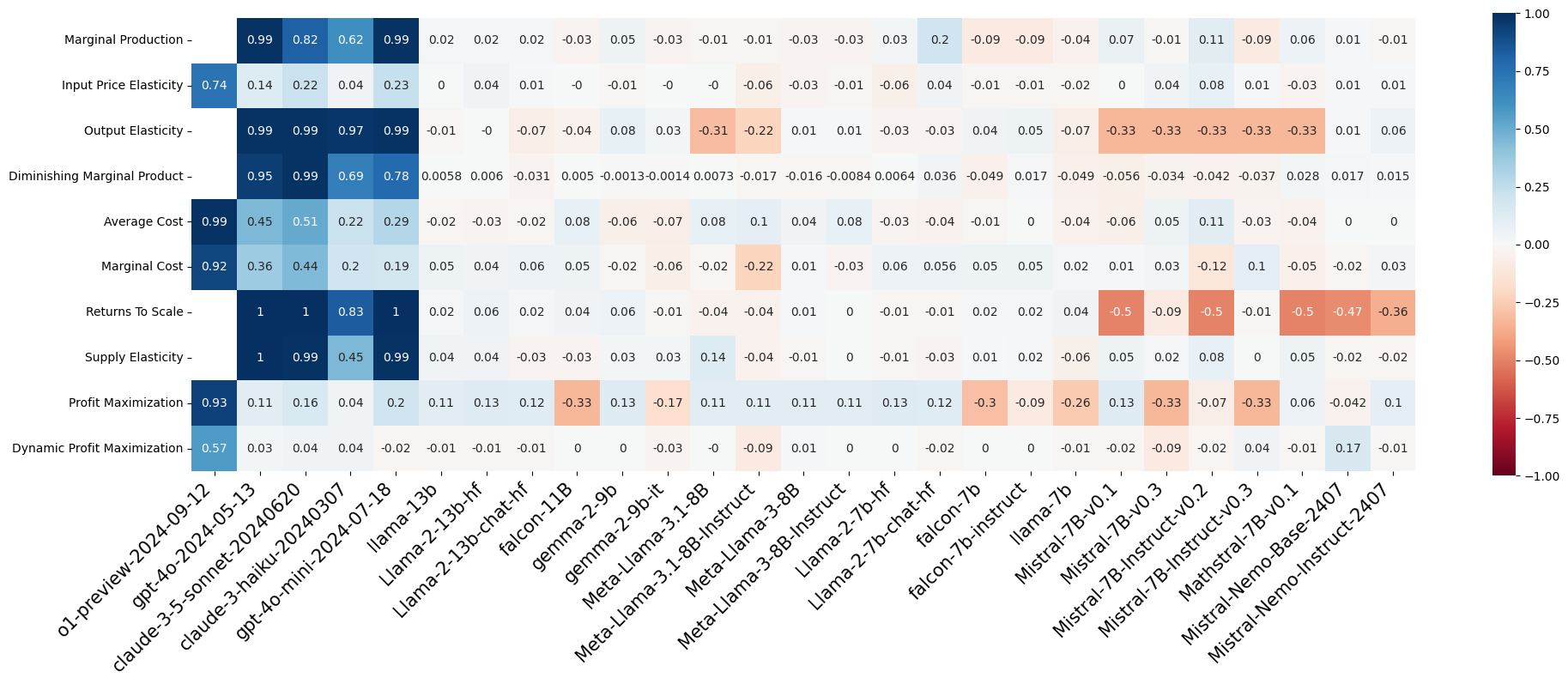}
        \caption{Heatmap plotting normalized accuracy performance of \models on elements within the \sixthParent \parent. Performance is on the \shown adaptation without CAR and we sort \models by parameter size (when available).}
        \begin{shownto}{arxiv}
            \Description[]{Heatmap plotting normalized accuracy performance of \models on elements within the \sixthParent \parent. Performance is on the \shown adaptation without CAR and we sort \models by parameter size (when available).}
        \end{shownto}
        \label{fig:producers_heatmap}
    \end{figure}

    Producer-side tasks in our benchmark often proved more challenging than their consumption-side counterparts. While \models generally performed well on simpler production-function properties (mirroring their success on basic consumer decision tasks), the more complex optimization elements exposed weaknesses. In particular, elements evaluating profit maximization were the most difficult elements in this \parent with \nameref{el:dynamic_profit_max}---a two-stage problem with interdependent calculations---being the most difficult in the entire benchmark. 

    In these complex elements, \models often failed to arrive at correct solutions through genuine optimization. Instead, we observed a ``cheating'' strategy in which the model, when appropriate, would plug in the candidate answer choices directly into the functions in the question and pick whichever yielded the best result, rather than solving via first principles (e.g., by taking derivatives). This was particularly evident in the \nameref{el:profit_max} element, where models were asked to determine the labor input that maximizes profit. Rather than identifying the profit-maximizing choice analytically, they simply tested each option and selected the highest-profit outcome. Indeed, every spot-checked instance of a correct response from gpt-4o and claude-3-5-sonnet used this shortcut (see \Cref{fig:profit_max_snippet} for an example). 
    
    While this approach is a common strategy for test-takers---humans and \models alike---it bypasses the economic reasoning that these elements were designed to assess. This is the issue our \emph{\hidden} adaptation was designed to highlight. When we evaluated \models using this adaptation, we observed significant performance degradation: gpt-4o's exact-match accuracy declined from \num{33.2}\%  (\shown) to \num{27.1}\% (\hidden), only \num{2}\% away from random guessing. This performance degradation was also visible on the CAR adaptation. When the correct answer was ``No other option is correct,'' claude-3-5-sonnet's exact-match accuracy dropped by \num{20.5}\%, o1-preview's by \num{15.1}\%, and gpt-4o's by \num{11.2}\%. Going even further, when evaluated on free-text QA, no model exceeded \num{10}\% exact-match accuracy.\footnote{We did not evaluate o1-preview on free-text QA because its reasoning tokens were not available at the time.}

    \begin{figure}
        \begin{minipage}{0.5\textwidth}
            \vspace{-1.85cm}
            \begin{tcolorbox}[
                colback=brown!10!white,
                colframe=brown!10!white,
                title=Question:,
                lower separated=true,
                coltitle=black,
                enhanced,
                top=4mm,
                boxed title style={colback=white!50!brown,colframe=white!50!brown},
                attach boxed title to top left={xshift=0.25cm,yshift=-3.5mm},
            ]
            
            \fontsize{9pt}{10pt}\selectfont
            John manages a company that produces premium leather wallets. The production output depends on the capital (K) he invests, represented by the function $3.97K^{0.45}$. At the moment, the wallets are sold at a price of $3.64$, and John has an initial capital of $K_1 = 3.3$. He needs to decide how much additional capital to invest for the next day's market. John is aware of the potential price distribution for his wallets for the next day, represented by price $2.7$ with probability $0.24$, price $2.14$ with probability $0.42$, price $3.52$ with probability $0.34$. There is also a cost associated with increasing his capital, given by $(K_1-K_2)^2$, and he applies a discount factor of $0.37$ to future revenues. To maximize his profits, how much additional capital should John invest?
            \begin{enumerate}[label=\Alph*., wide, labelwidth=!,labelindent=0.5pt]
                \item $0.44$ {\textcolor[HTML]{008000}{\textit{[Correct Answer.]}}}
                \item $0.14$
                \item $1.1$
                \item $0.09$
            \end{enumerate}
            \end{tcolorbox}
        \end{minipage}
        \hfill
        \begin{minipage}{0.48\textwidth}
        \begin{tcolorbox}[
            colback=blue!5!white,
            colframe=blue!5!white,
            title=GPT-4o:,
            lower separated=true,
            coltitle=black,
            enhanced,
            top=4mm,
            boxed title style={colback=white!80!blue,colframe=white!80!blue},
            attach boxed title to top left={xshift=0.25cm,yshift=-3.5mm},
        ]
            \fontsize{9pt}{10pt}\selectfont
            
            {\color{red}\textit{[Steps 1--6 omitted for brevity.] 
            }}%

            \textbf{Step 7: Evaluate the Net Profit for Different Values of \(\Delta K\)}
    
            We need to evaluate the net profit for the given options of additional capital investment:
    
            \colorbox{yellow}{Option A: \(\Delta K = 0.44\)}
            \[
            K_2 = 3.3 + 0.44 = 3.74
            \]
            \[
            \text{Production Output} = 3.97 \times 3.74^{0.45}
            \]
            \[
            \text{Revenue} = 2.7436 \times 3.97 \times 3.74^{0.45}
            \]
            \[
            \text{Discounted Revenue} = 0.37 \times \text{Revenue}
            \]
            \[
            \text{Cost} = (0.44)^2 = 0.1936
            \]
            \[
            \text{Net Profit} = 0.37 \times \text{Revenue} - 0.1936
            \]

            {\color{red}\textit{[Continues for the three remaining options.]
            }}%

            \textbf{Step 8: Compare the Net Profits}
    
            We need to compare the net profits for each option to determine which one
            maximizes John's profit. After calculating the net profits for each option, 
            we find that:

            \begin{itemize}[leftmargin=*]
                \item[--] Option A: \(\Delta K = 0.44\) yields the highest net profit.
            \end{itemize}
    
            Therefore, the best option for John is to invest an additional capital of 0.44.
    
            \textbf{Answer: A. 0.44}
    
        \end{tcolorbox}
        \end{minipage}
        \caption{This figure depicts a question in the \nameref{el:profit_max} element and corresponding snippet of GPT-4o's response. The response shows the model getting to the answer using the given options rather than deriving the answer from first principles. See the full response in \Cref{app:profit_max_example}.}
        \begin{shownto}{arxiv}
            \Description[]{This figure depicts a question in the \nameref{el:profit_max} element and corresponding snippet of GPT-4o's response. The response shows the model getting to the answer using the given options rather than deriving the answer from first principles. See the full response in \Cref{app:profit_max_example}.}
        \end{shownto}
        \label{fig:profit_max_snippet}
    \end{figure}


    \subsection{\seventhParent}

    \begin{figure}[H]
        \centering
        \includegraphics[width=\linewidth]{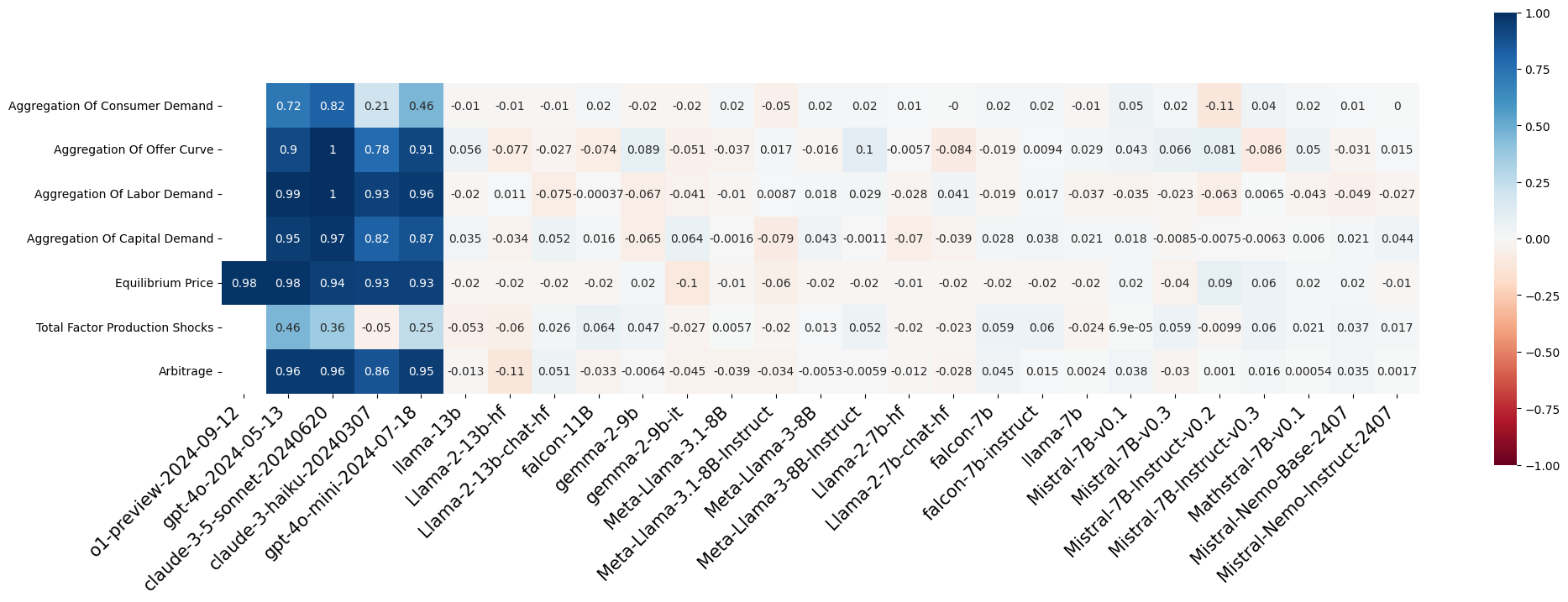}
        \caption{Heatmap plotting normalized accuracy performance of \models on elements within the \seventhParent \parent. Performance is on the \shown adaptation without CAR and we sort \models by parameter size (when available).}
        \begin{shownto}{arxiv}
            \Description[]{Heatmap plotting normalized accuracy performance of \models on elements within the \seventhParent \parent. Performance is on the \shown adaptation without CAR and we sort \models by parameter size (when available).}
        \end{shownto}
        \label{fig:multiagent_heatmap}
    \end{figure}

    \models faced a new twist in this section: economic outcomes depend on multiple consumers or producers acting in aggregate. Tasks included market-wide demand or supply aggregation, computing market-clearing prices, and identifying arbitrage opportunities. Overall accuracy here was higher than in earlier {\parent}s, largely because we presented the \models with already-derived demand and supply functions. Summation of given functions is inherently simpler than derivation from first principles and is thus less prone to the mathematical issues that often tripped up the \models in prior elements.
    
    \begin{wrapfigure}{r}{0.5\textwidth}
        \centering
        \includegraphics[trim={0cm, 0, 0, 0},clip,width=0.95\linewidth]{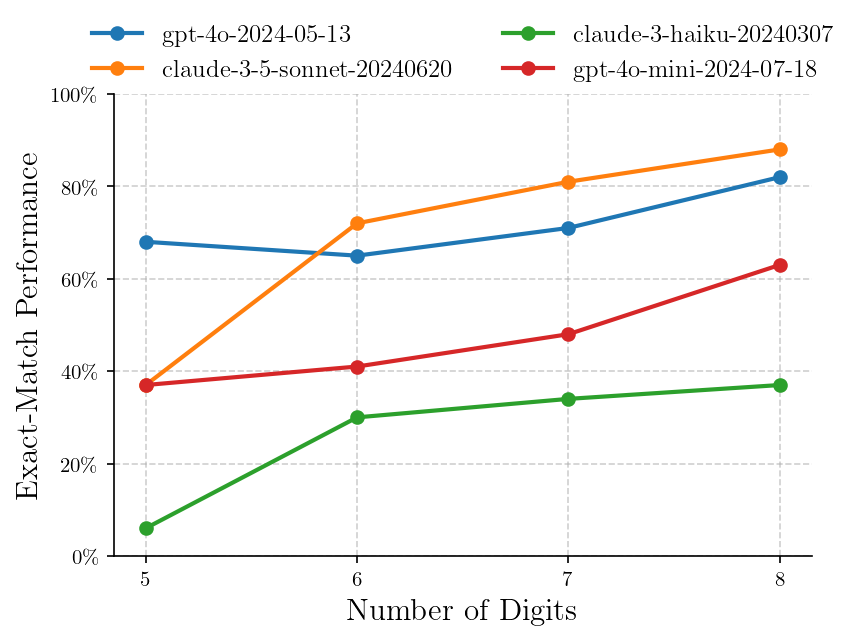}
        \caption{This figure plots exact-match performance on \nameref{el:agg_consumer_demand} on the \shown implementation of 0-CoT for the closed-source models against the number of digits of the correct answer.}
        \begin{shownto}{arxiv}
            \Description[]{This figure plots exact-match performance on \nameref{el:agg_consumer_demand} on the \shown implementation of 0-CoT for the closed-source models against the number of digits of the correct answer.}
        \end{shownto}
        \label{fig:agg_consumer_demand_main}
    \end{wrapfigure}

    Despite this relative success, one consistent pattern emerged: \models appeared the use the multiple-choice options when reasoning about how to aggregate demand or supply functions. This was particularly prevalent in the \nameref{el:agg_consumer_demand} element, which requires summing individual demand functions for a large number of consumers; here the performance gap between the \hidden and \shown adaptations was among the highest across our benchmark. In the \hidden version, models frequently failed to multiply the single-consumer demand by the number of consumers, yielding an incorrect quantity. By contrast, in the \shown version, \models correctly multiplied by the number of consumers. \Cref{app:agg_consumer_demand_example} illustrates how gpt-4o changed its response when only the adaptation was switched. 
    
    \Cref{fig:agg_consumer_demand_main} quantifies this phenomenon: As the number of digits in the correct answer increased from \num{5} to \num{6}, claude-3-5-sonnet's exact-match accuracy rose from \num{38}\% to \num{72}\%. This suggests that large numeric answers can serve as a ``nudge'' for models to think more carefully about summation or aggregation steps. This underscores how sensitive even state-of-the-art \models remain to textual cues, even in tasks that are otherwise conceptually straightforward.

    \subsection{\eighthParent}

    \begin{figure}[H]
        \centering
        \includegraphics[width=\linewidth]{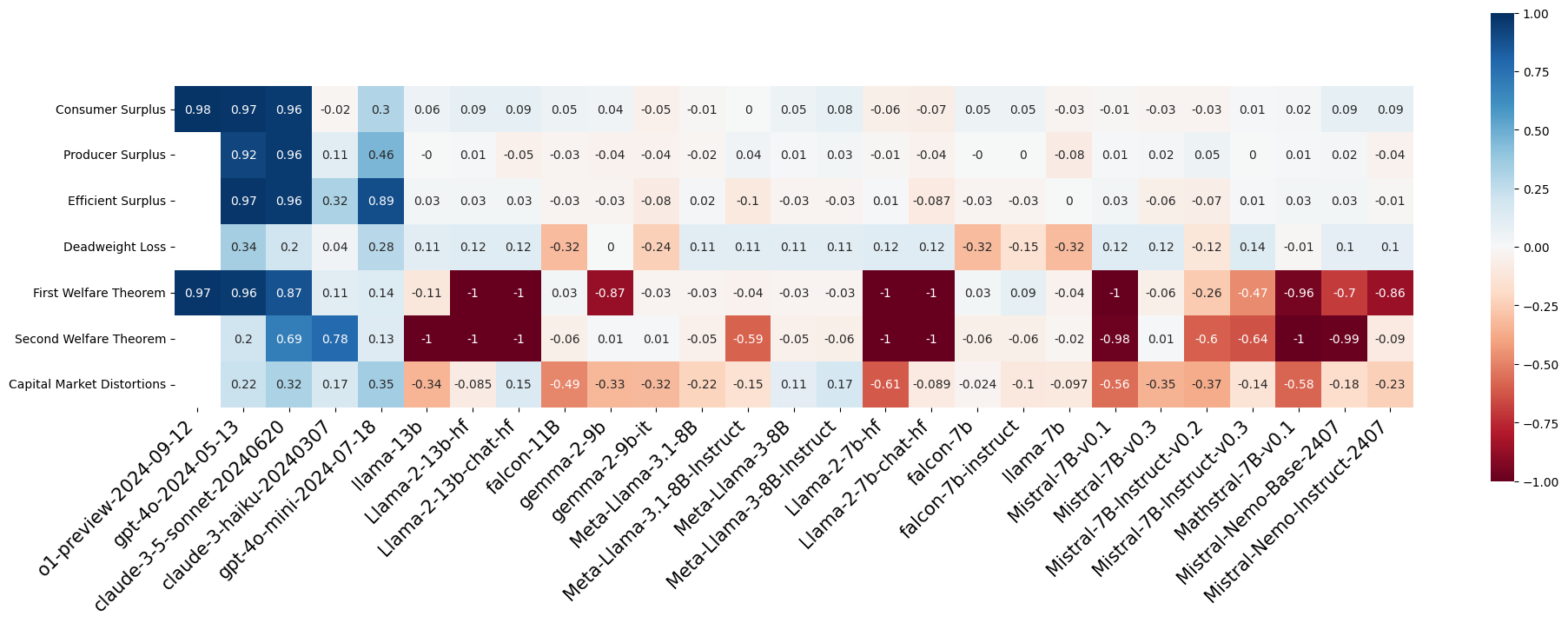}
        \caption{Heatmap plotting normalized accuracy performance of \models on elements within the \eighthParent \parent. Performance is on the \shown adaptation without CAR and we sort \models by parameter size (when available).}
        \begin{shownto}{arxiv}
            \Description[]{Heatmap plotting normalized accuracy performance of \models on elements within the \eighthParent \parent. Performance is on the \shown adaptation without CAR and we sort \models by parameter size (when available).}
        \end{shownto}
        \label{fig:equilibria_heatmap}
    \end{figure}
    
    We observed larger gaps in performance between the base and distilled versions of closed-source models on the elements in this \parent. While the base versions of the closed-source \models generally excelled on surplus-related elements, their distilled counterparts frequently failed to reach the correct answer. We found this gap surprising, given the relatively straightforward economic concepts and mathematics involved.
    
    More surprising, perhaps, was that none of the closed-source \models, except for o1-preview, was able to consistently compute the \nameref{el:deadweight_loss}; an element whose primary mathematical requirement is computing the area of a triangle. We discovered that models like claude-3-5-sonnet and gpt-4o often used an incorrect formula for computing deadweight loss and made errors in interpreting the marginal cost, a crucial step in the problem-solving process. To investigate these errors, we scored \models on the free-text QA adaptation. \Cref{fig:dwl_errors} shows the distribution of correct responses and specific errors for claude-3-5-sonnet and gpt-4o. While gpt-4o displayed performance better than random guessing, the majority of responses consisted of errors stemming from the use of an incorrect formula. claude-3-5-sonnet exhibited an even higher prevalence of incorrect formula errors, with nearly \num{44}\% of its responses relying on a particular incorrect formula for deadweight loss. Furthermore, gpt-4o was more susceptible to compounding issues, incorrectly computing marginal cost and using an incorrect deadweight loss formula, than claude-3-5-sonnet. We describe these errors in more detail in \Cref{app:deadweight_loss}.

    \begin{figure}
        \centering
        \begin{minipage}{0.9\textwidth}  
            \centering
            \includegraphics[trim={0 17.5cm 0cm 0cm},clip,width=\textwidth]{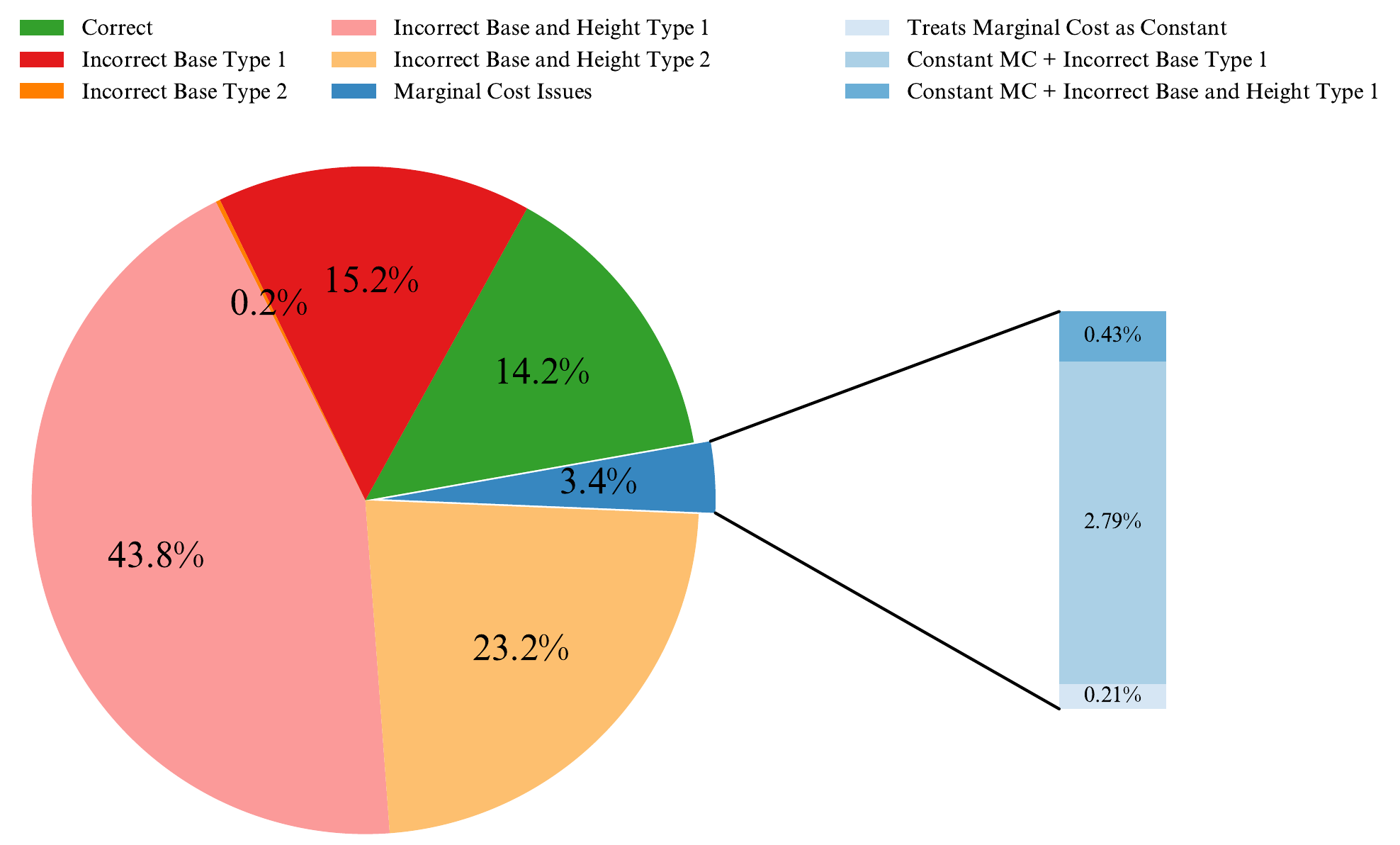}
        \end{minipage}
    
        \vspace{1em}  
    
        \begin{subfigure}[t]{0.48\textwidth}
            \centering
            \includegraphics[trim={0 0 0 3.5cm},clip,width=\textwidth]{images/claude-3-5-sonnet-20240620_pie_bar_chart.pdf}
            \caption{Claude-3.5 Sonnet}
            \label{fig:sonnet_dwl}
        \end{subfigure}
        \hfill
        \begin{subfigure}[t]{0.48\textwidth}
            \centering
            \includegraphics[trim={0 0 0 3.5cm},clip,width=\textwidth]{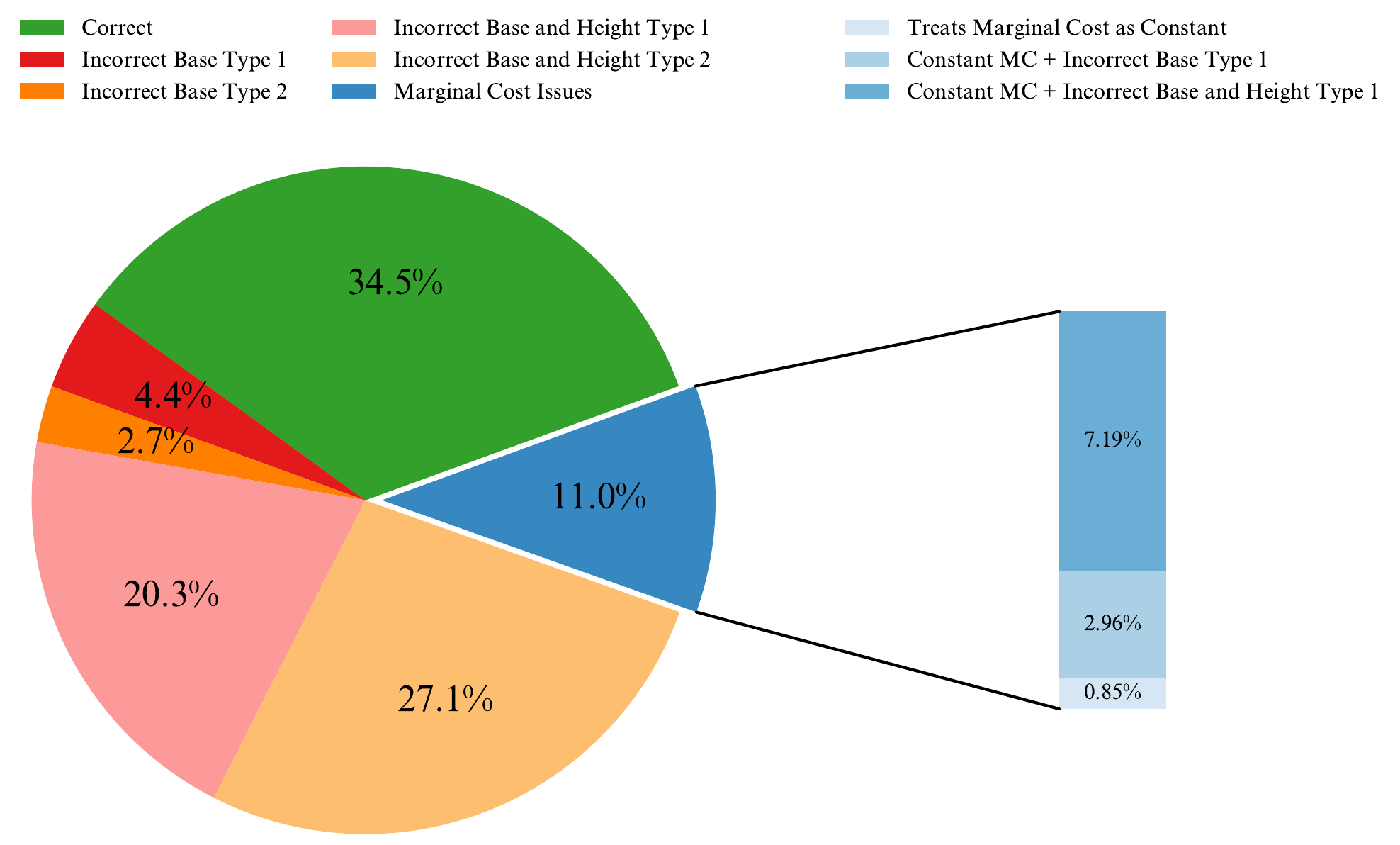}
            \caption{GPT-4o}
            \label{fig:4o_dwl}
        \end{subfigure}
    
        \caption{Error analyses of claude-3-5-sonnet and gpt-4o on the \nameref{el:deadweight_loss} element. In reds and oranges are failures due to incorrect computations of the deadweight loss area; in blue and further broken down are errors due to incorrectly interpreting the marginal cost. A more detailed description of what each error means can be found in \Cref{app:deadweight_loss}.}
        \begin{shownto}{arxiv}
            \Description[]{Error analyses of claude-3-5-sonnet and gpt-4o on the \nameref{el:deadweight_loss} element. In reds and oranges are failures due to incorrect computations of the deadweight loss area; in blue and further broken down are errors due to incorrectly interpreting the marginal cost. A more detailed description of what each error means can be found in \Cref{app:deadweight_loss}.}
        \end{shownto}
        \label{fig:dwl_errors}
    \end{figure}
    
    \subsection{Other Insights}
    \subsubsection{Adaptations.} We examined how providing multiple-choice options versus hiding them (free-text QA) impacted performance on four elements (\nameref{el:intertemporal_consumption_smoothing}, \nameref{el:profit_max}, \nameref{el:agg_consumer_demand}, and \nameref{el:producer_surplus}) that exhibited the largest gap in accuracy between hidden and shown adaptations. Our analysis revealed that performance was almost always worse under the free-text QA adaptation compared to the hidden adaptation, see \Cref{fig:hidden_vs_no_mc}. This performance gap appears to stem from the models' tendency to select the closest option to the free-text answer. \Cref{fig:percent_closest} shows the percentage of times that models were correct under the hidden adaptation but incorrect under the free-text adaptation due to guessing the closest answer. In almost all cases the majority of the gap was due to this phenomenon. We offer more discussion in \Cref{app:free_vs_hidden}.

    We also examined the effect of CAR on model performance. \Cref{fig:nota_frequency} shows that the fraction of times each model selects ``No other option is correct'' (NOTA) varies substantially across our benchmark, indicating different tendencies toward selecting NOTA vs. choosing among the listed options. On average, gpt-4o tended to select NOTA more frequently than the true rate, whereas claude-3-5-sonnet rarely did so; interestingly, the distilled models (gpt-4o-mini and claude-3-haiku) displayed these tendencies even more strongly.

    To assess how well each model identified cases where NOTA truly was the right choice, we aggregated element-level results into overall precision, recall, and F1 scores (\Cref{tab:car_results}). Precision measures how often the \model is right when it selects NOTA, whereas recall measures how often a model is right when NOTA is the correct answer. The F1 score is the harmonic mean of recall and precision, meaning it penalizes large differences between the two; both must be high for a strong F1. 

    \Cref{fig:nota_frequency} (in the appendix) shows the distribution of NOTA selection rates across individual elements. Notably, gpt-4o exhibited comparable precision and recall, whereas o1-preview outperformed it on both, selecting NOTA only when it was highly confident. In contrast, gpt-4o-mini had low precision, indicating a tendency to overpredict NOTA. Lastly, the Anthropic models rarely selected NOTA---but when they did, they seldom selected it correctly.

    \begin{table}[ht!]
        \centering
        \begin{tabular}{lccc}
        \toprule
        \textbf{Model} & \textbf{Recall} & \textbf{Precision} & \textbf{F1 Score} \\
        \midrule
        o1-preview        & 0.780 & 0.744 & 0.762 \\
        gpt-4o            & 0.704 & 0.698 & 0.701 \\
        claude-3-5-sonnet & 0.438 & 0.607 & 0.509 \\
        gpt-4o-mini       & 0.747 & 0.451 & 0.562 \\
        claude-3-haiku    & 0.051 & 0.580 & \num{0.093} \\
        \bottomrule
        \end{tabular}
        \caption{Aggregated performance on the CAR adaptation across \num{20} elements. Overall, o1-preview attains the best balance between correctly catching true CAR questions (recall) and avoiding extraneous CAR predictions. In contrast, claude-3-haiku is extremely conservative about labeling CAR (low recall), which leads it to miss the vast majority of CAR questions. The random guessing baseline is \num{0.25} for all metrics.}
        \begin{shownto}{arxiv}
            \Description[]{Aggregated performance on the CAR adaptation across \num{20} elements. Overall, o1-preview attains the best balance between correctly catching true CAR questions (recall) and avoiding extraneous CAR predictions. In contrast, claude-3-haiku is extremely conservative about labeling CAR (low recall), which leads it to miss the vast majority of CAR questions. The random guessing baseline is \num{0.25} for all metrics.}
        \end{shownto}
        \label{tab:car_results}
    \end{table}

    \subsubsection{Type Robustness.} Here, we examine how \model performance can fluctuate across several functional families commonly used in economic reasoning: Cobb-Douglas, Leontief, and linear.\footnote{We offer technical descriptions and the economic interpretations to these functions in \Cref{app:desc_functions}.} Each family poses its own difficulties---ranging from corner solutions in Leontief to exponent-based operations in Cobb-Douglas---that can trip up language models in non‐trivial ways. In fact, while some models may excel at one concept (such as summing exponents to determine returns to scale in a Cobb-Douglas production function), they can stumble on another (for instance, capturing how inputs must remain in fixed proportions under Leontief). As illustrated in \Cref{fig:type_robustness_main}, even among the three highlighted elements (e.g., \nameref{el:input_elasticity}, \nameref{el:diminishing_marginal_product}, and \nameref{el:returns_to_scale}), differences in performance underscore how the interplay between economic interpretation and mathematical representation can yield different outcomes across function families. 

    \begin{figure}
        \centering
        \includegraphics[width=\linewidth]{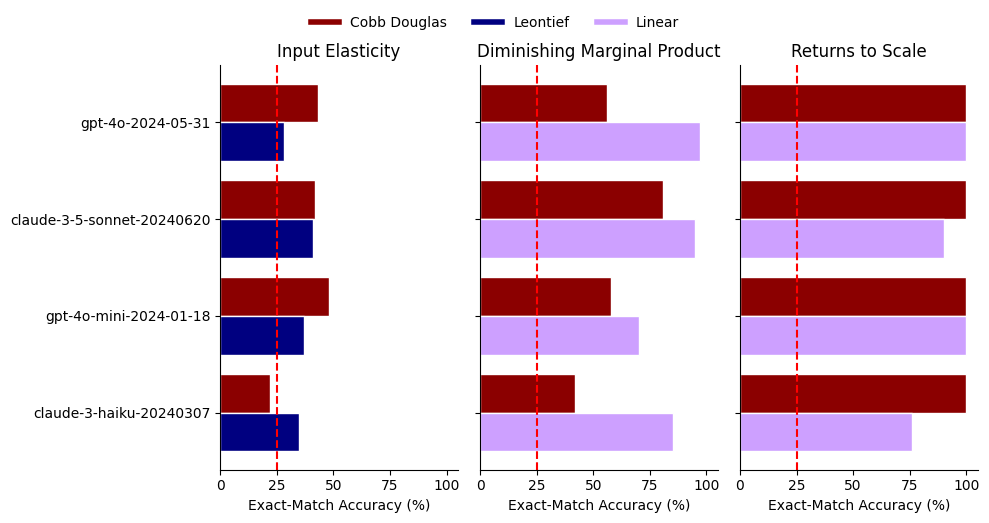}
        \caption{Exact‐match accuracy of all closed‐source models on three elements (\nameref{el:input_elasticity}, \nameref{el:diminishing_marginal_product}, and \nameref{el:returns_to_scale}) across three functional families (Cobb‐Douglas, Leontief, and Linear). In general, while Cobb‐Douglas often poses greater difficulty, performance does not consistently align with one family being the hardest for every element. For instance, while accuracy on Cobb-Douglas functions on \nameref{el:diminishing_marginal_product} is considerably lower than for Linear functions, the opposite trend is seen for the \nameref{el:returns_to_scale} element. The red dashed line indicates the random‐guessing baseline for comparison.}
        \label{fig:type_robustness_main}
        \begin{shownto}{arxiv}
            \Description[]{Exact‐match accuracy of all closed‐source models on three elements (\nameref{el:input_elasticity}, \nameref{el:diminishing_marginal_product}, and \nameref{el:returns_to_scale}) across three functional families (Cobb‐Douglas, Leontief, and Linear). In general, while Cobb‐Douglas often poses greater difficulty, performance does not consistently align with one family being the hardest for every element. For instance, while accuracy on Cobb-Douglas functions on \nameref{el:diminishing_marginal_product} is considerably lower than for Linear functions, the opposite trend is seen for the \nameref{el:returns_to_scale} element. The red dashed line indicates the random‐guessing baseline for comparison.}
        \end{shownto}
    \end{figure}
    
    \subsubsection{Domain Robustness.} While overall the variation across domains was limited, we observed noticeable differences in specific elements. In particular,  elements testing conceptual understanding of foundational principles (e.g., first welfare theorem) showed that certain domains provided more effective contextual cues for the \models. For example, in the consumer goods domain—where items like apples, chairs, or mugs are familiar in economic word problems—\models were more likely to recognize the task as an economic problem and anchor their reasoning in classical economic principles. 
    
    In contrast, the technology \domain, where the economic context could be interpreted as a real-world scenario, presented more challenges. The \models often failed to recognize what was being asked and equivocated when reasoning about the problem. The largest performance gaps appeared in the \nameref{el:welfare_theorem_1} and \nameref{el:welfare_theorem_2} elements. To give some examples, the gap in accuracy between the consumer goods and technology domains for claude-3-5-sonnet, claude-3-haiku, and gpt-4o-mini was respectively $0.657$, $0.48$, and $0.278$. 
    
\end{shownto}

\section{Discussion and Conclusions}

Our work introduces a novel benchmark specifically designed to evaluate \models' performance in non-strategic microeconomics, focusing on tasks that require a deep understanding of optimization, marginal analysis, and economic reasoning in individual decision-making contexts. This benchmark provides a comprehensive tool to assess the strengths and weaknesses of current models, revealing where they excel and where they struggle in applying foundational economic concepts. By identifying these areas, our benchmark can guide users in determining when \models can be trusted to perform well in economic analyses and when further development is needed.

In cases where models fall short, our benchmark serves as a practical resource for targeted improvements, e.g., via fine-tuning models, curating more specific datasets, or developing architectures better suited for microeconomic reasoning. These enhancements have the potential to impact a variety of economic applications, such as simulating consumer behavior, analyzing market dynamics, or conducting policy evaluations.

Looking ahead, we plan to expand our benchmark by incorporating additional elements from the microeconomics literature, deepening the evaluation of non-strategic decision-making. We encourage suggestions on new elements to include and make \autosteer public for others to add more elements or expand on the elements we have currently. We also intend to explore further experimentation with additional \models, adaptation strategies, and prompt configurations, along with more detailed analyses of model performance.

\section{Acknowledgements}

This work was funded by an NSERC Discovery Grant, a DND/NSERC Discovery Grant Supplement, a CIFAR Canada AI Research Chair (Alberta Machine Intelligence Institute), awards from Facebook Research and Amazon Research, and DARPA award FA8750-19-2-0222, CFDA \#12.910 (Air Force Research Laboratory). This paper draws on research supported by the Social Sciences and Humanities Research Council.

\clearpage 

\bibliography{ref}
\begin{shownto}{iclr}
    \bibliographystyle{iclr2025_conference}
\end{shownto}
\begin{shownto}{arxiv}
    \bibliographystyle{ACM-Reference-Format}
\end{shownto}

\newpage

\appendix

\section{Taxonomy of Non-Strategic Microeconomics}\label{app:taxonomy}



\subsection{\fifthParent}\label{setting:fifth}

We begin by characterizing the space of elements that test an agent's ability to optimally allocate their limited resources to goods and services they desire.  In economics and decision theory, the most primitive approach to describing the preferences of decision-makers is to use a function that maps a set of possible choices to the agent's optimal choice within that set.  Under a set of intuitive assumptions, such as \textit{transitivity} (i.e., if bundle $X$ is preferred to bundle $Y$, and $Y$ is preferred to bundle $Z$, then $X$ must be preferred to $Z$), it becomes possible to ``rationalize'' preferences by instead describing a utility function. This function assigns a real number to each bundle, and the agent selects the bundle with the highest utility.  

In this paper, we focus on these ``rationalizable'' preferences, where agent choice can be implemented as utility maximization constrained by prices and income. The solution to these consumer choice problems provides us with, among other things, individual demand functions, which describe the choice of each good or service as a function of prices and income. The individual demand functions for each good are essential when aggregating to the market demand in \nameref{mod:consumer_aggregation}, which in turn is used to find the price in a non-strategic equilibrium.  In addition, we test variations on the framework such as the agents ability to make tradeoffs between the quantity of goods they would need to be able to purchase for an increase in the amount of work they provide for a given wage (i.e., the elasticity of labor supply), as well as cases of choice under uncertainty where the agent is choosing between possible lotteries under rationalizability assumptions required for von Neuman expected utility.



\subsubsection{Properties of Utility Functions}\label{mod:utility_properties}
In this section, we test the ability of the agent to use utility functions as a means to compare preferences over different ``bundles'' of goods or services.  A key feature of economic reasoning in this context is for agents to consider how substitution between different goods in a bundle might achieve the same utility (i.e., map out the ``indifference curves'').  Key tests include correctly distinguishing between substitutes and complements in consumption, and calculating the marginal rate of substitution at a point on an indifference curve.  This logic is essential for both agents acting as a planner as we will see in \Cref{setting:eighth} and when fulfilling the role of choice under budget and income constraints, in \nameref{mod:deriving_demand}.  

\begin{el}[Marginal Utility]
    {The ability to calculate marginal utility for different types of demand curves such as quasilinear, Cobb-Douglas, and Leontief.}
\end{el}

\begin{el}[Diminishing Marginal Utility]
    {The ability to recognize the role of diminishing marginal utility in consumption decisions and the role of achieving interior solutions.}
\end{el}

\begin{el}[Marginal Rate of Substitution]\label{el:mrs_utility}
    {The ability to calculate the marginal rate of substitution between two goods in a consumption decision.}
\end{el}

\begin{el}[Tangency and the Marginal Rate of Substitution]
    {The ability to calculate the marginal rate of substitution between two goods in a consumption decision at a given point in the budget constraint as tangent to the indifference curve.}
\end{el}

\begin{el}[Substitutes and Complements]
    {The ability to distinguish between substitutes and complements in consumption decisions.}
\end{el}

\subsubsection{Deriving Demand}\label{mod:deriving_demand}
The \child in this section tests an agent's ability to solve a constrained utility maximization problem to derive a demand function---relying on the results of \nameref{mod:utility_properties}. We test the canonical classes of demand functions, check the duality of Marshallian demand and Hicksian demand, and ask the agent to derive these demand functions from first principles.

\begin{el}[Derivation of Marshallian Demand]
    {The ability to calculate the demand curve for a good given a utility function and a budget constraint.}
\end{el}



\begin{el}[Derivation of Hicksian Demand from Expenditure Minimization]\label{el:deriving_hicksian_demand}
    {The ability to calculate the demand curve for a good given a utility function and a budget constraint.}
\end{el}

\begin{el}[Duality of Hicksian Demand]
    {The ability to recognize that Hicksian demand (expenditure minimization) is dual to maximization in Marshallian Demand.}
\end{el}

\subsubsection{Comparative Statics of Demand}\label{mod:properties_demand}

This \child considers how agents reason about changes in prices or income, and their effects on the quantity of each good they would purchase. We test the classic law of demand, different types of goods (e.g., normal, inferior, and Giffen), and derive Engel curves from first principles.  The key tests are to ensure the agent rationally responds to changes in relative prices, and investigate their substitution between goods in a bundle.  In practice, these tests involve comparative statics of the $\mathrm{argmax}$ from the utility maximization of the previous section on \nameref{mod:deriving_demand}---i.e., using an Envelope theorem and perturbing prices or income.

\begin{el}[Law of Demand]
    {The ability to calculate the change in demand with the change in price for normal goods.}
\end{el}

\begin{el}[Price Elasticity of Demand]\label{el:price_elasticity_demand}
    {The ability to calculate the price elasticity of demand for a good given a utility function and a budget constraint.}
\end{el}

\begin{el}[Consumption Changes]
    {The ability to change the relative expenditures on goods given changes in relative prices with ordinary or Giffen goods.}
\end{el}



\begin{el}[Engel Curves]
    {The ability to calculate the Engel curve for a good given a utility function and a budget constraint.}
\end{el}

\begin{el}[Income Elasticity of Demand]
    {The ability to calculate the income elasticity of demand for a good given a utility function and a budget constraint.}
\end{el}

    

\subsubsection{Labor Supply}\label{mod:labor_supply}
While the proceeding elements tested tradeoffs in choices of bundles with different goods, services (in \nameref{mod:deriving_demand} and over lotteries in \nameref{mod:portfolio_choice}), often agents need to make a choice trading off between leisure and consumption. The elements in this \child test an agent's ability to optimally make that tradeoff by balancing the consumption goods required to compensate for decreased leisure---which leads to the labor supply elasticity central to many branches of economics.   Since goods must be purchased, agents will consider the relative wage from additional work compared to the price of goods.  This leads us to be able to test an agent's ability to distinguish real from nominal prices.

\begin{el}[Deriving Labor Supply]
    {The ability to calculate the labor supply curve given specific preference parameterizations such as separable preferences or homothetic preferences.}
\end{el}


\begin{el}[Labor Supply Elasticity]
    {The ability to calculate the elasticity of labor supply.}
\end{el}

\begin{el}[Marginal Rate of Substitution in Labor Supply]
    {The ability to calculate the marginal rate of substitution between consumption and leisure in a labor supply decision.}
\end{el}



\subsubsection{Dynamic Consumption Decisions}\label{mod:portfolio_choice}
Individuals often face decisions about how to trade off more consumption today at the cost of additional debt and less consumption in the future, and how best to plan for consumption with various contingencies with the future is uncertain.  Among other applications, this provides a formal model of how to best choose a mixture of financial assets---i.e., portfolios. Consequently, this subsection tests intertemporal consumption choices, optimal portfolio choice---which involves selecting a mix of assets that maximizes expected utility given the risks and returns associated with each asset. Understanding portfolio choice helps explain how consumers manage risk and make investment decisions, which is vital for financial planning and economic stability.

\begin{el}[Price of Risk with Mean-Variance Utility]
{The ability to calculate the price of risk for a mean-variance utility function.}
\end{el}

\begin{el}[State-Contingent Consumption]
    {The ability to calculate the optimal consumption given a utility function and a set of state-contingent consumption bundles.}
\end{el}

\begin{el}[Arbitrage]
    {The ability to recognize and execute arbitrage opportunities given two goods and prices you can resell.}
\end{el}

\begin{el}[Optimal Portfolio Choice with Bid-Ask Spreads]
    {The ability to calculate the optimal portfolio given bid-ask spreads.}
\end{el}

\begin{el}[Exponential Discounting]
    {The ability to exponentially discount future rewards or costs.}
\end{el}

\begin{el}[Intertemporal Consumption Smoothing]\label{el:intertemporal_consumption_smoothing}
    {The ability to calculate a smoothed consumption path and determine whether it is preferred to a non-smoothed path.}
\end{el}

\subsection{\sixthParent}\label{setting:sixth}

In the previous section, we derived how an agent facing a set of prices would choose the quantity demanded of each good or service to maximize their utility function.  We also tested the amount of time that an agent might choose to wok (i.e., the quantity of labor supplied) given market wages---where the agent trades off the additional goods they might purchase against the lost leisure time they must forgo.  Here, we look at the other side of the market and test an agent's ability to operate a production technology to maximize profits.  Facing market prices for all production factors (e.g., wages and the capital) and the market price of the good or service they produce, the agent chooses the quantity of each factor of production and the total output.  Parallel to \fifthParent, in \nameref{mod:properties_production} we first test general properties of production functions to ensure the agent can reason about substitution between factors, economies of scale in production, etc.  Then in \nameref{mod:deriving_factor_demand}  we solve the firms optimal profit maximization problem to determine the optimal choice of factors of production and output given a set of market prices.  Finally, in \nameref{mod:production_statics} we test the agents ability to reason about comparative statics on prices and their impact on factor demand and firm output. 

\subsubsection{Properties of Production Functions}\label{mod:properties_production}
Production functions in these environments take continuous inputs of each factor, which lets us test an agent's ability to conduct marginal thinking when choosing the composition of inputs.  For example, by knowing the hourly wage of hiring an additional worker, the additional output the worker might produce using the particular production process, and the price they can sell the firm's output, they can decide whether hiring the additional worker is profitable.  In the absence of prices, this section tests basic decision making of the agent for understanding substitution between factor of production, marginal products for each input, and the understanding of the returns to scale of a production process.

\begin{el}[Marginal Products]
    {The ability to calculate separate marginal products for a production function with multiple inputs (e.g., labor and capital).}
\end{el}

\begin{el}[Input Price Elasticity]\label{el:input_elasticity}
    {The ability to calculate the responsiveness of output to a proportional change in a specific input's cost, holding all other inputs constant.}
\end{el}

\begin{el}[Output Elasticity]\label{el:output_elasticity}
    {The ability to calculate the output elasticity of an input in a production function.}
\end{el}

\begin{el}[Elasticity of Substitution]
    {The ability to calculate the marginal elasticity of substitution between inputs in a production function.}
\end{el}

\begin{el}[Diminishing Marginal Products]\label{el:diminishing_marginal_product}
    {The ability to calculate the diminishing marginal products for a production function with multiple inputs.}
\end{el}

\begin{el}[Average and Marginal Costs]\label{el:average_cost}
    {The ability to calculate average and marginal costs given a production function and input prices, and use them to determine scale.}
\end{el}

\begin{el}[Returns to Scale]\label{el:returns_to_scale}
    {The ability to determine the proportional change in output resulting from a proportional change in all inputs in a production function.}
\end{el}

\subsubsection{Deriving Factor Demand}\label{mod:deriving_factor_demand}
This \child tests the agent's ability to act in the role of a profit maximizer in non-strategic situations where they take as given the price which they could sell goods they produce, and must pay for inputs to their production process at market rates (e.g., a competitive wage).  Whereas in \nameref{mod:deriving_demand}, the agent was solving a utility maximization problem subject to a budget constraint, here they solve a profit maximization problem constrained by a production function.  We test decisions on the quantity and composition of inputs, and the quantity of output for canonical production functions such as Cobb-Douglas and Leontief production functions given the agent's understanding of production functions from \nameref{mod:properties_production}.  The agent is asked to derive the factor demand functions from first principles from profit maximization and test their ability to reason with the dual cost-minimization formulation---analogous to the Hicksian vs. Marshallian demand of \nameref{mod:deriving_demand}.

\begin{el}[Profit Maximization]\label{el:profit_max}
    {The ability to calculate the optimal input bundle for a firm given a production function and input prices. Examples of given production functions: Cobb-Douglas, Leontief, Perfect Substitutes, CES production, CRS production, fixed costs.}
\end{el}

\begin{el}[Expenditure Minimization]
    {The ability to calculate the optimal input bundle for a firm given a production function and input prices.}
\end{el}

\begin{el}[Duality of Profit Maximization and Expenditure Minimization]
    {The ability to recognize that profit maximization is dual to expenditure minimization in production decisions and achieve consistent solutions.}
\end{el}

\subsubsection{Comparative Statics with Production}\label{mod:production_statics}

This \child considers how agents reason about changes in the prices at which they can sell their goods, as well as changes in the costs of producing those goods.  In particular, we can test how this affects their optimal choice of inputs to their production process (e.g., how many people to hire or robots to lease). We test comparative statics on the prices of inputs to the production function, changes to the underlying production technology, and substitution between goods for classic production functions such as Cobb-Douglas and Leontief.  Analogous to the relationship between \nameref{mod:deriving_demand} and \nameref{mod:properties_demand}, these tests involve comparative statics of the $\mathrm{argmax}$ from the profit maximization of \nameref{mod:deriving_factor_demand}---i.e., using an Envelope theorem and perturbing factor prices.

\begin{el}[Price Elasticity of Supply] 
{The ability to calculate the price elasticity of supply for a good given a production function and input prices.}
\end{el}

\begin{el}[Shephard's Lemma]
    {The ability to calculate factor demands given a cost function using the derivatives with respect to prices.}
\end{el}

\begin{el}[Input Price Elasticity]
    {The ability to calculate how the optimal input bundle changes with changes in input prices for a given production function.}
\end{el}

\begin{el}[Total Factor Productivity]
    {The ability to calculate total factor productivity given a production function and input prices}
\end{el}

\subsubsection{Dynamic Production Decisions}\label{mod:dynamic_production}
While \nameref{mod:deriving_factor_demand} tested the ability of agents to make static (i.e., within-period) decisions on the mix of input factors to maximize profits, many producer problems are inherently dynamic.  For example, we can test if an agent can optimally choose the amount of capital to purchase given forecasts of future consumer demand and prices or choose how much to adjust the labor force in cases when labor is difficult to relocate due to frictions such as hiring and firing costs.  Finally, agents are tested on their ability to make optimal entry and exit decisions based on their forecasted profits in an evolving market.

\begin{el}[Dynamic Profit Maximization]\label{el:dynamic_profit_max}
    {The ability to calculate the optimal investment decision given a production function and input prices.}
\end{el}

\begin{el}[Entry and Exit Decisions]
    {The ability to calculate the optimal entry and exit decisions given a production function and fixed costs.}
\end{el}

\subsection{\seventhParent}\label{setting:seventh}
This \parent tests the core logic of the relationship between supply-and-demand and prices, building on the tests of optimal behavior in \cref{setting:sixth} and \cref{setting:fifth}.  Economists refer to ``general equilibrium'' as the process where equilibrium prices and quantities emerge with a large number of non-strategic, price-taking market participants interact.  Unlike the strategic models found in \steer, the assumption is that the market interactions that lead to this equilibrium occur through an unspecified process that clears markets (i.e., a ``Walrasian auctioneer'' or ``invisible hand'').

In particular, for non-strategic settings, all market participants take prices as given and choose the quantity demanded or supplied in each market.  For example, consumers jointly decide on the quantity demanded of goods and services given relative prices, and the quantity of labor supplied given a wage.  Simultaneously, producers choose the quantity supplied of the good and the demand of each factor of production.  With a large number of non-strategic market participants we can test the agents ability aggregate all of their supply and demand functions to calculate a market-level supply and demand.  Finally, given the aggregated supply and demand functions for each market, we can test whether an agent can find the market clearing price where supply is equal to demand in equilibrium---given their internal model of all the market participants.

 In this section, we organize by markets rather than by the role of a decision maker, as in the previous sections.  For example, in the goods market we first ensure agents understand how individual demand functions from \nameref{mod:deriving_demand} aggregate to a market demand function for the good given a price, then that the agent understands how to aggregate the output from each producer at a given price from \nameref{mod:production_statics}, and finally that the agent is able to calculate the price which would equate demand and supply and clear the market in a non-strategic setting.  Factor markets are treated similarly.

Finally, given a system of equations that defines an equilibrium price we can perturb primitives (e.g., technological factors, distortions on decisions such as tax rates, or exogenous prices not determined in equilibrium) to see how the market clearing price would respond.  That is an essential tool for agents to be able to reason about the impact of interventions and distortions in \Cref{setting:eighth}.

\subsubsection{Consumer Goods Market Aggregation}\label{mod:consumer_aggregation}
The market clearing prices in general equilibrium arise from the separate market-level demand and supply curves, which sums the demand or supply across all market participants at a given price.  Here we test the aggregation of demand functions derived from individual preferences, as in \nameref{mod:deriving_demand} and  \nameref{mod:properties_demand}, to a market demand function that summarizes the total quantity demanded across all agents at a given price.  Central to the tests is to verify that the agent can aggregate the demands of market participants with heterogeneous preferences.  On the other side of the market, we test if the agent can aggregate the ``supply functions'' resulting from the optimal choice of factors in \nameref{mod:deriving_factor_demand} and \nameref{mod:production_statics}.

\begin{el}[Aggregation of Consumer Demand] \label{el:agg_consumer_demand}
    {The ability to calculate the aggregate demand for a good given primitives of demand into expenditure shares.}   
\end{el}





\begin{el}[Aggregation of Offer Curve for the Good]
    {The ability to calculate the aggregate supply of a good given primitives of supply into production functions.}
\end{el}

\subsubsection{Factor Market Aggregation}\label{mod:factor_aggregation}
As with the case of the goods market in \nameref{mod:consumer_aggregation} the market demand and supply for factors of production are essential to find the market clearing price.  For example, we test whether the agent can aggregate the individual labor supply curve decisions from market participants who work at a particular wage, following \nameref{mod:labor_supply}, into a market labor supply curve.  On the other side of the market, we test whether the agent can aggregate the labor demand in \nameref{mod:deriving_factor_demand} from producers into a market labor demand curve.  The same tests are essential for all factors of production, including capital.

\begin{el}[Aggregation of Labor Demand]
    {The ability to calculate the aggregate demand for labor given primitives of demand into expenditure shares.}
\end{el}

\begin{el}[Aggregation of Capital Demand]
    {The ability to calculate the aggregate demand for capital given primitives of demand into expenditure shares.}
\end{el}

\begin{el}[Aggregation of Labor Supply]
    {The ability to calculate the aggregate supply of labor given primitives of supply into production functions.}
\end{el}

\begin{el}[Aggregation of Fixed Factor Supply]
    {The ability to calculate the aggregate supply of capital given primitives of supply into production functions.}
\end{el}






\subsubsection{Prices in Static Market Equilibrium}\label{mod:static_equilibrium}
In this \parent we test the agent's ability to reason about how prices emerge in non-strategic setting as a process of equating supply and demand, which in turn relies on their ability to aggregate those market demand functions from consumer and producer behavior.

More specifically, the core logic of general equilibrium is to find the equilibrium price by taking the aggregated demand and supply functions for each market  and find the prices which would equate demand and supply.  For example, the supply and demand functions for the good, as a function of the price, in \nameref{mod:consumer_aggregation}; or the supply and demand functions for factors of production, as a function of factor prices in \nameref{mod:factor_aggregation}.  This is done market by market, taking all other prices as given---which requires the agent reason through comparative statics of the solution to a system of equations while keeping everything else fixed.

\begin{el}[Find Equilibrium Price]\label{el:find_eq_price}
    {The ability to calculate the equilibrium prices given a production function and a demand function.}
\end{el}

\begin{el}[Factor Shares in Equilibrium]
    {The ability to calculate the factor shares in a competitive equilibrium given a production function and input prices.}
\end{el}


\subsubsection{Comparative Statics of Equilibrium Prices}\label{mod:comparative_equilibrium}
Here, we test whether agents can reason about how prices and allocations (e.g., labor, capital, and goods) would respond to changes in the environment.  The canonical tests are to see how changes in model primitives (e.g., productivity of the production process) or exogenous forces from outside the model (e.g., impact of weather), change the equilibrium price and allocations of labor, capital, etc. that would clear the market and equate demand and supply.

\begin{el}[Comparative Statics with Total Factor Production Shocks]\label{el:tfp_shocks}
    {The ability to calculate how equilibrium prices change with changes in input prices for a Cobb-Douglas production function.}
\end{el}

\begin{el}[Comparative Statics with Inelastic or Perfectly Elastic Supply]
    {The ability to calculate how equilibrium prices change with changes in input prices for a production function with inelastic or perfectly elastic supply.}
\end{el}


\subsection{\eighthParent}\label{setting:eighth}
In \seventhParent, we tested an agents ability to reason about equilibrium prices and quantities arising from supply and demand decisions in a non-strategic setting.  Although preferences were reflected in the underlying supply and demand functions themselves (i.e., utility maximization in the consumption decisions of \fifthParent and profit maximization in the production decisions of \sixthParent), the equilibria in \seventhParent do not necessarily reflect broader social preferences.

However, we can still ask whether the resulting ``allocations'' (i.e., the physical goods produced and how they are distributed to individuals, the amount of hours worked, and the physical capital installed) from the ``invisible hand'' in \seventhParent compare to a alternative ways of allocating resources which may directly take social preferences into account.  A central result of economics in non-strategic settings is that absent market imperfections and market power (i.e., when self-interested agents cannot directly manipulate prices because they are too small) the competitive equilibria of \seventhParent typically yields the same allocations a benevolent planner might choose.

In this section, we consider how a social planner would evaluate the underlying welfare, efficiency, and inequality that comes about in non-strategic equilbria with prices derived from equating supply and demand.  This leads to testing the ability of the agent to evaluate Pareto efficiency, consider the welfare theorems, evaluate Pigouvian externalities, and weigh the welfare impact of various market interventions which change the equilibria derived in \seventhParent.  

\subsubsection{Welfare and Decentralization}\label{mod:welfare}
In this section, we test whether the agent can determine cases where the the competitive equilibrium they calculate would yield the same distribution of resources and consumer welfare as that of a benevolent social planner directly making the consumption and production decisions of all agents directly (also known as the ``Welfare Theorems").  In cases where the supply-and-demand relationships lead to the same results as those of a planner, the competitive equilibrium and its prices are said to ``decentralize'' the problem of a social planner.  We then test that the agent recognizes cases where the welfare theorems fail, and can calculate the degree of welfare loss due to the distortions.

\begin{el}[First Welfare Theorem]\label{el:welfare_theorem_1}
    {The ability to recognize that a competitive equilibrium is Pareto efficient.}
\end{el}

\begin{el}[Second Welfare Theorem]\label{el:welfare_theorem_2}
    {The ability to recognize that any Pareto efficient allocation can be achieved as a competitive equilibrium with prices.}
\end{el}

\begin{el}[Consumer Surplus]\label{el:consumer_surplus}
    {The ability to calculate consumer surplus given a demand curve and a price.}
\end{el}

\begin{el}[Producer Surplus]\label{el:producer_surplus}
    {The ability to calculate producer surplus given a supply curve and a price.}
\end{el}

\begin{el}[EFficient Surplus]
    {The ability to calculate the total surplus in a competitive equilibrium and recognize that it is maximized in the competitive equilibrium.}
\end{el}

\begin{el}[Deadweight Loss of a Monopoly]\label{el:deadweight_loss}
    {The ability to calculate the deadweight loss of a monopoly given a demand curve and a supply curve.}
\end{el}

\subsubsection{Welfare Analysis of Market Equilibrium}\label{mod:analysis_equilibrium}
In this section, we focus on the agent's ability to evaluate welfare implications of various forms of market equilibrium, particularly how different policies and distortions impact overall efficiency and resource allocation. The agent is tested on their understanding of how different interventions---such as taxes, subsidies, and price controls---affect welfare outcomes, and their ability to distinguish between distortionary and non-distortionary policies.

\begin{el}[Identify Non-Distortionary Taxes]
    {The ability to identify taxes which do not distort the allocation of resources.}
\end{el}

\begin{el}[Irrelevance of Tax Incidence]
    {The ability to recognize that the incidence of a tax does not depend on who is legally responsible for paying the tax.}
\end{el}

\begin{el}[Labor Supply Distortions]
    {The ability to determine the extent that labor taxes will distort labor supply and change aggregates and prices.}
\end{el}

\begin{el}[Capital Market Distortions]\label{el:cap_market_distortions}
    {The ability to identify that taxing a fixed factor is non-distortionary, but distorts with dynamic accumulation.}
\end{el}

\section{Mitigating Data Contamination with \autosteerbig}\label{app:autosteer}

Data contamination, where training data inadvertently includes information from test sets, poses significant challenges in machine learning, leading to overestimated model performance and compromised generalization capabilities. To address this, we implemented a structured dataset generation methodology incorporating human oversight, controlled data generation, and style transfer techniques. This appendix details our approach and its alignment with best practices in the literature.

The \autosteer methodology provides a systematic approach to generating datasets that mitigates the risk of data contamination, ensuring the integrity of benchmarks and the validity of results. Below, we outline the key aspects of \autosteer that address this issue:

\subsection{Challenging Models with Rephrasings:}

Rephrasings are known to cause significant variance in model performance, as demonstrated in the GSM-Symbolic dataset \cite{mirzadeh2024gsmsymbolicunderstandinglimitationsmathematical} and other studies \cite[e.g.,][]{zhu2024promptrobustevaluatingrobustnesslarge, wang2023largelanguagemodelsreally} highlighting how syntactic or stylistic changes can challenge generalization. In \Cref{app:rephrasing_var}, we also show that much of the observed variance in LLM performance arises from these rephrasings, underscoring their role in robust evaluations. \autosteer leverages this phenomenon to craft diverse rephrased questions that test beyond rote learning.

\subsection{Dynamic Question Generation:}
 
\autosteer generates new questions through a structured process that balances diversity and consistency. Questions are systematically rephrased or style-transferred to ensure they are different enough from the original templates to prevent memorization while retaining the same core meaning. This approach reduces the risk of overlap with pre-trained data while preserving the focus of the assessment.

The rapid advancement of large language models necessitates benchmarks that can evolve just as quickly. To address this, \autosteer incorporates a user interface that allows users to regenerate entire datasets with minimal effort. By modifying domains, seeds, or even resampling numerical values, users can quickly produce an entirely new dataset with minimal effort. This adaptability ensures that benchmarks remain fresh and resistant to contamination as models advance.

\section{Technical Descriptions of Adaptations}\label{app:adaptations}

\subsection{RPM (Conditioning):}

Given the \model's output distribution over all possible tokens, filter to include only those that correspond to valid options. For example, if a question has four options then get the probabilities corresponding to `A', `B', `C', and `D'. Then, compute softmax over the valid options to normalize the filtered probabilities into a distribution. 

\subsection{RPM (Mixing):}
Alternatively, we restrict the output distribution to only valid option tokens $O$ as follows:
$
\alpha \cdot p(o) + (1-\alpha)\nicefrac{1}{|O|},
$ where $o\in O$, $p(o)$ is the probability the \model assigns to each token it outputs, and $\alpha = \sum_{o\in O} p(o)$. We then compute the softmax to normalize the resulting probabilities into a distribution. 

In the mixing approach, if an \model is confident in a valid option token the resulting distribution will place high probability on that token, but if an \model places negligible probability on the valid option tokens then the resulting distribution will more closely resemble a uniform distribution.

\subsection{Prompt for Getting Answers from Free-Text QA}\label{app:free_text_prompt}

\begin{tcolorbox}[colback=black!5!white,colframe=black!5!white,title=User Message:,coltitle=black,left=5pt,toptitle=0.25mm,fonttitle=\bfseries]
Report the answer that the following reasoning ended up with. Do not solve the question just look at the explanation text and report the answer if it exists or None if not. Just report the number or None.

Q: \{question\_text\}

Explanation: \{model\_response\}
\end{tcolorbox}

\section{Technical Descriptions of Metrics}\label{app:metrics}

\subsection{Accuracy.}\label{app:accuracy}
Accuracy is the most broadly used metric for evaluating \models. We define accuracy metrics as metrics that only look at the \outputtoken that the \model outputs. 

\subsubsection{Exact-Match accuracy}\label{metric:exact_match}
This is the fraction of questions answered correctly.

\subsubsection{Normalized accuracy} \label{metric:normalized_accuracy}
Elements can differ in their number of multiple choice options, leading to differences in the exact-match accuracy of random guessing. We can compensate for this by reporting the gap between the \model's exact-match accuracy and random guessing \citep{budescu1993guess}. We compute normalized accuracy for an element as follows:
$\sum_{i = 1}^{N} a_i(t_i) - \frac{1 - a_i(t_i)}{|O_i| - 1},$
where $t_i$ is the \outputtoken the \model outputs for question $i$, $a_i$ is the indicator describing whether the \outputtoken is correct or not, $N$ the number of questions in the element, and $|O_i|$ the number of options in the question. In other words, normalized accuracy rewards an \model with 1 point for every correct answer and penalizes an \model by 1 over the number of options minus 1 for each incorrect answer.

\subsection{Calibration}\label{app:calibration}
It can also be useful to understand how confident an \model is in its  responses and the extent to which these confidence levels align with accuracy.

\subsubsection{Expected Calibration Error} \label{metric:ece}
We follow \citet{liang2022holistic} and \citet{ramansteer} in measuring the confidence of an \model's response and computing the expected calibration error \citep[ECE;][]{naeini2015obtaining, guo2017calibration}. ECE measures how closely the probability an \model assigns to its top answer matches the actual probability of the correct answer, which in our case is 1. ECE first splits the data into $M$ equally spaced bins, where each bin contains the probabilities the model assigned to their top token in that range: e.g., let $p^{\max}$ be the set of most probable tokens for each question then if $M = 2$, then the first bin $B_1 = \{p  ~|~ p\in p^{\max} ~\mbox{and} ~p\in [0, 0.5]\}$. It is then defined as $\sum_{i \in [M]} \nicefrac{|B_i|}{N}\cdot |\texttt{acc}(B_i) - \texttt{conf}(B_i)|,$ where $\texttt{conf}(B_i)$ is the average probability the \model assigned to its top token in bin $B_i$, and $\texttt{acc}(B_i)$ denotes the exact-match accuracy in bin $B_i$. We allow users to choose the number of bins, however, we set $M=10$ uniformly spaced over the interval $[0, 1]$ as is standard.

\subsubsection{Brier Score}\label{metric:brier_score}
The Brier Score of an element is defined as
$$
\sum_{i=1}^{N}\frac{1}{|O_i|}\sum_{o\in O_i}(p_i(o) - a_i(o))^2,
$$ where $p_i(o)$ is the probability the \model assigns to option $o$ in question $i$. Thus, if an \model is overly confident in an incorrect answer (e.g., assigns a probability of 0.9 to a wrong option), the Brier Score will penalize it more heavily.

\subsubsection{Expected Probability Assignment}\label{metric:epa}
EPA measures how much probability mass an \model assigns to the correct answer option out of all possible options. It is defined as: $\nicefrac{1}{N}\sum_{i\in[N]}p^*_i$, where $p^*_i$ is the predicted probability that the \model assigns to the correct option for question $i$.

\subsection{Robustness} \label{app:robustness}
Elements are not the lowest level of granularity in our benchmark and aggregating on an element-by-element basis may hide where {\model} performance varies. We introduce three robustness metrics that aggregate on the component fields of an element: \domains, \qtypes, and \Tags. We compute the \domain (\qtype, \tags) robustness on each element by taking the minimum normalized accuracy over all \domains (\qtypes, \Tags). 

\section{Technical Descriptions of Functional Families}\label{app:desc_functions}

In this section, we describe the functional forms that we use in testing economic concepts. Each can be applied to \emph{consumer} problems (as utility functions) or \emph{producer} problems (as production functions). We highlight the canonical mathematical form and note any technical differences in interpretation when modeling consumers versus producers.

\subsection{Cobb-Douglas}

The \emph{Cobb-Douglas} functional form is one of the most frequently used due to its  tractable properties and partial elasticities interpretation. Suppose there are \(n\) goods (or inputs). For a producer with input vector \(\mathbf{x} = (x_1,x_2,\dots,x_n)\), a typical Cobb-Douglas production function can be written as:
\[
  f(\mathbf{x}) \;=\; A \, x_1^{\alpha_1} \, x_2^{\alpha_2} \,\cdots\, x_n^{\alpha_n},
\]
where \(A > 0\) is a scale parameter and each \(\alpha_i \geq 0\). For a consumer's utility function, the same functional family looks like:
\[
  u(\mathbf{q}) \;=\; q_1^{\beta_1} \, q_2^{\beta_2} \,\cdots\, q_n^{\beta_n},
\]
where \(\mathbf{q} = (q_1, q_2,\dots,q_n)\) are quantities of goods consumed, and 
\(\beta_i \geq 0\). Economically, \(\alpha_i\) (or \(\beta_i\)) often reflect 
the relative importance (or expenditure share) of each input (or good).

\subsection{Leontief}

A \emph{Leontief} functional form encodes strict complementarity. A producer's Leontief production function is described as:
\[
  f(\mathbf{x}) 
  \;=\; \min\biggl\{\tfrac{x_1}{a_1},\; \tfrac{x_2}{a_2},\;\dots,\;\tfrac{x_n}{a_n}\biggr\},
\]
where each \(a_i > 0\) captures a fixed proportion in which inputs must be combined. For a consumer, their Leontief utility function is of the form:
\[
  u(\mathbf{q}) 
  \;=\; \min\biggl\{\tfrac{q_1}{\gamma_1},\; \tfrac{q_2}{\gamma_2},\;\dots,\;\tfrac{q_n}{\gamma_n}\biggr\}.
\]
This implies goods are perfect complements: the consumer gains utility only when goods are consumed in the specific ratio \(\gamma_1 : \gamma_2 : \ldots : \gamma_n\). In production, perfect complementarity imposes that a shortage of any one input 
strictly limits total output.

\subsection{Linear}

The \emph{linear} family is the simplest and assumes perfect substitutability. For a producer, the linear production function with inputs \(\mathbf{x}\) takes the form:
\[
  f(\mathbf{x}) 
  \;=\; b_1 x_1 + b_2 x_2 + \dots + b_n x_n,
\]
where \(b_i \geq 0\). This means each input contributes additively (and independently) to output. A consumer's linear utility function with goods \(\mathbf{q}\) is:
\[
  u(\mathbf{q}) 
  \;=\; \theta_1 q_1 + \theta_2 q_2 + \dots + \theta_n q_n,
\]
where \(\theta_i > 0\) captures the marginal utility for good \(i\). 
In both contexts, linear forms imply a constant rate of technical (or preferential)  substitution, reflecting strong substitutability among inputs (or goods).

\subsection{Non-Linear}

Beyond the classic forms, we use low-degree polynomial specifications 
for both production and utility functions. Such functions can capture nonlinearities 
without resorting to strictly Cobb-Douglas or other functional families. For a producer with inputs \(\mathbf{x}\), a second-degree polynomial looks like:
\[
  f(\mathbf{x}) 
  \;=\; \alpha_0 \;+\; \sum_{i=1}^n \alpha_i \, x_i 
        \;+\; \sum_{i=1}^n \sum_{j=1}^n \alpha_{ij} \, x_i \, x_j,
\]
where \(\alpha_0,\alpha_i,\alpha_{ij}\) are specified parameters. 
Similarly, for a consumer’s utility function, we write:
\[
  u(\mathbf{q}) 
  \;=\; \beta_0 \;+\; \sum_{i=1}^n \beta_i \, q_i 
        \;+\; \sum_{i=1}^n \sum_{j=1}^n \beta_{ij} \, q_i \, q_j.
\]
These forms allow more flexible curvature than linear or Cobb-Douglas, but remain 
computationally tractable for low degrees. The main difference in interpretation is that producers typically focus on \textit{return-to-scale} questions or  \textit{marginal products}, while consumers focus on \textit{marginal utilities} and \textit{marginal rates of substitution}.

\section{Examples}

\subsection{GPT-4o Aggregation of Consumer Demand (Shown)}\label{app:agg_consumer_demand_example}
\begin{tcolorbox}[colback=black!5!white,colframe=black!5!white,title=User Message:,coltitle=black,left=5pt,toptitle=0.1mm,fonttitle=\bfseries]
\fontsize{9pt}{10pt}\selectfont
Q: As the owner of a premium bedding brand, I cater to a diverse group of customers. I've discovered that there are $1204$ potential buyers for my luxury bedding products. Out of these, $820$ shoppers in urban markets each follow the demand function $18.1 - 1.51P$, while $384$ shoppers in suburban markets each adhere to the demand function $75.44 - 8.68P$. How can I calculate the total demand for my bedding at a price point of $8.4$?
\begin{enumerate}[label=\Alph*., wide, labelwidth=!,labelindent=0.5pt]
    \item $2779.36$
    \item $1832.01$
    \item $5411.87$ \emph{(Correct Answer)}
    \item $5050.03$
\end{enumerate}

Let's think step by step. Explain your reasoning.
\end{tcolorbox}

\begin{tcolorbox}[colback=blue!5!white,colframe=blue!5!white,title=GPT-4o Message:,coltitle=black,left=5pt,toptitle=0.1mm,fonttitle=\bfseries]
\fontsize{9pt}{10pt}\selectfont
To calculate the total demand for your luxury bedding products at a price point of $8.4$, we need to evaluate the demand functions for both urban and suburban markets separately and then sum the results.

\textbf{Step 1: Calculate the demand in the urban market}

The demand function for the urban market is given by:
\[ Q_u = 18.1 - 1.51P \]

Substitute \( P = 8.4 \) into the urban demand function:
\[ Q_u = 18.1 - 1.51 \times 8.4 \]
\[ Q_u = 18.1 - 12.684 \]
\[ Q_u = 5.416 \]

Since there are 820 shoppers in the urban market, the total demand in the urban market is:
\[ \text{Total Urban Demand} = 820 \times 5.416 \]
\[ \text{Total Urban Demand} = 4441.12 \]

\fontsize{9pt}{10pt}\selectfont
\textbf{Step 2: Calculate the demand in the suburban market}

The demand function for the suburban market is given by:
\[ Q_s = 75.44 - 8.68P \]

Substitute \( P = 8.4 \) into the suburban demand function:
\[ Q_s = 75.44 - 8.68 \times 8.4 \]
\[ Q_s = 75.44 - 72.912 \]
\[ Q_s = 2.528 \]

Since there are 384 shoppers in the suburban market, the total demand in the suburban market is:
\[ \text{Total Suburban Demand} = 384 \times 2.528 \]
\[ \text{Total Suburban Demand} = 970.88 \]

\textbf{Step 3: Calculate the total demand}

Now, sum the total demand from both markets:
\[ \text{Total Demand} = \text{Total Urban Demand} + \text{Total Suburban Demand} \]
\[ \text{Total Demand} = 4441.12 + 970.88 \]
\[ \text{Total Demand} = 5412 \]

Therefore, the total demand for your bedding at a price point of $8.4$ is approximately $5411.87$.

The correct answer is:

C. $5411.87$
\end{tcolorbox}

\subsection{GPT-4o Aggregation of Consumer Demand (Hidden)}

\begin{tcolorbox}[colback=black!5!white,colframe=black!5!white,title=User Message:,coltitle=black,left=5pt,toptitle=0.25mm,fonttitle=\bfseries]
\fontsize{9pt}{10pt}\selectfont
Q: As the owner of a premium bedding brand, I cater to a diverse group of customers. I've discovered that there are $1204$ potential buyers for my luxury bedding products. Out of these, $820$ shoppers in urban markets each follow the demand function $18.1 - 1.51P$, while $384$ shoppers in suburban markets each adhere to the demand function $75.44 - 8.68P$. How can I calculate the total demand for my bedding at a price point of $8.4$?

Let's think step by step. Explain your reasoning.
\end{tcolorbox}

\begin{tcolorbox}[colback=blue!5!white,colframe=blue!5!white,title=GPT-4o Message:,coltitle=black,left=5pt,toptitle=0.25mm,fonttitle=\bfseries]
\fontsize{9pt}{10pt}\selectfont
To calculate the total demand for your luxury bedding at a price point of 8.4, we need to evaluate
the demand functions for both urban and suburban markets separately and then sum the results. Here's
a step-by-step breakdown:

1. \textbf{Identify the demand functions:}

   - Urban market demand function: \( Q_u = 18.1 - 1.51P \)
   
   - Suburban market demand function: \( Q_s = 75.44 - 8.68P \)

2. \textbf{Substitute the price point (P = 8.4) into each demand function:}

   - For the urban market:
     \[
     Q_u = 18.1 - 1.51 \times 8.4
     \]
     Calculate the value:
     \[
     Q_u = 18.1 - 1.51 \times 8.4 = 18.1 - 12.684 = 5.416
     \]

   - For the suburban market:
     \[
     Q_s = 75.44 - 8.68 \times 8.4
     \]
     Calculate the value:
     \[
     Q_s = 75.44 - 8.68 \times 8.4 = 75.44 - 72.912 = 2.528
     \]

3. \textbf{Sum the demands from both markets to get the total demand:}
   \[
   Q_{total} = Q_u + Q_s = 5.416 + 2.528 = 7.944
   \]

Therefore, at a price point of 8.4, the total demand for your luxury bedding products from both
urban and suburban markets is approximately 7.944 units.

\end{tcolorbox}

\begin{tcolorbox}[colback=black!5!white,colframe=black!5!white,title=User Message:,coltitle=black,left=5pt,toptitle=0.25mm,fonttitle=\bfseries]
\fontsize{9pt}{10pt}\selectfont
\begin{enumerate}[label=\Alph*., wide, labelwidth=!,labelindent=0.5pt]
    \item $2779.36$
    \item $1832.01$
    \item $5411.87$ \emph{(Correct Answer)}
    \item $5050.03$
\end{enumerate}

Answer by writing the option letter corresponding to the correct option. WRITE ONLY A SINGLE LETTER.

A:
\end{tcolorbox}

\begin{tcolorbox}[colback=blue!5!white,colframe=blue!5!white,title=GPT-4o Message:,coltitle=black,left=5pt,toptitle=0.25mm,fonttitle=\bfseries]
\fontsize{9pt}{10pt}\selectfont
B
\end{tcolorbox}

\subsection{GPT-4o Profit Maximization}\label{app:profit_max_example}

\begin{tcolorbox}[colback=black!5!white,colframe=black!5!white,title=User Message:,coltitle=black,left=5pt,toptitle=0.25mm,fonttitle=\bfseries]
\fontsize{9pt}{10pt}\selectfont
Q: John manages a company that produces premium leather wallets. The production output depends on the capital (K) he invests, represented by the function $3.97K^{0.45}$. At the moment, the wallets are sold at a price of $3.64$, and John has an initial capital of $K_1 = 3.3$. He needs to decide how much additional capital to invest for the next day's market. John is aware of the potential price distribution for his wallets for the next day, represented by price $2.7$ with probability $0.24$, price $2.14$ with probability $0.42$, price $3.52$ with probability $0.34$. There is also a cost associated with increasing his capital, given by $(K_1-K_2)^2$, and he applies a discount factor of $0.37$ to future revenues. To maximize his profits, how much additional capital should John invest?
\begin{enumerate}[label=\Alph*., wide, labelwidth=!,labelindent=0.5pt]
    \item $0.44$
    \item $0.14$
    \item $1.1$
    \item $0.09$
\end{enumerate}
Let's think step by step. Explain your reasoning.
\end{tcolorbox}

\begin{tcolorbox}[colback=blue!5!white,colframe=blue!5!white,title=GPT-4o Message:,coltitle=black,left=5pt,toptitle=0.25mm,fonttitle=\bfseries]
\fontsize{9pt}{10pt}\selectfont
To determine how much additional capital John should invest to maximize his profits, we need to consider the expected revenue from the different potential prices, the cost of increasing capital, and the discount factor applied to future revenues. Let's break this down step by step.

\textbf{Step 1: Calculate the Expected Revenue}

First, we need to calculate the expected price of the wallets for the next day using the given probabilities.

\[
\text{Expected Price} = (2.7 \times 0.24) + (2.14 \times 0.42) + (3.52 \times 0.34)
\]

\[
\text{Expected Price} = 0.648 + 0.8988 + 1.1968 = 2.7436
\]

\textbf{Step 2: Determine the Production Function}

The production function is given by \(3.97K^{0.45}\). We need to calculate the production output for different levels of capital \(K\).

\textbf{Step 3: Calculate the Revenue for Different Capital Levels}

Let's denote the initial capital as \(K_1 = 3.3\). We need to consider the additional capital \( \Delta K \) that John might invest. The new capital \(K_2\) will be \(K_1 + \Delta K\).

The revenue function for a given capital \(K\) is:

\[
\text{Revenue} = \text{Expected Price} \times \text{Production Output}
\]

\[
\text{Revenue} = 2.7436 \times 3.97K^{0.45}
\]

\textbf{Step 4: Calculate the Cost of Increasing Capital}

The cost of increasing capital is given by \((K_1 - K_2)^2\). Since \(K_2 = K_1 + \Delta K\), the cost becomes:

\[
\text{Cost} = (\Delta K)^2
\]

\textbf{Step 5: Calculate the Discounted Future Revenue}

John applies a discount factor of 0.37 to future revenues. Therefore, the discounted future revenue is:

\[
\text{Discounted Revenue} = 0.37 \times \text{Revenue}
\]

\textbf{Step 6: Calculate the Net Profit}

The net profit is the discounted revenue minus the cost of increasing capital:

\[
\text{Net Profit} = 0.37 \times (2.7436 \times 3.97K_2^{0.45}) - (\Delta K)^2
\]
\end{tcolorbox}

\begin{tcolorbox}[colback=blue!5!white,colframe=blue!5!white,title=GPT-4o Message (Continued):,coltitle=black,left=5pt,toptitle=0.25mm,fonttitle=\bfseries]
\fontsize{9pt}{10pt}\selectfont
\textbf{Step 7: Evaluate the Net Profit for Different Values of \(\Delta K\)}

We need to evaluate the net profit for the given options of additional capital investment:

Option A: \(\Delta K = 0.44\)
\[
K_2 = 3.3 + 0.44 = 3.74
\]
\[
\text{Production Output} = 3.97 \times 3.74^{0.45}
\]
\[
\text{Revenue} = 2.7436 \times 3.97 \times 3.74^{0.45}
\]
\[
\text{Discounted Revenue} = 0.37 \times \text{Revenue}
\]
\[
\text{Cost} = (0.44)^2 = 0.1936
\]
\[
\text{Net Profit} = 0.37 \times \text{Revenue} - 0.1936
\]

Option B: \(\Delta K = 0.14\)
\[
K_2 = 3.3 + 0.14 = 3.44
\]
\[
\text{Production Output} = 3.97 \times 3.44^{0.45}
\]
\[
\text{Revenue} = 2.7436 \times 3.97 \times 3.44^{0.45}
\]
\[
\text{Discounted Revenue} = 0.37 \times \text{Revenue}
\]
\[
\text{Cost} = (0.14)^2 = 0.0196
\]
\[
\text{Net Profit} = 0.37 \times \text{Revenue} - 0.0196
\]

Option C: \(\Delta K = 1.1\)
\[
K_2 = 3.3 + 1.1 = 4.4
\]
\[
\text{Production Output} = 3.97 \times 4.4^{0.45}
\]
\[
\text{Revenue} = 2.7436 \times 3.97 \times 4.4^{0.45}
\]
\[
\text{Discounted Revenue} = 0.37 \times \text{Revenue}
\]
\[
\text{Cost} = (1.1)^2 = 1.21
\]
\[
\text{Net Profit} = 0.37 \times \text{Revenue} - 1.21
\]

Option D: \(\Delta K = 0.09\)
\[
K_2 = 3.3 + 0.09 = 3.39
\]
\[
\text{Production Output} = 3.97 \times 3.39^{0.45}
\]
\[
\text{Revenue} = 2.7436 \times 3.97 \times 3.39^{0.45}
\]
\[
\text{Discounted Revenue} = 0.37 \times \text{Revenue}
\]
\[
\text{Cost} = (0.09)^2 = 0.0081
\]
\[
\text{Net Profit} = 0.37 \times \text{Revenue} - 0.0081
\]

\textbf{Step 8: Compare the Net Profits}

We need to compare the net profits for each option to determine which one maximizes John's profit. After calculating the net profits for each option, we find that:

- Option A: \(\Delta K = 0.44\) yields the highest net profit.

Therefore, the best option for John is to invest an additional capital of 0.44.

**Answer: A. 0.44**
\end{tcolorbox}

\begin{tcolorbox}[colback=black!5!white,colframe=black!5!white,title=User Message:,coltitle=black,left=5pt,toptitle=0.25mm,fonttitle=\bfseries]
\fontsize{9pt}{10pt}\selectfont
Answer by writing the option letter corresponding to the correct option. WRITE ONLY A SINGLE LETTER. 

A: 
\end{tcolorbox}

\begin{tcolorbox}[colback=blue!5!white,colframe=blue!5!white,title=GPT-4o Message:,coltitle=black,left=5pt,toptitle=0.25mm,fonttitle=\bfseries]
\fontsize{9pt}{10pt}\selectfont
A 
\end{tcolorbox}

\section{Figures}

\begin{shownto}{iclr}

    \begin{figure}[H]
        \centering
        \includegraphics[width=\linewidth]{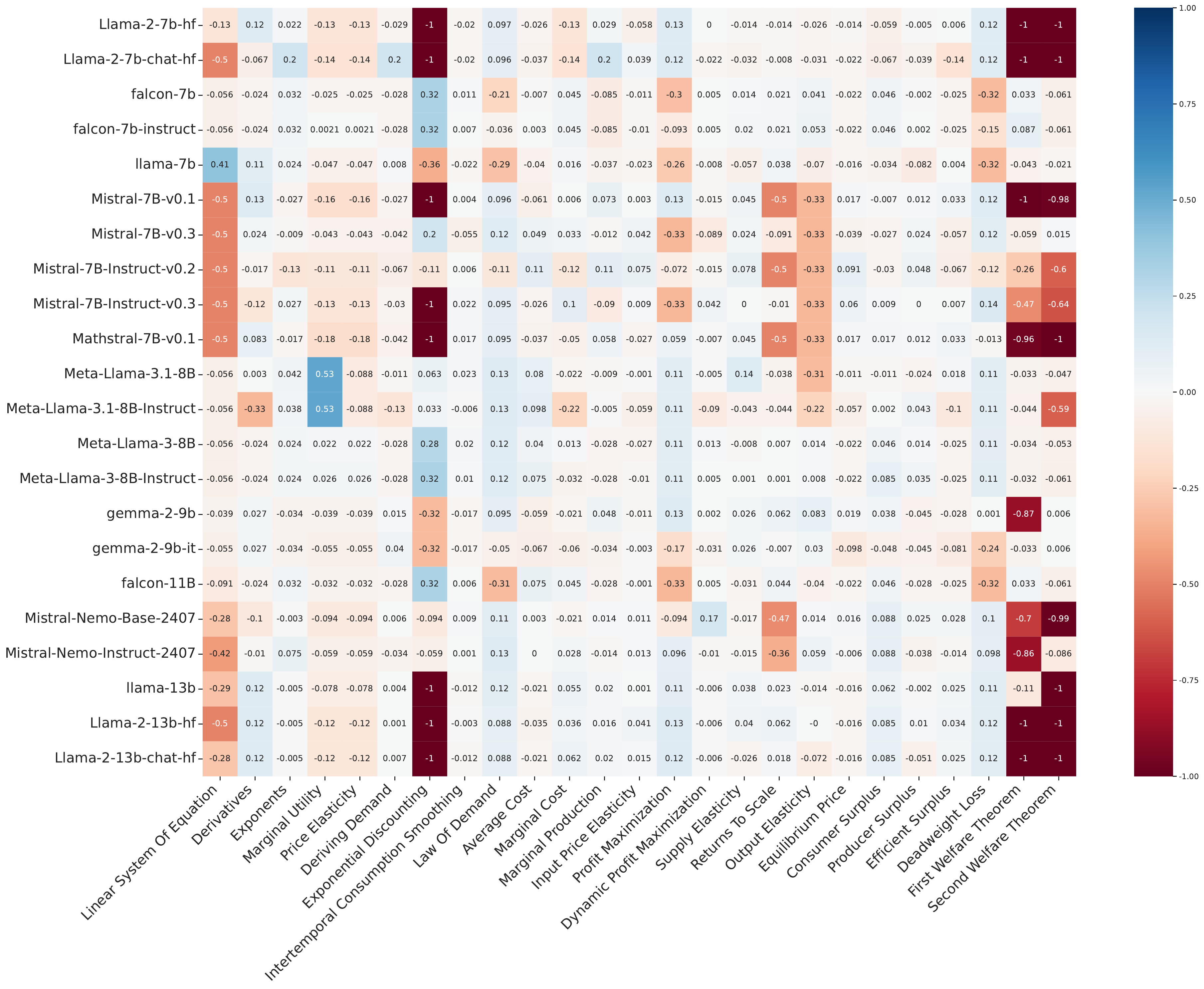}
        \caption{Heatmap of normalized accuracy of open-source models.}
        \label{fig:small_model_heatmap}
    \end{figure}

    \begin{figure}[H]
        \centering
        \includegraphics[width=\linewidth]{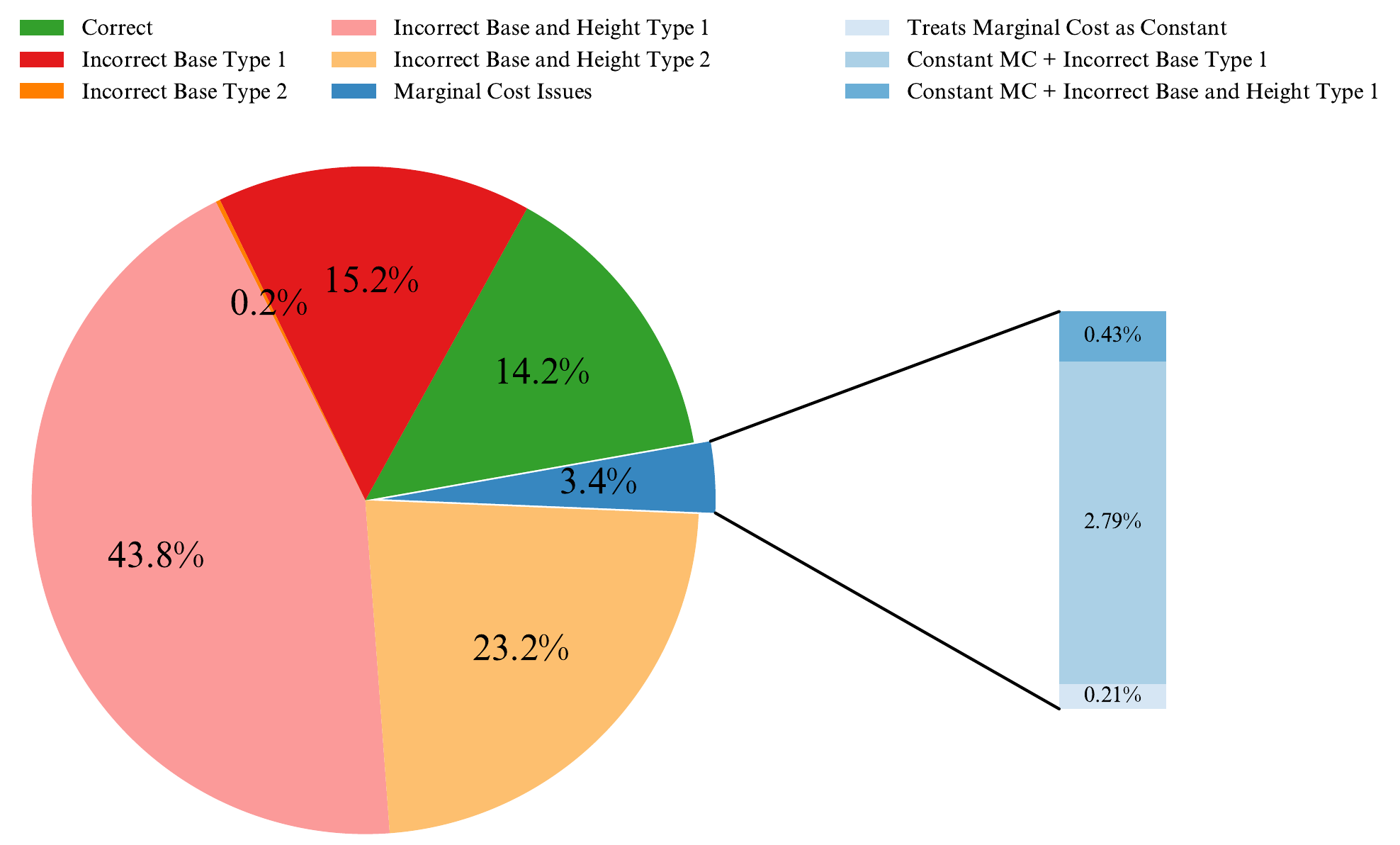}
        \caption{Error analysis of claude-3-5-sonnet on the \nameref{el:deadweight_loss} element. We further breakdown the errors by incorrectly interpreting the marginal cost.}
        \label{fig:sonnet_dwl}
    \end{figure}
    
    \begin{figure}[H]
        \centering
        \includegraphics[width=\linewidth]{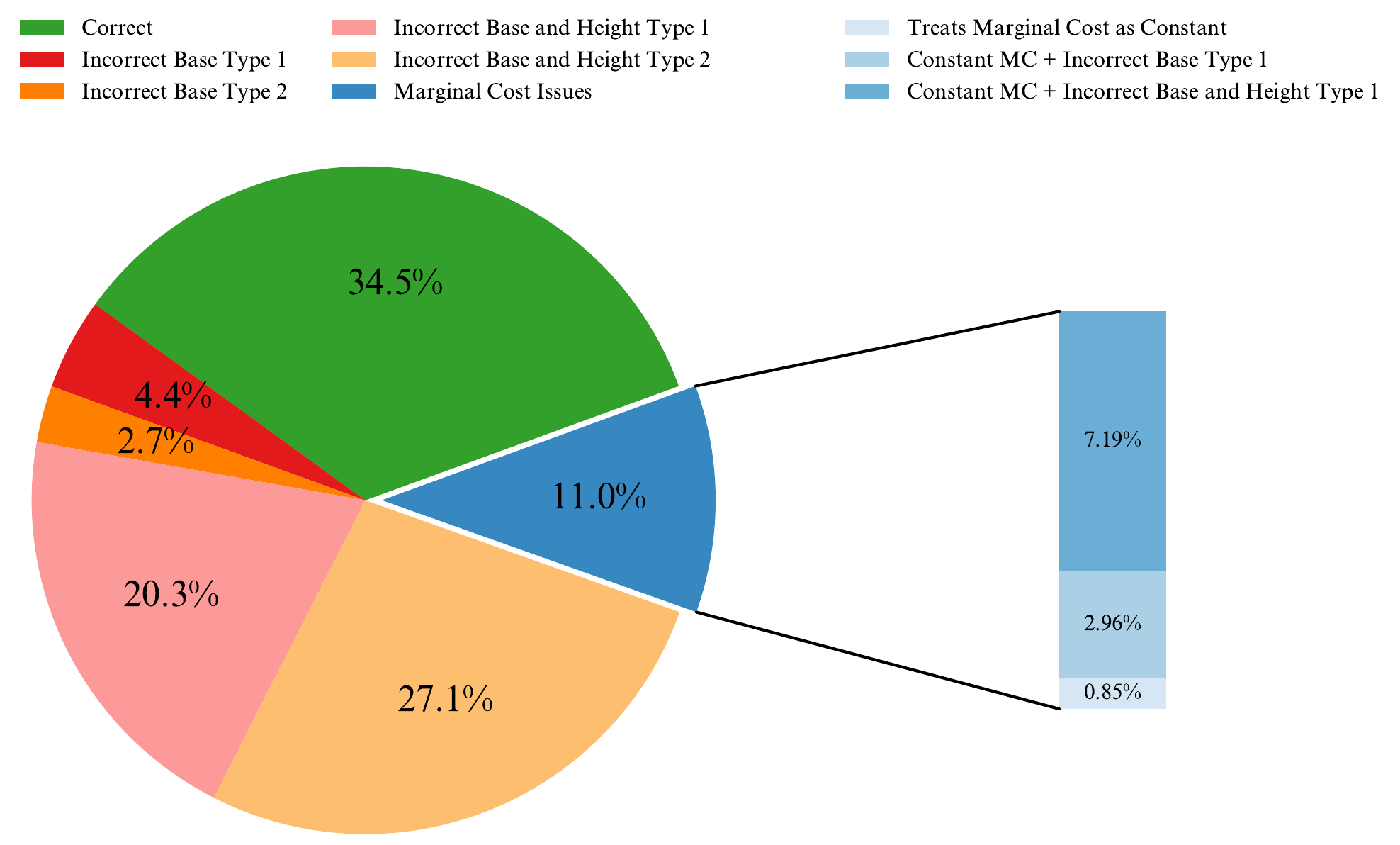}
        \caption{Error analysis of gpt-4o on the \nameref{el:deadweight_loss} element. We further breakdown the errors by incorrectly interpreting the marginal cost.}
        \label{fig:4o_dwl}
    \end{figure}

    \begin{figure}[H]
        \centering
        \includegraphics[width=\linewidth]{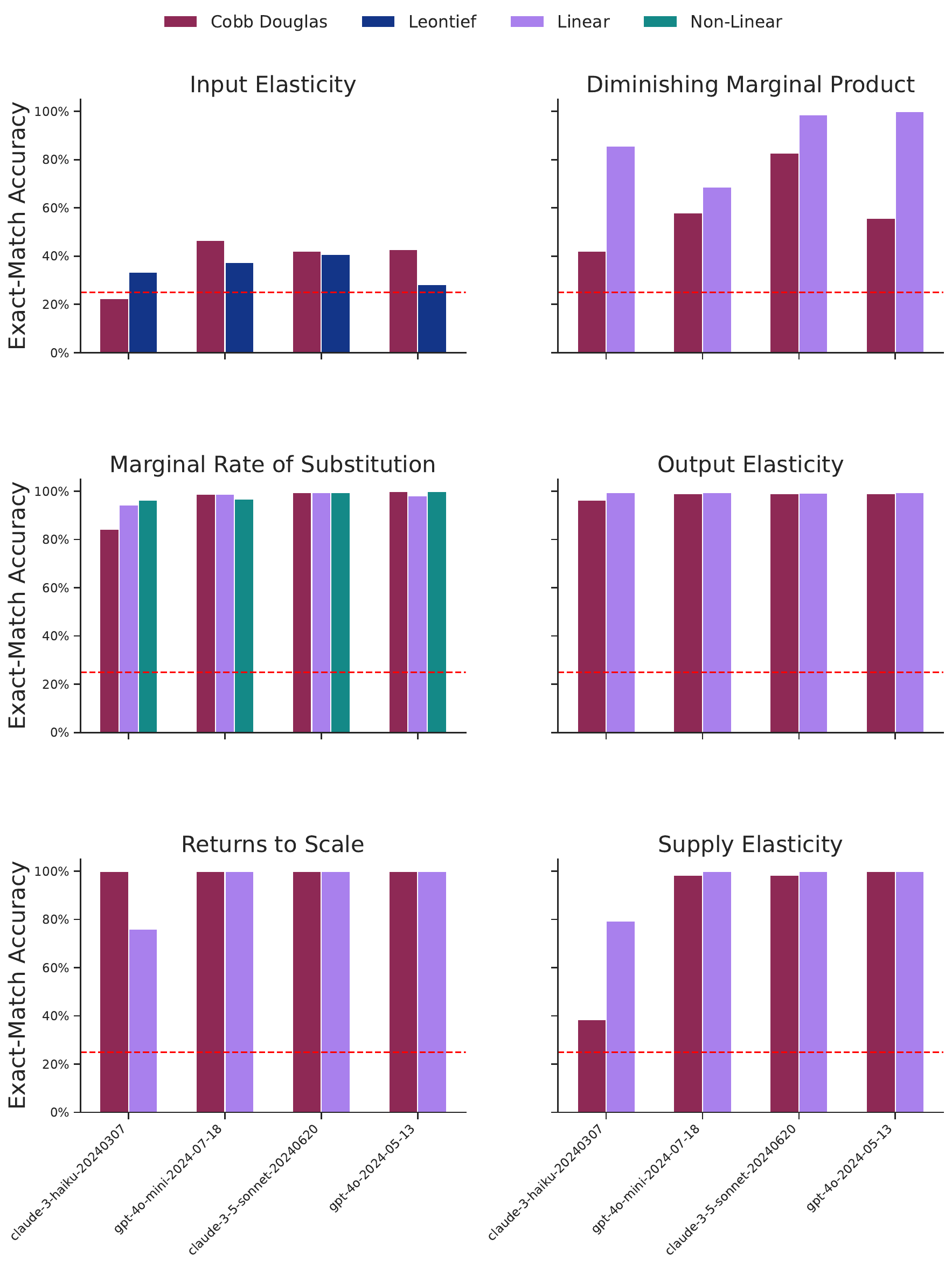}
        \caption{Exact-match accuracy of all closed-source models across six elements and four functional families (Cobb-Douglas, Leontief, Linear, and Non-Linear). The results demonstrate varying type robustness, with Cobb-Douglas being a generally more challenging functional family but not consistently harder for all elements. For instance, accuracy remains high for elements such as \nameref{el:output_elasticity} and \nameref{el:mrs_utility}, even on the Cobb-Douglas functions, while elements like \nameref{el:input_elasticity} and \nameref{el:returns_to_scale} show more variability across functional types. The red dashed line indicates the random guessing baseline for comparison.}
        \label{fig:type_robustness}
    \end{figure}

    \begin{figure}[H]
        \centering
        \includegraphics[trim={1cm, 0, 0, 0},clip,width=\linewidth]{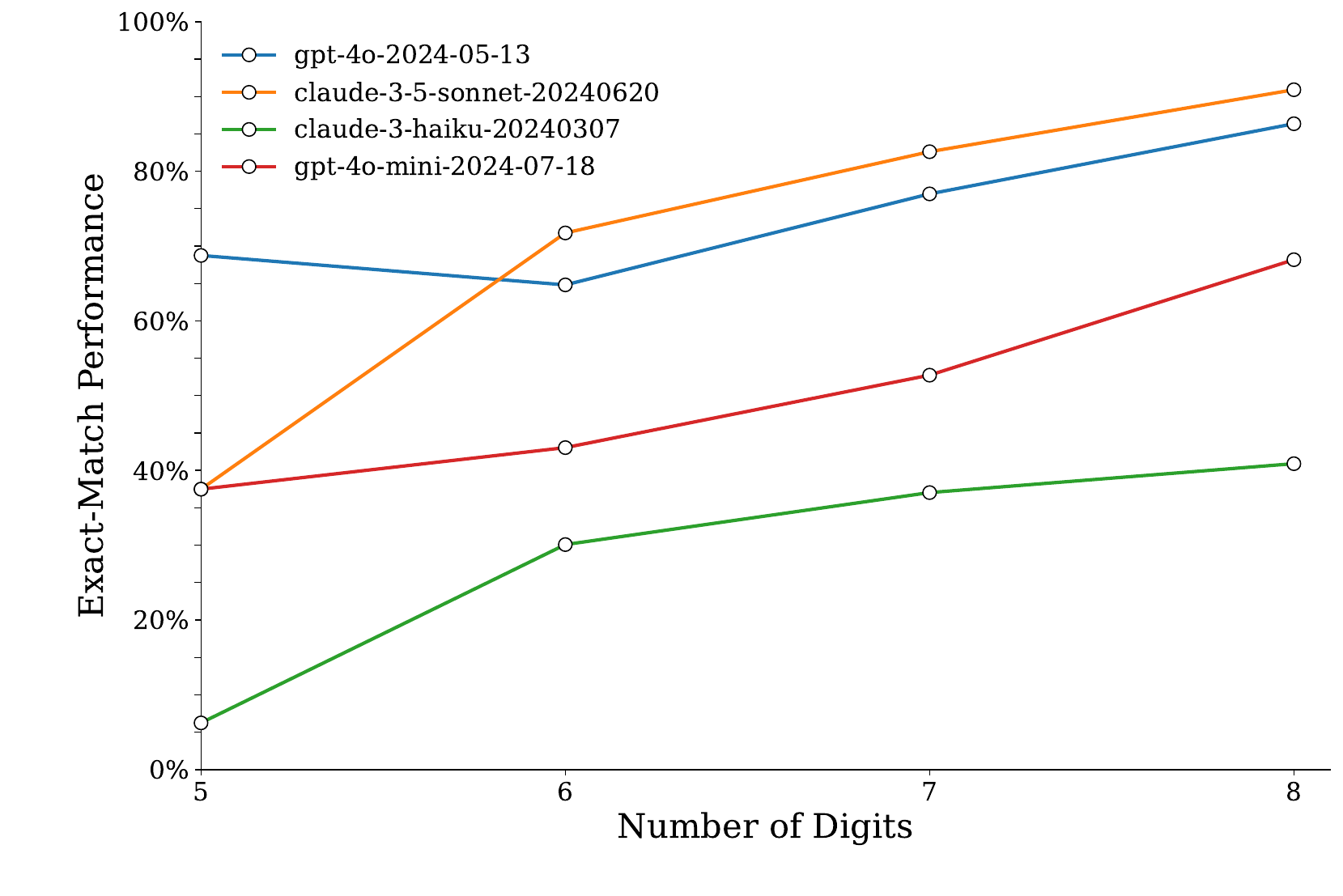}
        \caption{This figure depicts exact-match performance on the \nameref{el:agg_consumer_demand} element on the \shown implementation of 0-CoT for the closed-source models against the number of digits of the correct answer.}
      \label{fig:agg_consumer_demand}
      \begin{shownto}{arxiv}
        \Description[]{This figure depicts exact-match performance on the \nameref{el:agg_consumer_demand} element on the \shown implementation of 0-CoT for the closed-source models against the number of digits of the correct answer.}
    \end{shownto}
    \end{figure}
\end{shownto}
                             
\begin{figure}[H]
    \centering
    \includegraphics[width=\linewidth]{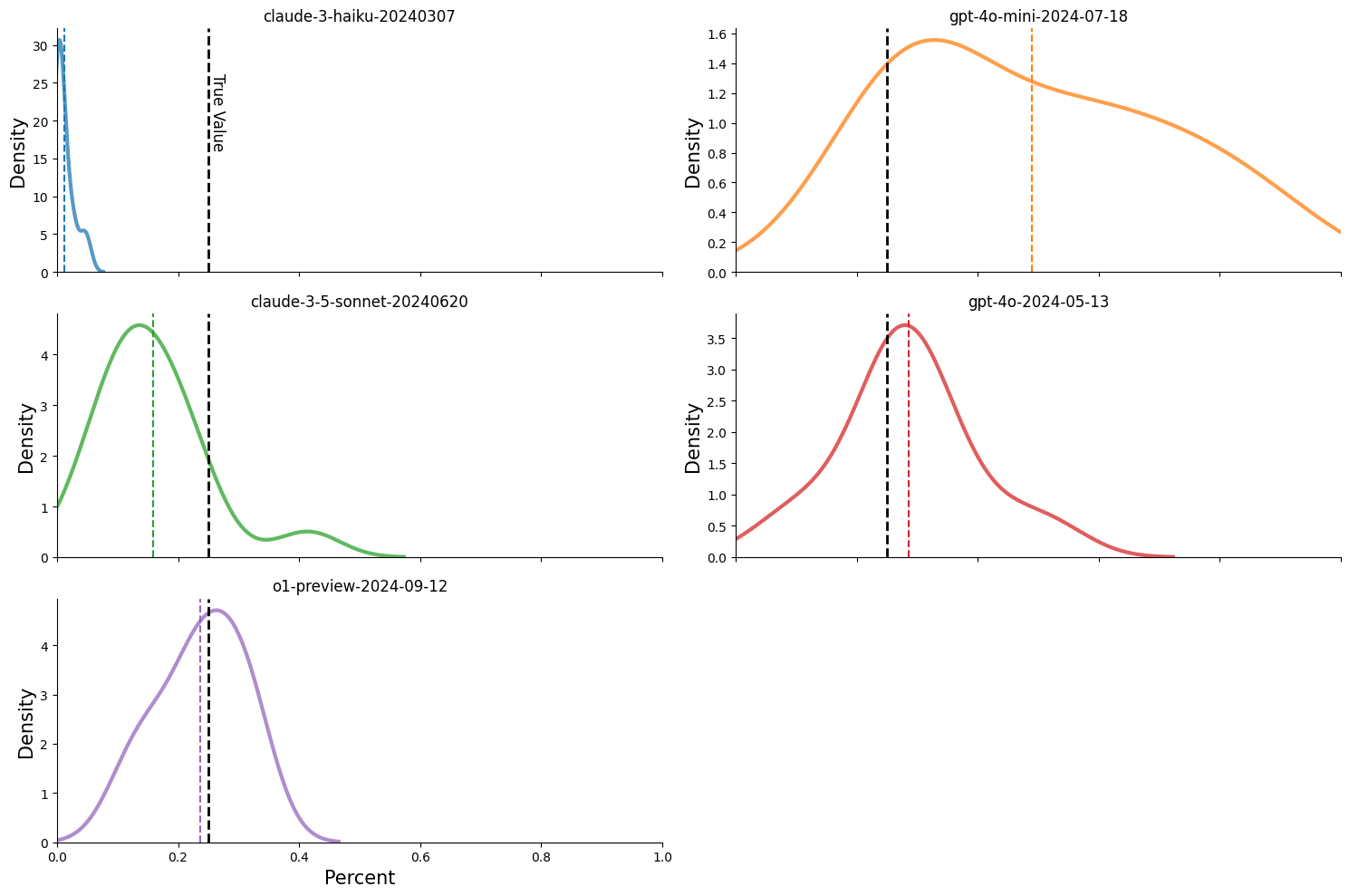}
    \caption{This figure plots the distribution of selection rates of ``No other option is correct'' (NOTA) element-by-element. The dotted black line represents the correct frequency of selecting NOTA. We see that Anthropic models rarely select NOTA, whereas the non-reasoning OpenAI models over-select NOTA.}
    \label{fig:nota_frequency}
    \begin{shownto}{arxiv}
        \Description[]{This figure plots the distribution of selection rates of ``No other option is correct'' (NOTA) element-by-element. The dotted black line represents the correct frequency of selecting NOTA. We see that Anthropic models rarely select NOTA, whereas the non-reasoning OpenAI models over-select NOTA.}
    \end{shownto}
\end{figure}

\pagebreak

\section{Analysis of Rephrasing Variance}\label{app:rephrasing_var}
To understand the role of question rephrasings in our dataset, we conducted an analysis of variance (ANOVA) on all other controllable features. These features include \qtype, \domain, and \tags. The goal of this analysis was to quantify the variance in \model performance attributable to these features and, by exclusion, infer the contribution of rephrasings to the remaining unexplained variance.

The results for the top-performing models, summarized in \Cref{tab:anova_o1-preview-2024-09-12} through \Cref{tab:anova_claude-3-haiku-20240307}, indicate that the explained variance attributable to the controlled features is consistently low across all evaluated models. This leaves approximately 56\% (for claude-3-5-sonnet) and up to 91\% (for o1-preview) of the variance unexplained by the features included in the analysis. Given that question rephrasings are a systematic element of our dataset design and were not included as a feature in this analysis, we infer that the majority of this residual variance is due to differences in how models respond to semantically equivalent but syntactically varied prompts.

\begin{table}[ht]
\centering
\begin{tabular}{lrrrr}
\toprule
Factor & Sum of Squares & Degrees of Freedom & F-Statistic & p-value \\
\midrule
\domain & 7.5572 & 11.0000 & 3.0199 & 0.0823 \\
\tags & 3.4351 & 5.0000 & 3.0199 & 0.0823 \\
CAR & 21.1235 & 1.0000 & 92.8524 & 0.0000 \\
element:\qtype & 178.6251 & 260.0000 & 3.0199 & 0.0823 \\
Residual & 2218.5356 & 9752.0000 &  &  \\
\cline{3-5}\noalign{\vspace{0.5ex}}
& & \multicolumn{1}{l}{\emph{R-squared}} & & 0.0941 \\
& & \multicolumn{1}{l}{\emph{Adjusted R-squared}} & & 0.0907 \\
\end{tabular}
\caption{ANOVA Results for o1-preview-2024-09-12}
\label{tab:anova_o1-preview-2024-09-12}
\end{table}

\begin{table}[ht]
\centering
\begin{tabular}{lrrrr}
\midrule
Factor & Sum of Squares & Degrees of Freedom & F-Statistic & p-value \\
\midrule
\domain & 0.2619 & 11.0000 & 0.0928 & 0.7607 \\
\tags & 0.1190 & 5.0000 & 0.0928 & 0.7607 \\
0-CoT & 229.7295 & 1.0000 & 895.0635 & 0.0000 \\
CAR & 141.3015 & 1.0000 & 550.5338 & 0.0000 \\
element:\qtype & 47.4945 & 1995.0000 & 0.0928 & 0.7607 \\
Residual & 16015.5015 & 62399.0000 &  &  \\
\cline{3-5}\noalign{\vspace{0.5ex}}
& & \multicolumn{1}{l}{\emph{R-squared}} & & 0.3368 \\
& & \multicolumn{1}{l}{\emph{Adjusted R-squared}} & & 0.3358 \\
\end{tabular}
\caption{ANOVA Results for gpt-4o-2024-05-13}
\label{tab:anova_gpt-4o-2024-05-13}
\end{table}

\begin{table}[ht]
\centering
\begin{tabular}{lrrrr}
\toprule
Factor & Sum of Squares & Degrees of Freedom & F-Statistic & p-value \\
\midrule
\domain & 0.4573 & 11.0000 & 0.1293 & 0.7191 \\
\tags & 0.2079 & 5.0000 & 0.1293 & 0.7191 \\
0-Cot & 34.0324 & 1.0000 & 105.8911 & 0.0000 \\
CAR & 159.1899 & 1.0000 & 495.3161 & 0.0000 \\
element:type & 82.9338 & 1995.0000 & 0.1293 & 0.7191 \\
Residual & 18642.5829 & 58006.0000 &  &  \\
\cline{3-5}\noalign{\vspace{0.5ex}}
& & \multicolumn{1}{l}{\emph{R-squared}} & & 0.2964 \\
& & \multicolumn{1}{l}{\emph{Adjusted R-squared}} & & 0.2953 \\
\end{tabular}
\caption{ANOVA Results for gpt-4o-mini-2024-07-18}
\label{tab:anova_gpt-4o-mini-2024-07-18}
\end{table}

\begin{table}[ht]
\centering
\begin{tabular}{lrrrr}
\toprule
Factor & Sum of Squares & Degrees of Freedom & F-Statistic & p-value \\
\midrule
\domain & 1.2389 & 11.0000 & 0.5004 & 0.6063 \\
\tags & 0.5032 & 5.0000 & 0.4472 & 0.5037 \\
0-CoT & 30.9486 & 1.0000 & 137.5118 & 0.0000 \\
CAR & 156.8135 & 1.0000 & 696.7577 & 0.0000 \\
element:\qtype & 224.0990 & 1995.0000 & 0.4991 & 0.6071 \\
Residual & 18025.1941 & 80090.0000 &  &  \\
\cline{3-5}\noalign{\vspace{0.5ex}}
& & \multicolumn{1}{l}{\emph{R-squared}} & & 0.4436 \\
& & \multicolumn{1}{l}{\emph{Adjusted R-squared}} & & 0.4430 \\
\end{tabular}
\caption{ANOVA Results for claude-3-5-sonnet-20240620}
\label{tab:anova_claude-3-5-sonnet-20240620}
\end{table}

\begin{table}[ht]
\centering
\begin{tabular}{lrrrr}
\toprule
Factor & Sum of Squares & Degrees of Freedom & F-Statistic & p-value \\
\midrule
\domain & 1.4703 & 11.0000 & 0.3814 & 0.5369 \\
\tags & 0.6683 & 5.0000 & 0.3814 & 0.5369 \\
0-Cot & 0.2629 & 1.0000 & 0.7502 & 0.3864 \\
CAR & 0.4909 & 1.0000 & 1.4007 & 0.2366 \\
element:\qtype & 266.6568 & 1995.0000 & 0.3814 & 0.5369 \\
Residual & 38842.9909 & 110826.0000 &  &  \\
\cline{3-5}\noalign{\vspace{0.5ex}}
& & \multicolumn{1}{l}{\emph{R-squared}} & & 0.2336 \\
& & \multicolumn{1}{l}{\emph{Adjusted R-squared}} & & 0.2330 \\
\end{tabular}
\caption{ANOVA Results for claude-3-haiku-20240307}
\label{tab:anova_claude-3-haiku-20240307}
\end{table}

\newpage

\section{Models}\label{appendix:models}
\begin{longtable}{>{\raggedright\arraybackslash}p{4cm} >{\raggedright\arraybackslash}p{6.5cm} >{\centering\arraybackslash}p{2cm}}
    \toprule
    \textbf{Model Name} & \textbf{Model Card} & \textbf{Chat/ Instruction Tuned} \\
    \midrule
    \endfirsthead

    \toprule
    \textbf{Model Name} & \textbf{Model Card} & \textbf{Chat/Instruction Tuned} \\
    \midrule
    \endhead

    \midrule
    \multicolumn{3}{r}{\textit{Continued on next page}} \\
    \midrule
    \endfoot

    \bottomrule
    \caption{Overview of the open- and closed-source \models we evaluated. The table includes their names, their model card links, and whether they have been chat or instruction tuned. Models are grouped by family and sorted by parameter size, with non-chat-tuned models listed first within each group.}
    \endlastfoot
    \textbf{Closed-Source} & & \\
    \cline{1-2}
    \addlinespace[1ex]
    \emph{OpenAI}\\
    \cline{1-1}
    \addlinespace[0.5ex]
    gpt-4o & & \checkmark \\
    gpt-4o mini & & \checkmark \\

    \addlinespace[1ex]
    \emph{Anthropic}\\
    \cline{1-1}
    \addlinespace[0.5ex]
    claude-3-5-sonnet & & \checkmark \\
    claude-3-haiku & & \checkmark \\

    
    
    \addlinespace[0.5ex]
    \emph{Huggy Llama} & & \\
    \cline{1-1}
    \addlinespace[0.5ex]
    llama-7b & \href{https://huggingface.co/huggyllama/llama-7b}{huggyllama/llama-7b} & $\times$ \\
    llama-13b & \href{https://huggingface.co/huggyllama/llama-13b}{huggyllama/llama-13b} & $\times$ \\
    llama-30b & \href{https://huggingface.co/huggyllama/llama-30b}{huggyllama/llama-30b} & $\times$ \\
    llama-65b & \href{https://huggingface.co/huggyllama/llama-65b}{huggyllama/llama-65b} & $\times$ \\

    \addlinespace[0.5ex]
    \emph{Meta Llama} & & \\
    \cline{1-1}
    \addlinespace[0.5ex]
    Llama-2-7b-hf & \href{https://huggingface.co/meta-llama/Llama-2-7b-hf}{meta-llama/Llama-2-7b-hf} & $\times$ \\
    Llama-2-13b-hf & \href{https://huggingface.co/meta-llama/Llama-2-13b-hf}{meta-llama/Llama-2-13b-hf} & $\times$ \\
    Llama-2-7b-chat-hf & \href{https://huggingface.co/meta-llama/Llama-2-7b-chat-hf}{meta-llama/Llama-2-7b-chat-hf} & \checkmark \\
    Llama-2-13b-chat-hf & \href{https://huggingface.co/meta-llama/Llama-2-13b-chat-hf}{meta-llama/Llama-2-13b-chat-hf} & \checkmark \\

    Llama-3-8B & \href{https://huggingface.co/meta-llama/Meta-Llama-3-8B}{meta-llama/Meta-Llama-3-8B} & $\times$ \\
    Llama-3-8B-Instruct & \href{https://huggingface.co/meta-llama/Meta-Llama-3-8B-Instruct}{meta-llama/Meta-Llama-3-8B-Instruct} & \checkmark \\

    Llama-3.1-8B & \href{https://huggingface.co/meta-llama/Meta-Llama-3.1-8B}{meta-llama/Meta-Llama-3.1-8B} & $\times$ \\
    Llama-3.1-70B & \href{https://huggingface.co/meta-llama/Meta-Llama-3.1-70B}{meta-llama/Meta-Llama-3.1-70B} & $\times$ \\
    Llama-3.1-70B-Instruct & \href{https://huggingface.co/meta-llama/Meta-Llama-3.1-70B-Instruct}{meta-llama/Meta-Llama-3.1-70B-Instruct} & \checkmark \\

    \addlinespace[0.5ex]
    \emph{Mistral} & & \\
    \cline{1-1}
    \addlinespace[0.5ex]
    Mistral-7B-v0.1 & \href{https://huggingface.co/mistralai/Mistral-7B-v0.1}{mistralai/Mistral-7B-v0.1} & $\times$ \\
    Mathstral-7B-v0.1 & \href{https://huggingface.co/mistralai/Mathstral-7B-v0.1}{mistralai/Mathstral-7B-v0.1} &$\times$ \\
    Mistral-7B-v0.3 & \href{https://huggingface.co/mistralai/Mistral-7B-v0.3}{mistralai/Mistral-7B-v0.3} & $\times$ \\
    Mistral-7B-Instruct-v0.3 & \href{https://huggingface.co/mistralai/Mistral-7B-Instruct-v0.3}{mistralai/Mistral-7B-Instruct-v0.3} & \checkmark \\
    Mistral-Nemo-Base-2407 (12.2B) & \href{https://huggingface.co/mistralai/Mistral-Nemo-Base-2407}{mistralai/Mistral-Nemo-Base-2407} &$\times$ \\
    Mistral-Nemo-Instruct-2407 (12.2B) & \href{https://huggingface.co/mistralai/Mistral-Nemo-Instruct-2407}{mistralai/Mistral-Nemo-Instruct-2407} & \checkmark \\

    \addlinespace[0.5ex]
    \emph{TIIUAE} & & \\
    \cline{1-1}
    \addlinespace[0.5ex]
    falcon-7B & \href{https://huggingface.co/tiiuae/falcon-7b}{tiiuae/falcon-7b} & $\times$ \\
    falcon-11B & \href{https://huggingface.co/tiiuae/falcon-11B}{tiiuae/falcon-11B} & $\times$ \\
    
    \addlinespace[0.5ex]
    \emph{AI21} & & \\
    \cline{1-1}
    \addlinespace[0.5ex]
    Jamba-v0.1 & \href{https://huggingface.co/ai21labs/Jamba-v0.1}{ai21labs/Jamba-v0.1} & $\times$ \\
    AI21-Jamba-1.5-Mini & \href{https://huggingface.co/ai21labs/AI21-Jamba-1.5-Mini}{ai21labs/AI21-Jamba-1.5-Mini} & $\times$ \\

    \label{tbl:models}
\end{longtable}


\newpage 

\section{Extra Results}\label{app:extra_results}

\subsection{Performance on Elements Generated by Claude 3.5 Sonnet}\label{app:claude_elements}
To assess whether the performance on our dataset was influenced by the choice of the generation \model, we re-generated three elements from scratch using claude-3-5-sonnet. We selected \nameref{el:find_eq_price} because it exhibited the largest performance gap between gpt-4o and claude-3-5-sonnet, \nameref{el:diminishing_marginal_product} was chosen as a random element with slight performance variation across the models, and \nameref{el:price_elasticity_demand} served as a control where no significant differences were expected.

\begin{figure}[H]
    \centering
    \includegraphics[width=\linewidth]{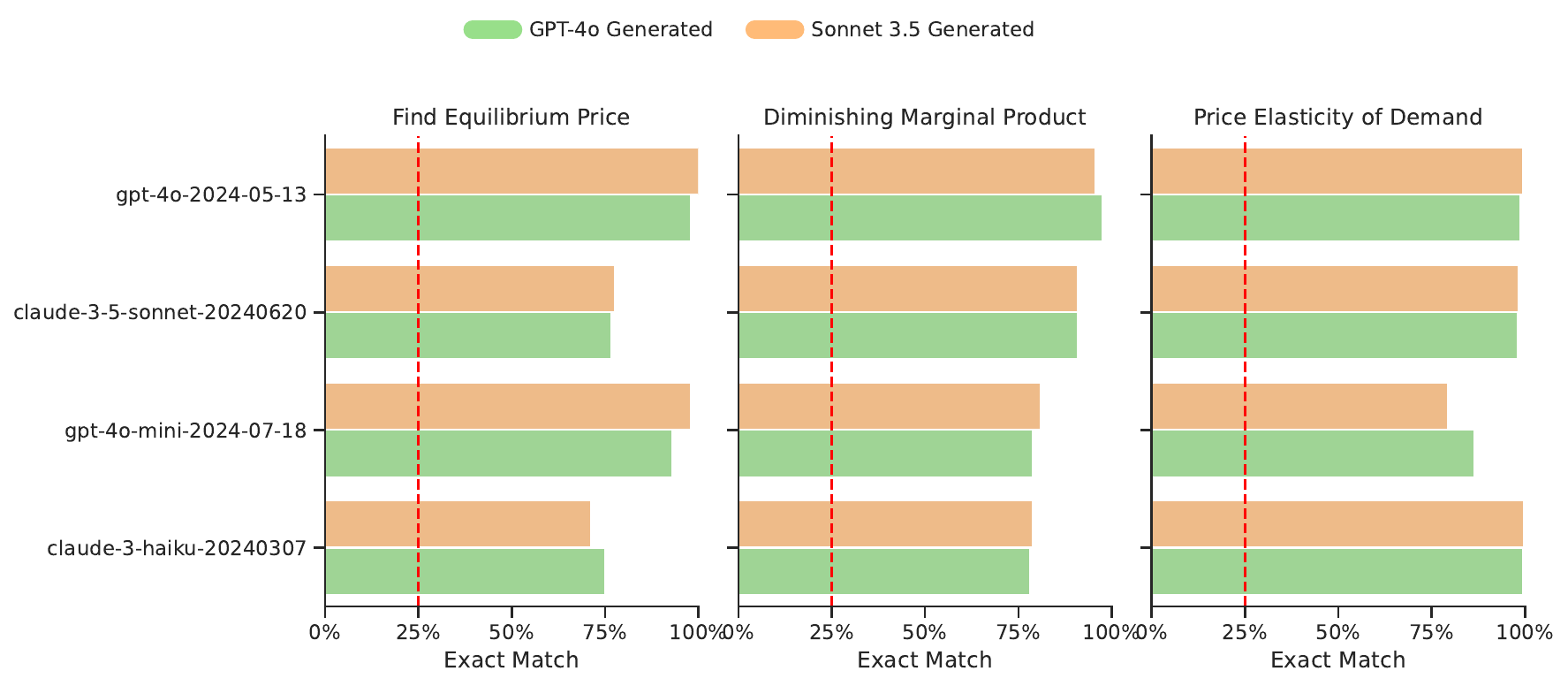}
    \caption{Exact-match performance comparison between closed-source models on three that were generated by gpt-4o and claude-3-5-sonnet. The elements were selected based on the observed performance differences across models, with \nameref{el:price_elasticity_demand} serving as a control. The figure shows no significant differences in performance between the models on these elements. Note that the red dotted line signifies random guessing performance.}
    \begin{shownto}{arxiv}
        \Description[]{Exact-match performance comparison between closed-source models on three that were generated by gpt-4o and claude-3-5-sonnet. The elements were selected based on the observed performance differences across models, with \nameref{el:price_elasticity_demand} serving as a control. The figure shows no significant differences in performance between the models on these elements. Note that the red dotted line signifies random guessing performance.}
    \end{shownto}
    \label{fig:claude_comparison}
\end{figure}

\Cref{fig:claude_comparison} shows the exact-match performance of both models on these three elements. We found no significant differences in performance between any of the models.

\newpage

\subsection{Intertemporal Consumption Smoothing}\label{app:intertemporal_consumption_smoothing}

When optimizing intertemporal consumption, the consumer maximizes the discounted utility
\[
\sum_{t=0}^T \beta^t u(c_t)
\]
subject to the intertemporal budget constraint. The first-order condition for an optimum leads to
the Euler equation:
\[
u'(c_t) = \beta (1+r) \, u'(c_{t+1}).
\]

For our purposes, we tested models using a constant relative risk aversion (CRRA) utility function. We used the following form:
\[
u(c) = \frac{c^{1-\gamma}}{1-\gamma} \quad (\gamma \neq 1),
\]
where \(\gamma\) is the coefficient of relative risk aversion. This function exhibits diminishing marginal utility, meaning that each additional unit of consumption adds less to utility when overall consumption is high compared to when it is low. Due to diminishing returns, an agent is motivated to smooth consumption over time, even with a discount factor of $1$; spending too much in one period reduces the marginal utility in that period, while having too little in another period results in a steep loss of satisfaction.

For CRRA utility, the Euler equation is given by:
\[
u'(c_t) = \beta (1+r) \, u'(c_{t+1}),
\]
where
\[
u'(c) = c^{-\gamma}.
\]
Rearranging yields:
\[
c_{t+1} = \left[\beta (1+r)\right]^{\frac{1}{\gamma}} c_t.
\]

However, we found that \models would often use linear utility functions in their analysis. For a linear utility function, the Euler equation---which equates the marginal benefit of consuming today with that of consuming tomorrow---simplifies significantly. If \( u'(c) \) is constant (say, equal to 1), then aside from the effects of discounting and interest, there is no curvature-driven motive to adjust consumption levels across periods. The optimal allocation would then depend solely on the intertemporal budget constraint and the returns on savings.

\subsection{Deadweight Loss}\label{app:deadweight_loss}
To conduct our error analysis, we ran all closed-source models on the free-text QA adaptation of the deadweight loss task. We began by inspecting a range of model outputs to identify distinct classes of errors that were common across responses. Once these error categories were established, we computed the answers corresponding to these errors and rescored the models based on whether their outputs were within 98\% of either the correct answer or any of the answers derived from specific error assumptions. We also ensured that when there was any overlap in incorrect responses that we chose the closest one to the model's response. This approach allowed us to capture not only the frequency of correct outputs but also the systematic nature of the models' reasoning flaws. Below, we provide a detailed breakdown of the primary error types: 

\begin{itemize}

    \item Incorrect Base for Deadweight Loss Type 1: This error incorrectly substitutes \( P_e - P_m \) (the difference between the competitive equilibrium price and the monopolist's price) in place of the correct term \( P_m - MC(Q_m) \) (the difference between the monopolist's price and the marginal cost at the monopolist's quantity). 
    
    \item Incorrect Base for Deadweight Loss Type 2: This error calculates the deadweight loss using the difference between the monopoly price and the competitive equilibrium price as the base of the triangle. 

    \item Incorrect Base and Height Type 1: This error replaces the base of the DWL triangle (\( Q_e - Q_m \)) with a miscalculated value for the equilibrium quantity and replaces the base with the Type 1 variant. 

    \item Incorrect Base and Height Type 2: This error assumes that \( P_e = MC(Q_m) \), leading to an incorrect height calculation where the DWL triangle's height becomes \( P_m - P_e \). 

    \item Treating Marginal Cost as Constant: Instead of recognizing marginal cost as a function derived from the supply curve (\( MC(Q) = a_{\text{supply}} \times Q + b_{\text{supply}} \)), several models treated marginal cost as a constant, often equal to the slope of the supply curve. This assumption led to errors in determining the monopolist's quantity and price, further propagating inaccuracies in the deadweight loss calculation.

    \item Combined Errors: A subset of models combined the two errors above, simultaneously using an incorrect formula for deadweight loss and assuming a constant marginal cost. This compounded error significantly reduced the likelihood of producing a correct answer and highlighted the systematic nature of the misunderstanding.
\end{itemize}

\newpage

\subsection{Performance Gaps Between Free Text QA and Hidden MCQA}\label{app:free_vs_hidden}

\begin{figure}[H]
    \centering
    \includegraphics[width=\linewidth]{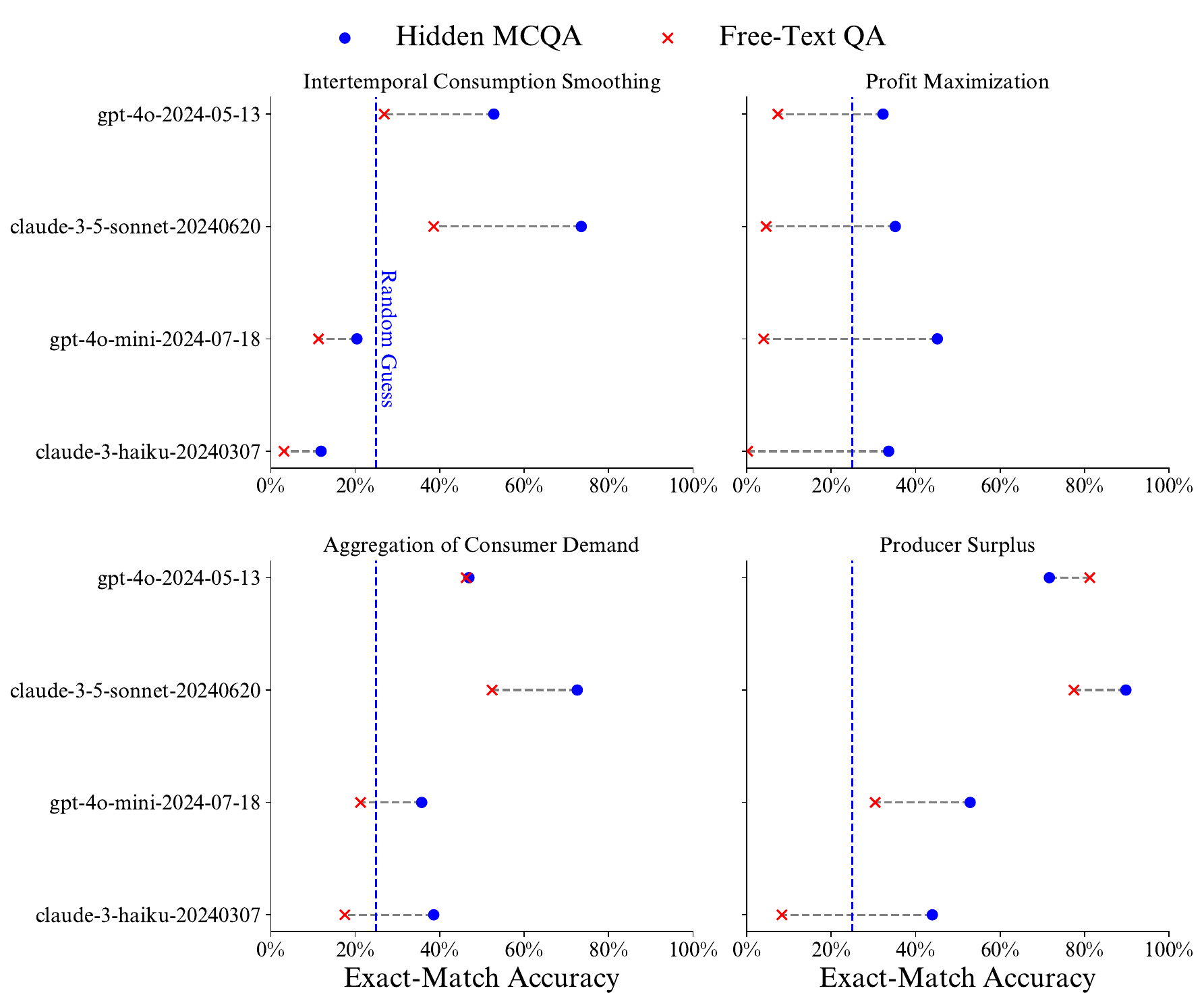}
    \caption{Comparison of exact-match accuracy for four elements (\nameref{el:intertemporal_consumption_smoothing}, \nameref{el:profit_max}, \nameref{el:agg_consumer_demand}, and \nameref{el:producer_surplus}) across all closed-source models. The plot illustrates the difference in performance under the hidden and free-text adaptations, highlighting the impact of multiple-choice options on reasoning accuracy. Plotted in blue is the accuracy random guessing would have in the hidden adaptation, note that this line is not relevant for the free-text QA adaptation. }
    \begin{shownto}{arxiv}
        \Description[]{Comparison of exact-match accuracy for four elements (\nameref{el:intertemporal_consumption_smoothing}, \nameref{el:profit_max}, \nameref{el:agg_consumer_demand}, and \nameref{el:producer_surplus}) across all closed-source models. The plot illustrates the difference in performance under the hidden and free-text adaptations, highlighting the impact of multiple-choice options on reasoning accuracy. Plotted in blue is the accuracy random guessing would have in the hidden adaptation, note that this line is not relevant for the free-text QA adaptation. }
    \end{shownto}
    \label{fig:hidden_vs_no_mc}
\end{figure}

While most cases show a negative gap between hidden and free-text QA performance, there are notable exceptions. \Cref{fig:hidden_vs_no_mc} shows that in the \nameref{el:producer_surplus} element, gpt-4o performed better in the free-text QA adaptation compared to the hidden adaptation. One might expect that the explanation for this positive gap is that the multiple-choice answers in the hidden adaptation were very similar, which may have caused confusion for the \model. However, our analysis shows that when the free-text QA was scored correctly, gpt-4o selected an incorrect answer in the hidden adaptation with at least a $10$\% difference from the correct answer 78\% of the time. This suggests that while the \model could reason effectively about the problem, it struggled to correctly match its reasoning to the multiple-choice options provided.

\begin{figure}[H]
    \centering
    \includegraphics[width=\linewidth]{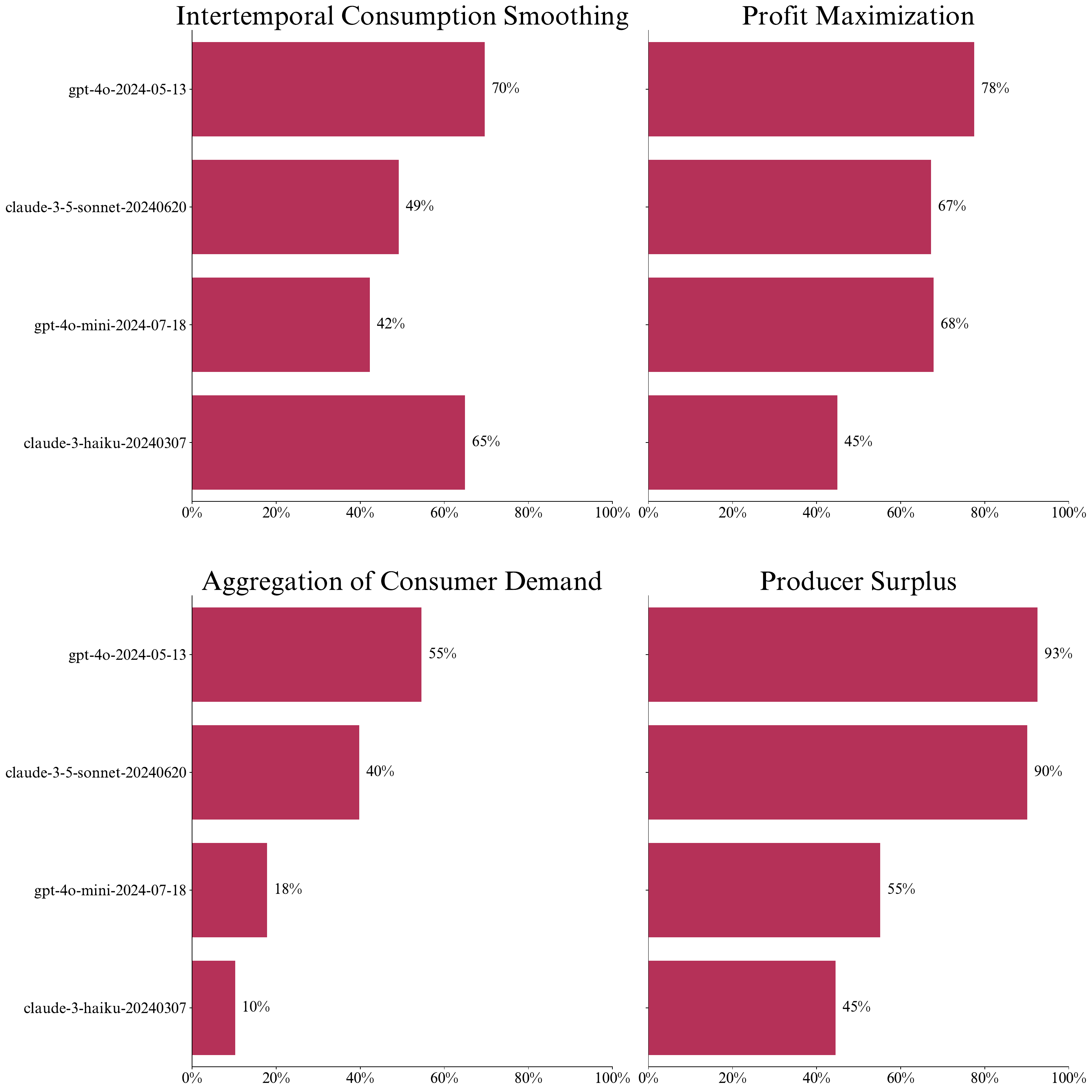}
    \caption{This figure depicts the percentage of time models were incorrect in the free-text adaptation but correct on the hidden adaptation due to choosing the closest answer. The plot compares the performance of four elements---\nameref{el:intertemporal_consumption_smoothing}, \nameref{el:profit_max}, \nameref{el:agg_consumer_demand}, and \nameref{el:producer_surplus}---across all closed-source models.}
    \begin{shownto}{arxiv}
        \Description[]{This figure depicts the percentage of time models were incorrect in the free-text adaptation but correct on the hidden adaptation due to choosing the closest answer. The plot compares the performance of four elements---\nameref{el:intertemporal_consumption_smoothing}, \nameref{el:profit_max}, \nameref{el:agg_consumer_demand}, and \nameref{el:producer_surplus}---across all closed-source models.}
    \end{shownto}
    \label{fig:percent_closest}
\end{figure}

The plot also suggests that the performance gap in the Profit Maximization element is primarily due to the benefit random guessing has on accuracy in the hidden adaptation compared to free-text QA. Furthermore, in the Aggregation of Consumer Demand element, the inclusion of options after the reasoning step offered limited benefit, highlighting that the true advantage lies in including these options during the reasoning process.

These observations highlight an important nuance: although multiple-choice formats generally offer helpful structure for models, they may also hinder performance in certain scenarios.

\newpage
\section{Section 3 Images}\label{appendix:sec3}

\begin{figure}[!ht]
    \centering
    \includegraphics[width=\linewidth]{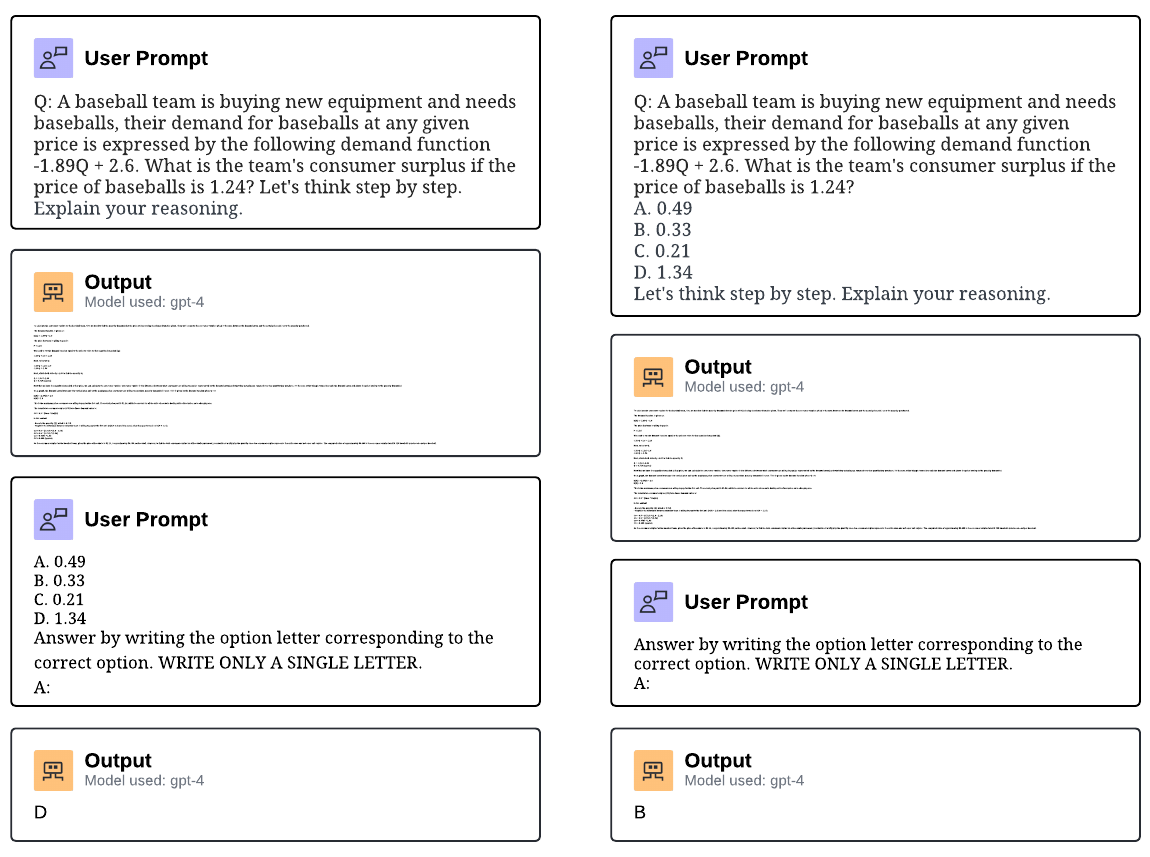}
    \caption{(Left) The \hidden approach to 0-CoT: the model is given only the question and asked to explain its reasoning before being provided with options. (Right) The \shown approach to 0-CoT: the model is presented with both the question and options before explaining its reasoning.}
    \begin{shownto}{arxiv}
        \Description[]{(Left) The \hidden approach to 0-CoT: the model is given only the question and asked to explain its reasoning before being provided with options. (Right) The \shown approach to 0-CoT: the model is presented with both the question and options before explaining its reasoning.}
    \end{shownto}
    \label{fig:hidden_vs_shown}
\end{figure}

\newpage
\section{Web Application}\label{appendix:web}

\begin{figure}[h]
    \centering
    \includegraphics[width=\linewidth]{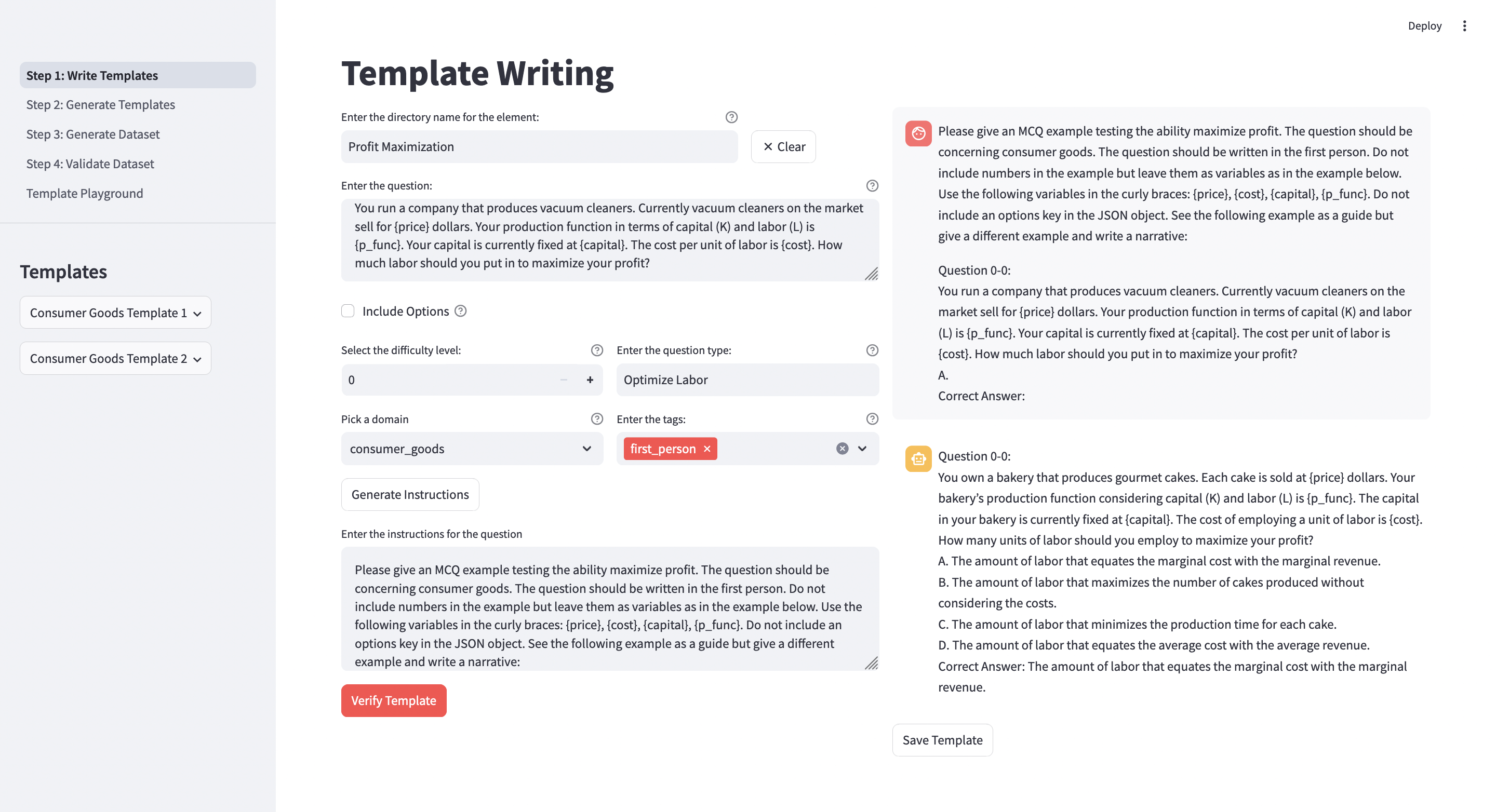}
    \caption{The web app user interface for template writing. This page includes fields for type, domain, grade level and tags (including perspectives). The right shows an example of template verification which uses a \model to generate another template using the example seed.}
    \begin{shownto}{arxiv}
        \Description[]{The web app user interface for template writing. This page includes fields for type, domain, grade level and tags (including perspectives). The right shows an example of template verification which uses a \model to generate another template using the example seed.}
    \end{shownto}
    \label{fig:template_writing}
\end{figure}

\begin{figure}[h]
    \centering
    \includegraphics[width=\linewidth]{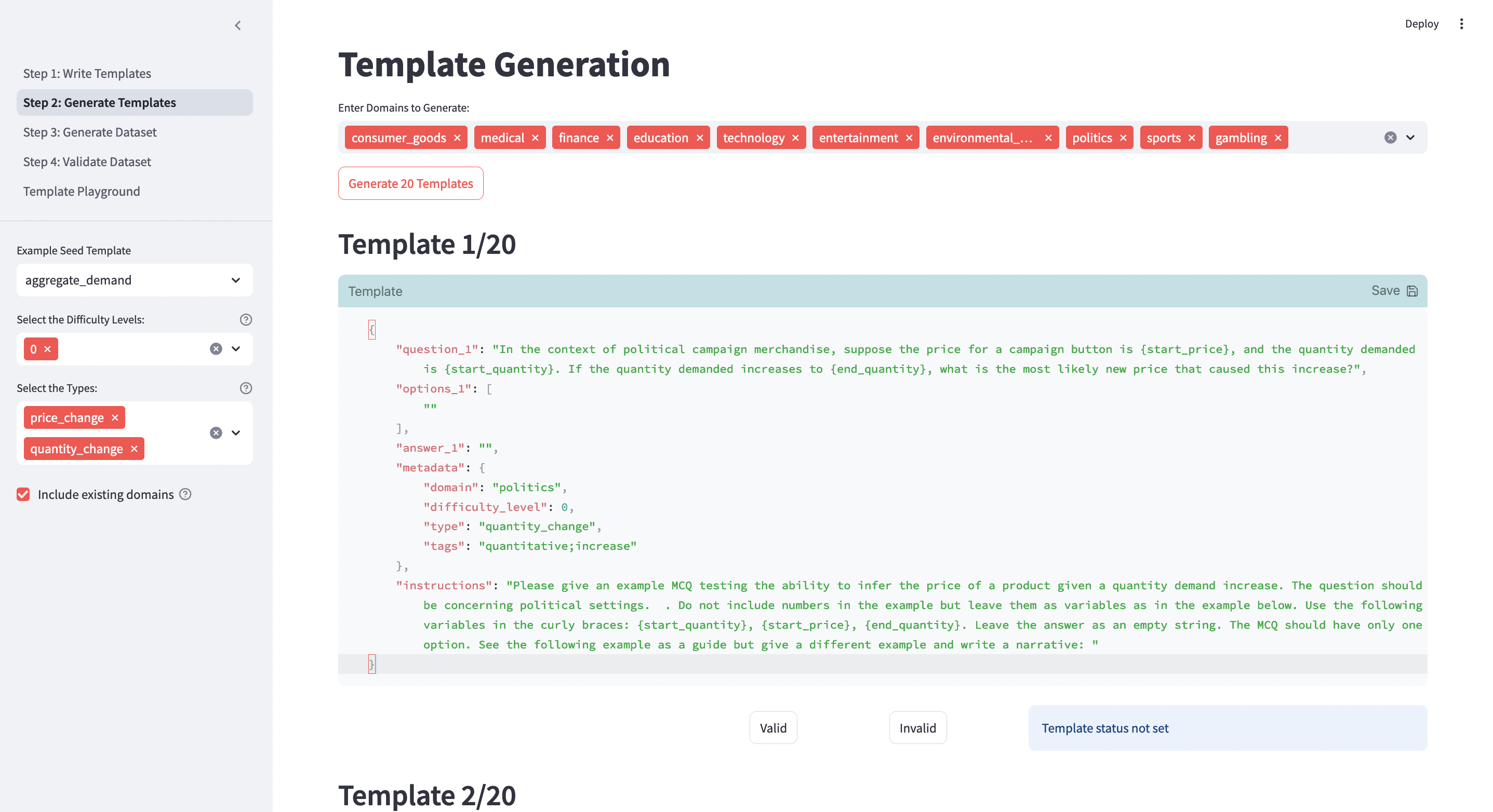}
    \caption{The web app user interface for template generation. This page allows for a selection of domains, and types for which templates will be generated using the available example seeds. Templates can then be verified and saved by the user.}
    \begin{shownto}{arxiv}
        \Description[]{The web app user interface for template generation. This page allows for a selection of domains, and types for which templates will be generated using the available example seeds. Templates can then be verified and saved by the user.}
    \end{shownto}
    \label{fig:template_generation}
\end{figure}

\begin{shownto}{iclr}
    \begin{figure}[h]
        \centering
        \includegraphics[width=\linewidth]{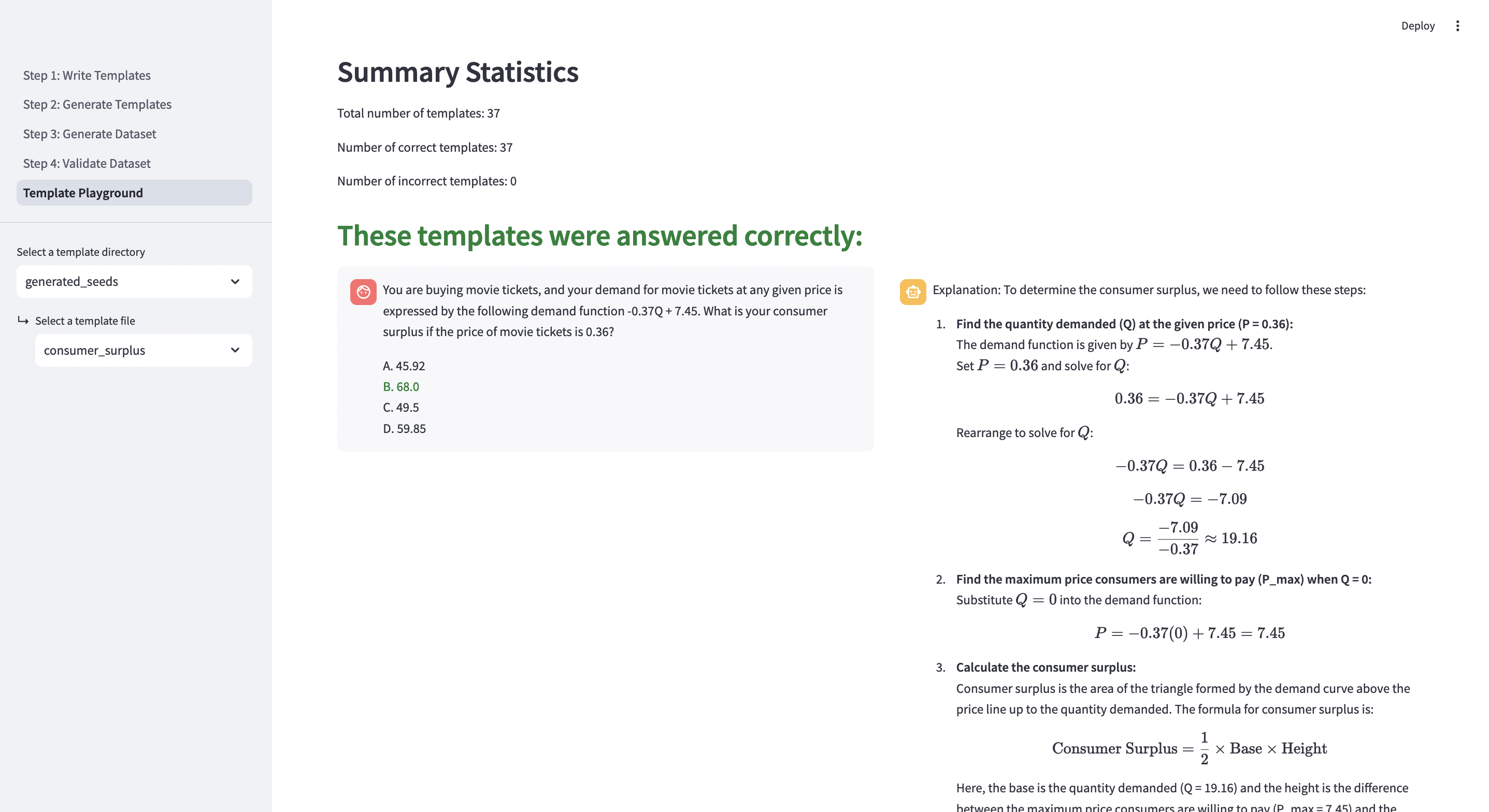}
        \caption{The web app user interface for template AI double-checking. This page instantiates and fills a set of question using a generated or example seed and then generates a response using an OpenAI model. The page also reports the number of questions answered correctly as well as the responses from the model.}
        \begin{shownto}{arxiv}
            \Description[]{The web app user interface for template AI double-checking. This page instantiates and fills a set of question using a generated or example seed and then generates a response using an OpenAI model. The page also reports the number of questions answered correctly as well as the responses from the model.}
        \end{shownto}
        \label{fig:template_ai_validation}
    \end{figure}
\end{shownto}

\begin{figure}[h]
    \centering
    \includegraphics[width=\linewidth]{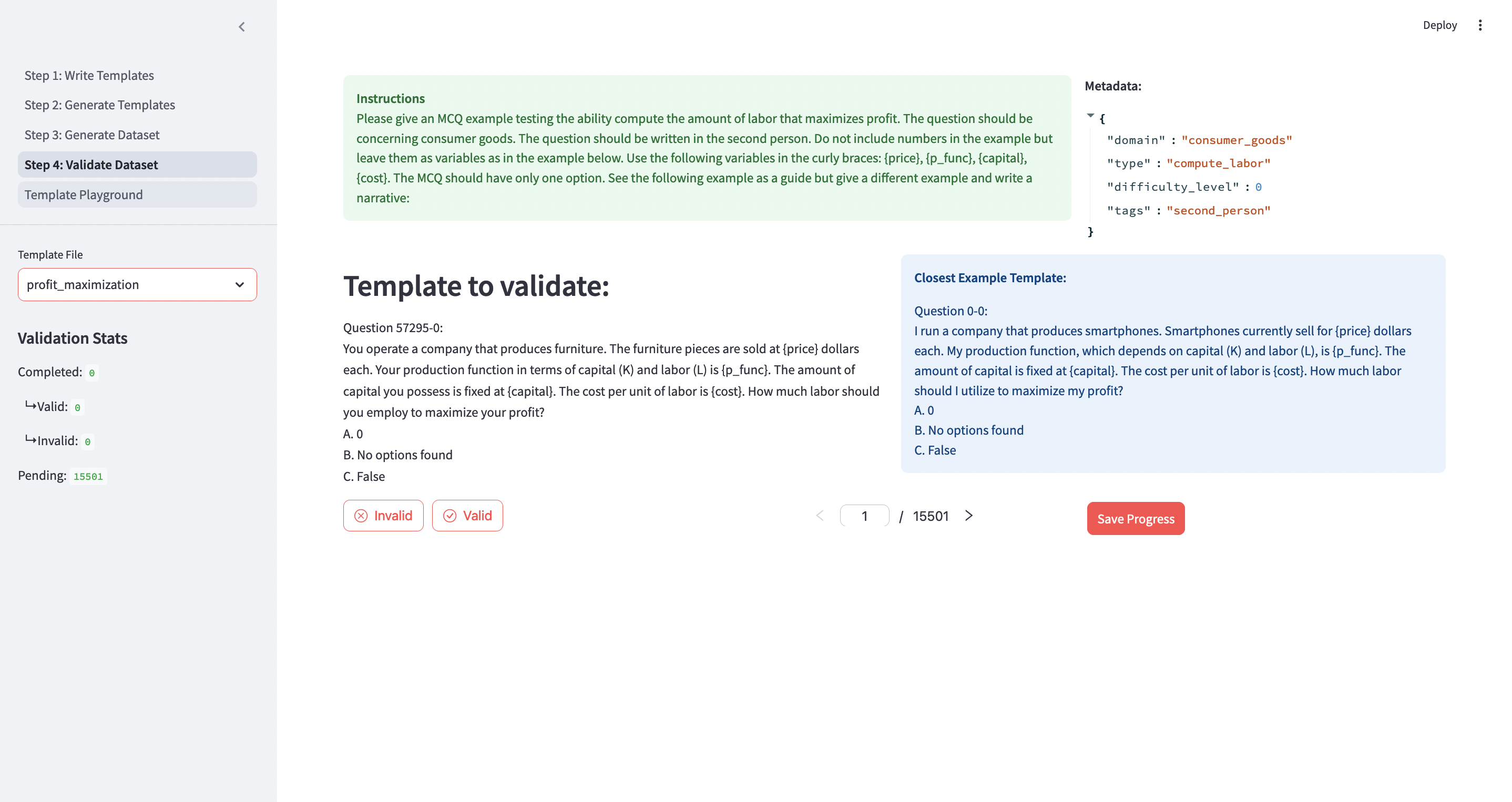}
    \caption{The web app user interface for template validation. This page displays all generated seeds returned by the model for manual validation.}
    \begin{shownto}{arxiv}
        \Description[]{The web app user interface for template validation. This page displays all generated seeds returned by the model for manual validation.}
    \end{shownto}
    \label{fig:template_validation}
\end{figure}

\end{document}